\pdfoutput=1

\documentclass[11pt]{article}

\usepackage[preprint]{acl}

\usepackage{times}
\usepackage{latexsym}

\usepackage[T1]{fontenc}

\usepackage[utf8]{inputenc}

\usepackage{microtype}

\usepackage{inconsolata}

\usepackage{multirow}
\usepackage{graphicx}
\usepackage{multirow}
\usepackage{makecell}
\usepackage{stackengine}
\setstackEOL{\cr}
\usepackage{hyperref}
\usepackage{float}
\usepackage[none]{hyphenat}
\usepackage{comment}
\usepackage{booktabs}
\usepackage{longtable}
\usepackage{listings}
\usepackage{array}
\usepackage{threeparttable}

\usepackage{tikz}
\def\checkmark{\tikz\fill[scale=0.2](0,.35) -- (.25,0) -- (1,.7) -- (.25,.15) -- cycle;}

\newcommand{\mycomment}[1]{}

\title{\textsc{Batayan}: A Filipino NLP benchmark for evaluating \\ Large Language Models}

\author{
 \textbf{Jann Railey Montalan\textsuperscript{1,2}},
 \textbf{Jimson Paulo Layacan\textsuperscript{3}},
 \textbf{David Demitri Africa\textsuperscript{4}},
 \textbf{Richell Isaiah Flores\textsuperscript{3}},
\\
 \textbf{Michael T. Lopez II\textsuperscript{3}},
 \textbf{Theresa Denise Magsajo},
 \textbf{Anjanette Cayabyab},
 \textbf{William Chandra Tjhi\textsuperscript{1,2}}
\\
\\
 \textsuperscript{1}AI Singapore,
 \textsuperscript{2}National University of Singapore,
 \\
 \textsuperscript{3}Ateneo de Manila University,
 \textsuperscript{4}University of Cambridge
\\
 \small{
   \textbf{Correspondence:} \href{mailto:email@domain}{railey@aisingapore.org}
 }
}

\begin{document}
\maketitle
\begin{abstract}
Recent advances in large language models (LLMs) have demonstrated remarkable capabilities on widely benchmarked high-resource languages. However, linguistic nuances of under-resourced languages remain unexplored. We introduce \textsc{Batayan}, a holistic Filipino benchmark that systematically evaluates LLMs across three key natural language processing (NLP) competencies: understanding, reasoning, and generation. \textsc{Batayan} consolidates eight tasks, three of which have not existed prior for Filipino corpora, covering both Tagalog and code-switched Taglish utterances. Our rigorous, native-speaker-driven adaptation and validation processes ensures fluency and authenticity to the complex morphological and syntactic structures of Filipino, alleviating the pervasive translationese bias in existing Filipino corpora. We report empirical results on a variety of open-source and commercial LLMs, highlighting significant performance gaps that signal the under-representation of Filipino in pre-training corpora, the unique hurdles in modeling Filipino's rich morphology and construction, and the importance of explicit Filipino language support. Moreover, we discuss the practical challenges encountered in dataset construction and propose principled solutions for building culturally and linguistically-faithful resources in under-represented languages. We also provide a public evaluation suite as a clear foundation for iterative, community-driven progress in Filipino NLP.
\end{abstract}

\section{Introduction}

\begin{figure}[t]
    \centering
    \includegraphics[width=\linewidth]{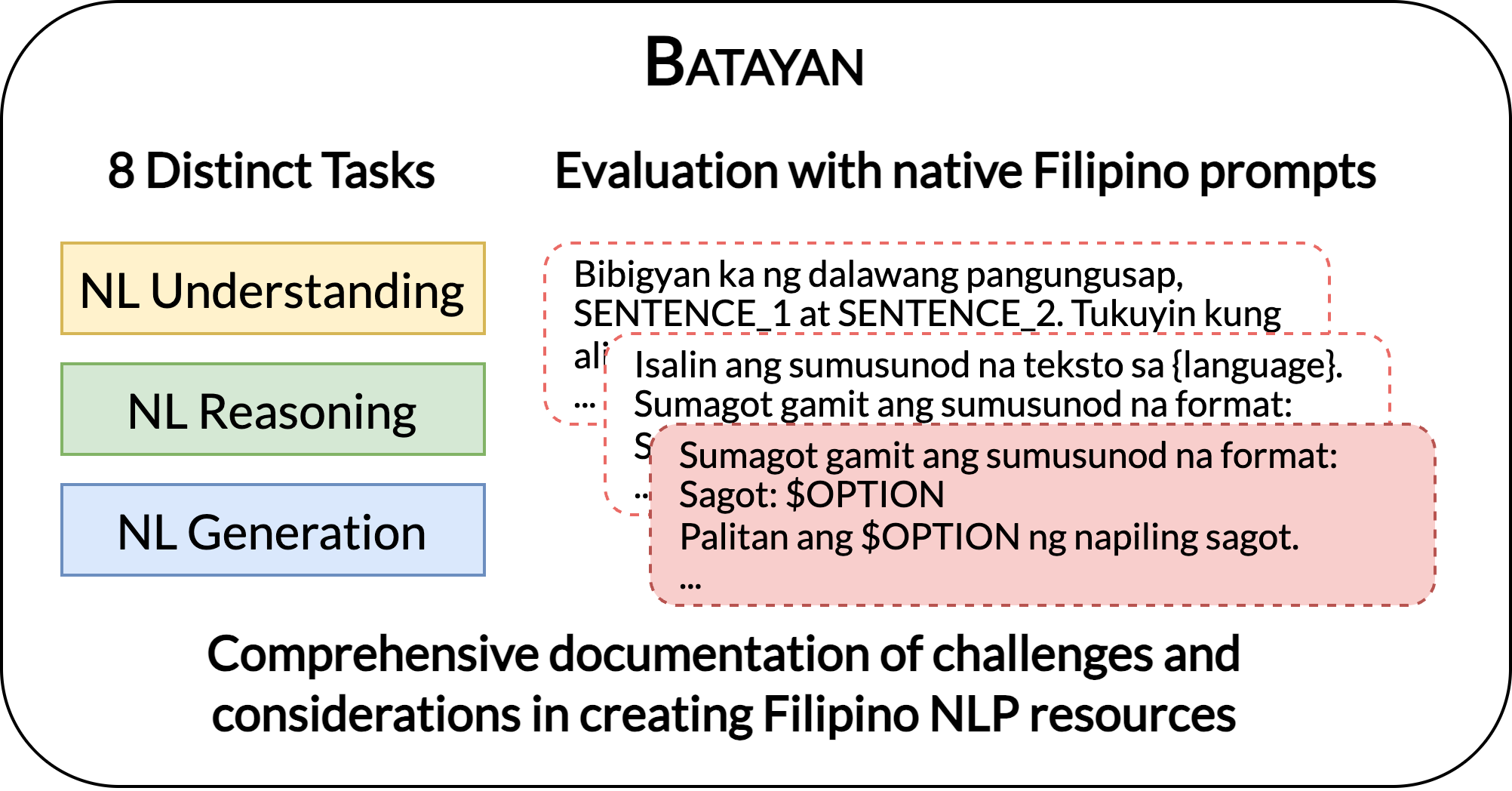}
    \caption{\textsc{Batayan} is a benchmark that holistically evaluates LLM capabilities on a wide range of Filipino language tasks. The rigorous curation and adaptation done by native Filipinos preserves the complexity and authenticity of Filipino language usage today.}
    \label{fig:batayan_contributions}
\end{figure}

Spurred on by recent advances in computing power, big data, and machine learning, LLMs have come into widespread use due to the emergence of a variety of novel and useful capabilities at scale 
\cite{hadi2023survey}. These capabilities have made user applications based on LLMs some of the fastest-growing consumer applications in human history, but have also rendered previous benchmarks insufficiently difficult and diverse \cite{brickscurrent, yang2023rethinking}.

Benchmarks are standard datasets used to measure and compare the performance of models against one another. In particular, the increasingly general capabilities of LLMs have necessitated holistic benchmarks which test a diversity of metrics like fairness, truthfulness, and robustness, as well as a variety of tasks like text summarization, casual reasoning, and translation \cite{yang2023rethinking, guo2023evaluating, liu2023trustworthy}. The vast majority of LLM benchmarks evaluate tasks in English, with non-English works being much fewer \cite{liu-etal-2021-visually, son2024mm}.

Filipino, despite being the national language of the Philippines and being spoken by over 80 million people, remains an under-resourced language. It is under-represented in multilingual LLMs, and existing corpora are domain-specific, non-multilingual \cite{cajote2024philippine}, and created by non-native speakers \cite{quakenbush2005philippine, dita2009building}.
Filipino is often excluded from multilingual LLM benchmarks, suffer from limited task diversity, or have serious deficiencies in grammatical correctness, completeness, and diversity \cite{bandarkar-etal-2024-belebele}.

Moreover, Filipino is a complex language which exhibits a highly complex linguistic structure, particularly in its rich morphological system \cite{ramos2021tagalog, go2017using}. Its agglutinative nature allows for extensive use of affixation and creating nuance through prefixes, infixes, suffixes, and circumfixes \cite{archibald2001contemporary, jubilado2004philippine}. These affixes, combined with root words, allow intricate verb conjugations for marking tense, focus, and mood \cite{zamar2022filipino}. Filipino incorporates elements from a wide array of linguistic influences, such as Spanish \cite{bowen1971hispanic, wolff2001influence}, Chinese \cite{gonzales2022interactions, reid2018modeling, chan1980hokkien}, and Malay \cite{wardana2022lexicostatistics, baklanova2017types}, with codeswitching between English and Filipino being common \cite{bautista1991code, bautista2004taglish}.


Our threefold contributions attempt to address these challenges. \textbf{First, we present \textsc{Batayan},\footnote{The word ``\textit{batayan}'' means ``basis'' or ``ground truth''.} a holistic Filipino benchmark for evaluating LLMs across 8 distinct natural language processing (NLP) tasks} spanning understanding, reasoning, and generation. We place a rigorous emphasis on authenticity to natural Filipino language use through native speaker translation and annotation, addressing translationese bias and other limitations of existing Filipino datasets. \textsc{Batayan} is released as part of SEA-HELM, a publicly available evaluation suite\footnote{\href{https://github.com/aisingapore/sea-helm}{https://github.com/aisingapore/sea-helm}} and leaderboard\footnote{\href{https://leaderboard.sea-lion.ai/}{https://leaderboard.sea-lion.ai}} for assessing LLMs across linguistic and reasoning tasks for Southeast Asian languages. \textbf{Second, we provide extensive evaluation results from a comprehensive set of open-source and commercial LLMs on \textsc{Batayan}}, from differing sizes (7B–671B parameters) and Filipino-language representation, revealing significant disparities in model performance across different linguistic capabilities in Filipino. \textbf{Third, we document systematic challenges and methodological considerations in creating high-quality Filipino NLP datasets}, particularly highlighting issues in translation fluency, vocabulary adaptation, and the preservation of Filipino's rich morphological features. These insights provide a practical framework for future development of Filipino language resources.

\section{Considerations for Filipino Evaluations}
\label{sec:considerations}

\mycomment{First, we motivate various linguistic considerations in creating \textsc{Batayan}, such as the choice to allow Taglish or choosing between grammatically valid word orders.}

\renewcommand{\arraystretch}{1.2}
\begin{table*}[t]
\centering
\tiny
\begin{threeparttable}
\begin{tabular}{lllrllllc}
\toprule
\textbf{Comp.} & \textbf{Task} & \textbf{Dataset} & \textbf{\# test samples} & \textbf{Target}\tnote{1} & \textbf{Language} & \textbf{Source}\tnote{2} & \textbf{Adaptation} & \textbf{Our contribution}\\
\midrule
\multirow{1}{*}{NLU} & PI & PAWS \cite{paws2019naacl} & 2,000 & label (2) & English & English-sourced & Native translation & \checkmark \\
& QA & Belebele \cite{bandarkar-etal-2024-belebele} & 900 & span (4) & Filipino & English-adapted & Native re-translation & \checkmark \\
& SA & PH Elections \cite{cabasag2019hate} & 5,160 & label (3) & Taglish & Natively-sourced & No adaptation & \\
& TD & PH Elections \cite{cabasag2019hate} & 5,160 & label (2) & Taglish & Natively-sourced  & No adaptation & \\
\midrule
\multirow{1}{*}{NLR} & CR & Balanced COPA \cite{kavumba-etal-2019-choosing} & 500 & label (2) & English & English-sourced & Native translation & \checkmark \\
& NLI & XNLI \cite{conneau2018xnli} & 5,010 & label (3) & English & English-sourced & Native translation & \checkmark \\
\midrule
\multirow{1}{*}{NLG} & AS & XL-Sum \cite{hasan-etal-2021-xl} & 11,535 & summary & English & English-sourced & Native translation & \checkmark \\
& MT & FLORES 200 \cite{nllb2022} & 1,012 & translation & Filipino & English-adapted & Native re-translation & \checkmark \\
\bottomrule
\end{tabular}
\begin{tablenotes}
    \item[1] Number of options are shown in parenthesis
    \item[2] English-adapted: previously translated from English; English-sourced: originally in English; Natively-sourced: originally in Taglish
\end{tablenotes}
\caption{Source datasets for each task in \textsc{Batayan}.}
\label{tab:dataset_sources}
\end{threeparttable}
\end{table*}
\renewcommand{\arraystretch}{1.0}

\subsection{Language of Evaluation}
\label{subsec:language_of_evaluation}
While Filipino is the official language of the Philippines, it has been argued in literature that the de facto \textit{lingua franca} is Taglish, the practice of code-switching between English and Tagalog \cite{go2013tagalog}. Naturally-occurring datasets in Filipino (e.g., mined from social media, recorded interviews) typically contain some code-switching \cite{bautista2004taglish}. We defer the problematization of the debated difference between Tagalog and Filipino, and use the two terms interchangeably.

Filipino, while text-rich, is considered under-resourced, lacking the linguistic data, tools, and resources for effective natural language processing \cite{miranda2023nlp}. To address data scarcity, developers take advantage of high-resource languages such as English through translation \cite{goyal-etal-2022-flores, doddapaneni-etal-2023-towards}. Furthermore, leveraging multilingual datasets is potent not only because it enhances the performance of models trained on limited resources, but also because code-switching and bilingualism is a common phenomenon in the Philippines \cite{tupas2017bilingual}.


\subsection{Language Design Principles}
The design of \textsc{Batayan} emphasized linguistic authenticity and representativeness in word choice, sentence structure, and grammar. In terms of language use and word choice, we employed common Filipino and Taglish vocabulary, balancing loanwords with native Filipino terms by prioritizing colloquial usage. This was also done when words had both English and Filipino variants, taking into account spelling and orthographic preferences prevalent among native speakers.

Regarding sentence structure, we prioritized sentences with attention to natural syntax and the choice of \textit{ayos} (sentence arrangement), namely direct (\textit{karaniwang ayos}, KA, \textit{lit.} usual order) or inverted (\textit{di-karaniwang ayos}, DKA, \textit{lit.} unusual order) forms \cite{tanawanetal2008istruktura}. In KA, Filipino constructions follow the typical predicate-initial word order \cite{malicsi2013gramar}. In contrast, DKA, which is also referred to linguistically as an \textit{ay}-inversion, is a type of construction in Filipino where non-predicative constituents (such as \textit{simuno}, \textit{lit.} grammatical subject) are ``fronted'' or shifted to precede the predicate, marking them as the ``topic'' of the sentence \cite{kroeger1993phrase}. Inverted constructions are often used in formal settings, and could therefore be deemed unnatural in most situations \cite{pizarro2010revisiting}.  This characteristic of Filipino is of interest due to our observation of machine translations preferring DKA (see Section \ref{subsec:issues-english}); hence, the further need for native re-translations with a preference for KA.

\subsection{Related Datasets and Benchmarks}

Previous work on Filipino language model evaluation has been largely fragmented. Researchers have made efforts in developing datasets for Filipino language tasks such as fake news detection \cite{cruz-etal-2020-localization}, named entity recognition \cite{miranda2023developing}, natural language inference \cite{cruz2021entailment},  sentiment analysis \cite{villavicencio2021twitter}, and others. However, the field lacks a unified, comprehensive benchmark for systematic evaluation of LLMs across different linguistic dimensions in Filipino. 

Unlike existing Southeast Asian benchmarks that either overlook Filipino or treat it through machine translation, \textsc{Batayan} is the first holistic, native-speaker-driven evaluation suite tailored specifically for Filipino NLP. For instance, BHASA provides a multilingual SEA benchmark covering Indonesian, Tamil, Thai, and Vietnamese, but it does not include Filipino \cite{leong2023bhasaholisticsoutheastasian}. Similarly, IndoNLU focuses exclusively on Indonesian understanding tasks and lacks any generative or reasoning evaluations \cite{wilie-etal-2020-indonlu}. SeaExam and SeaBench introduce local exam and open-ended Vietnamese, Thai, and Indonesian queries but exclude Filipino entirely \cite{liu-etal-2025-seaexam}, and SailCompass evaluates robustness for SEA languages without any Filipino components \cite{guo2024sailcompass}. SeaEval provides a broad multilingual and multicultural benchmark across 29 datasets for Southeast Asian languages, including Filipino, but still relies primarily on machine translations and generic multilingual prompts \cite{wang-etal-2024-seaeval}. 

In contrast, \textsc{Batayan} integrates eight tasks—including three novel Filipino-centric ones—grounded in native translations and rigorous acceptability judgments, ensuring fluency, authenticity, and sensitivity to Filipino’s complex morphological and code-switching patterns.

\section{Task and Dataset Curation}
\label{sec:task_and_dataset_curation}

\subsection{Task Selection} 
\label{subsec:task_selection}


Through \textsc{BATAYAN}, we significantly expand the scope of Filipino LLM evaluation by introducing tasks that assess a gamut of NLP competencies: understanding (NLU), generation (NLG), and reasoning (NLR). Our Filipino benchmark incorporates machine translation (MT), natural language inference (NLI), question answering (QA), sentiment analysis (SA), and toxicity detection (TD). We also release three novel tasks that have not been previously explored in the Filipino context, namely abstractive summarization (AS), causal reasoning (CR), and paraphrase identification (PI).

Our benchmark design and selection of tasks are informed by established multilingual evaluation frameworks, particularly BHASA \cite{leong2023bhasaholisticsoutheastasian} and XTREME \cite{10.5555/3524938.3525348}.

\subsection{Dataset Selection}
\label{sec:dataset_selection}

To construct \textsc{Batayan}, we curated open-source corpora with clear provenance when possible, prioritizing datasets that exhibit authentic language use across various domains such as social media, news articles, and other publicly available texts. We identified existing Filipino-language datasets for MT, QA, SA, and TD. Meanwhile, the AS, CR, NLI, and PI tasks lacked an expert-annotated, natively-sourced Filipino corpus. Given the scarcity of high-quality Filipino corpora for these tasks, we chose to natively translate and adapt existing English corpora into Filipino following the methods detailed in Section \ref{subsec:dataset_adaptation_and_annotation}. In translation, we deliberately ``naturalized'' the language use by preferring KA, colloquial terms, and other adaptations to preserve native Filipino speech. 

Table \ref{tab:dataset_sources} outlines key attributes of each dataset, such as output type, domain, language, source, and any adaptations that we applied to ensure alignment with natural Filipino language construction. More detailed statistics can be found in Appendix \ref{sec:dataset_overview}.

\subsection{Dataset Adaptation and Annotation}
\label{subsec:dataset_adaptation_and_annotation}

\begin{figure}[t]
    \centering
    \includegraphics[width=\linewidth]{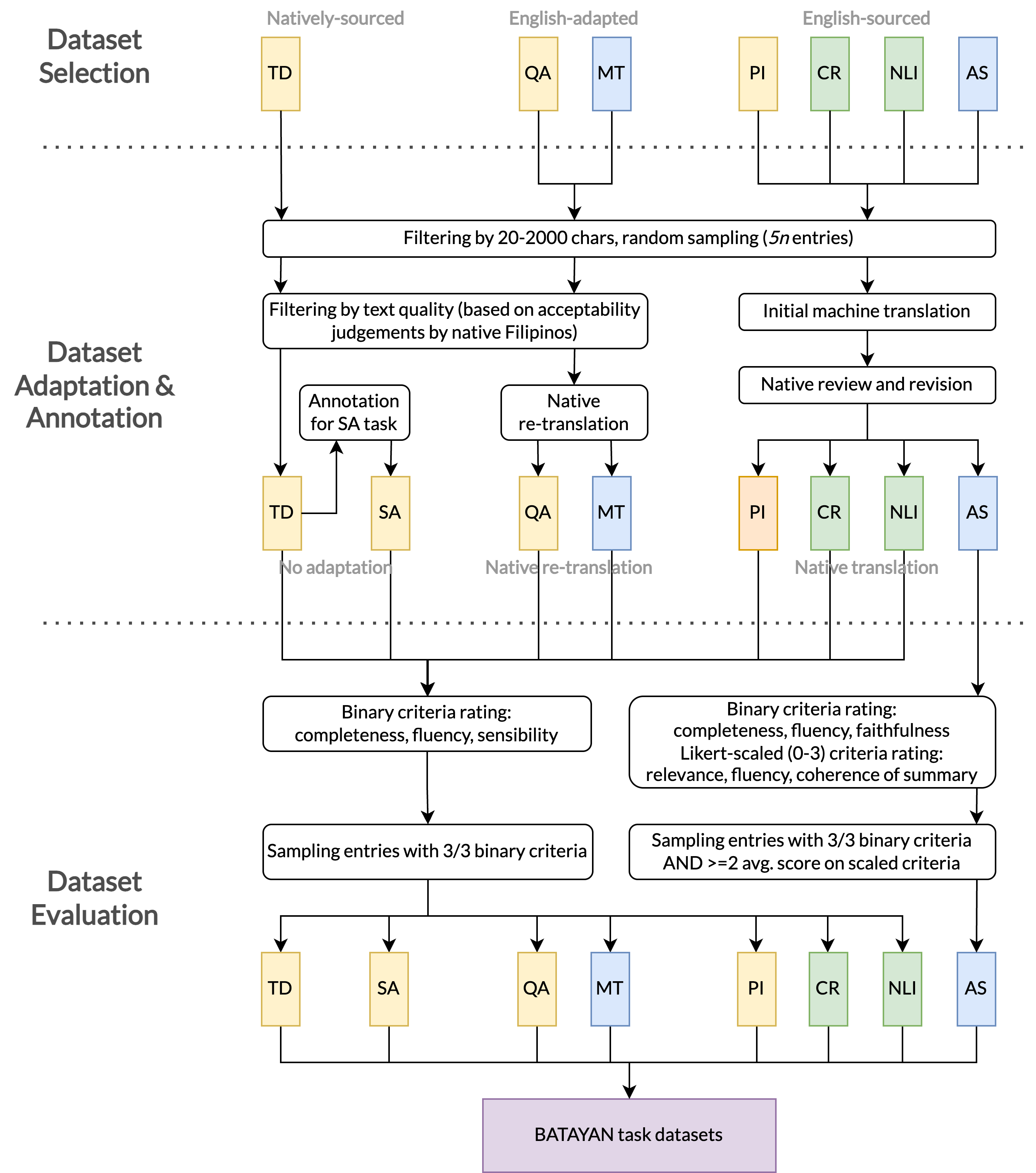}
    \caption{Dataset curation process for \textsc{Batayan}.}
    \label{fig:batayan_process}
\end{figure}

For each task, entries were filtered by length (20–2000 characters), then randomly sampled $5n$ entries for a target size of $n$, maintaining balanced class distributions. 

\textbf{Native speaker acceptability judgments for quality filtering.} For tasks that were natively-sourced (such as TD) or were previously translated from English to Filipino (such as MT, QA), we filtered entries by text quality. Our evaluation of language quality was driven by acceptability judgments from native Filipino speakers\footnote{Demographics: Working professionals, research faculty, and university students, aged between 21–30 years old. Native bilingual speakers and code-switchers of Filipino and English. Lived in the Philippines for a majority of their lives.} in groups of three so that context and natural language variation are appropriately considered. Relying on standardized automated metrics would be less effective for languages with rich morphological and syntactic traits \cite{rivera2022machine} such as Filipino.

\textbf{Annotation for SA task.} We chose to repurpose the natively-sourced TD dataset due to the authenticity and diversity of registers in the dataset. Three native speakers annotated the Philippine election-related tweets \cite{cabasag2019hate} dataset for the SA task with three sentiment polarities: positive, negative, and neutral. Inter-annotator agreement was calculated using Cohen's kappa \cite{cohen1960coefficient}, which is the difference between observed agreement and the probability of chance agreement over the probability that the raters did not agree by chance, and Krippendorff’s alpha \cite{krippendorff2018content}, which is 1 minus the ratio of observed disagreement by raters and expected chance disagreement). The resulting Cohen’s kappa of 0.8202 and Krippendorf’s alpha of 0.8268 indicate substantial agreement. For both TD and SA tasks, we chose to maintain the naturally-occurring language to capture the nuances of code-switched Taglish usage prevalent in social media in the Philippines.

\textbf{Authoritative translation guidelines.} For datasets that were adapted into Filipino, a critical aspect of dataset design was ensuring compliance with official guidelines set by the \textit{Komisyon sa Wikang Filipino} (Commission on the Filipino Language) of the Philippines. In translation, we followed prescribed principles aimed at preserving the source material's intent while adjusting for cultural and contextual relevance \cite{almario2016batayang} and effective writing \cite{almario2014masinop}. Translations were performed by native Filipino speakers in the identified \textit{lingua franca}, whenever appropriate, to maintain semantic equivalence without resorting to overly literal phrasing.

\textbf{Native re-translation of previously translated datasets.} We observed that the QA and MT datasets were of subpar quality and suffered from issues with inaccuracies and translationese \cite{gellerstam1986translationese, riley_translationese_2020} such as the use of unnaturally formal language or inappropriate word choices. As such, these were re-translated to enhance contextual and linguistic accuracy. 

\textbf{Native translation of English datasets.} For multilingual tasks that had no equivalent Filipino subsets (such as AS, CR, NLI, and PI), we generated initial translations using Helsinki NLP's \href{https://huggingface.co/Helsinki-NLP/opus-mt-en-tl}{OPUS} model. This model was selected over other off-the-shelf tools, such as Google Translate, as it demonstrated superior performance in terms of Filipino translation quality and accuracy as tested by the authors. These translations were then manually revised by native Filipino speakers to generate high-quality and fluent translations. Our native-speaker-driven evaluation process focused on cultural and linguistic relevance to ensure that the translated texts resonate with Filipino contexts.

\subsection{Dataset Evaluation}
\label{subsec:dataset_evaluation}

\textbf{Human evaluation for data quality.} Each sample in each dataset underwent further quality assessment by teams of three raters against three binary criteria: completeness, fluency, and sensibility. Samples were assessed as complete only if they contained well-formed sentences, excluding standalone dependent clauses or fragmentary headers and titles lacking complete meaning. Fluency was evaluated on native-like constructions based on multiple factors including appropriate verb conjugations (e.g., avoiding constructions like ``*\textit{Nagpapasalamatan ako sa iyo}'', \textit{lit.} ``*I give thanks to you.''), natural word choices (avoiding awkward phrases like ``*\textit{Si Pacquiao ang kamao ng bansa}'' instead of ``\textit{Si Pacquiao ang pambansang kamao}'', \textit{lit.}: ``Pacquiao is the nation's fist''), and preference for KA sentence structure where contextually appropriate. Samples were assessed as sensible if relevant to the task, screening for confounding factors language capability (e.g., incorrect answers in QA mislabeled as correct or requiring re-translation).

\textbf{Additional criteria for abstractive summarization.} We employed additional criteria based on previous work \cite{leong2023bhasaholisticsoutheastasian} to assess the appropriateness of the summary in reference to the content of the provided passage. We scored samples on binary criteria (completeness, fluency, and faithfulness) and a Likert-scaled (0-3) criteria (relevance, fluency, and coherence of the summary), with three independent raters per entry. For faithfulness, raters evaluated the factual consistency between the summary and source article, with particular attention to fact preservation and penalization of hallucinated content. For relevance, raters evaluated both coverage of essential information and information filtering, penalizing summaries containing redundancies or excess details. For fluency, raters assessed for proper formatting, appropriate capitalization, and grammatical construction, with higher scores awarded to summaries exhibiting natural Filipino language patterns. For coherence, raters assessed how effectively information flowed between sentences, with higher scores given to summaries that built logically from sentence to sentence to create a cohesive narrative about the topic.

\subsection{Dataset Statistics}
\label{subsec:dataset_statistics}

\renewcommand{\arraystretch}{1.0}
\begin{table}[t]
    \centering
    \tiny
    \begin{tabular}{lllr}
        \toprule
        \textbf{Competency} & \textbf{Task} & \textbf{Label} & \textbf{\# test samples} \\
        \midrule
        \multirow{1}{*}{NLU} & PI  & True  & 200 \\
        & & False  & 200 \\
        & QA  & span & 100 \\
        & SA  & Negative (\textit{Negatibo}) & 200 \\
        & & Neutral (\textit{Neutral})  & 200 \\
        & & Positive (\textit{Positibo})  & 200 \\
        & TD  & Clean (\textit{Malinis}) & 200 \\
        & & Toxic (\textit{Mapoot})  & 200 \\
        \midrule
        \multirow{1}{*}{NLR} & CR  & Cause (\textit{Sanhi})  & 200 \\
        & & Effect (\textit{Bunga})  & 200 \\
        & NLI  & Contradiction & 200 \\
        & & Entailment & 200 \\
        & & Neutral & 200 \\
        \midrule
        \multirow{1}{*}{NLG} & AS  & summary & 100 \\
        & MT & \textit{eng}→\textit{tgl} translation & 600 \\
        & & \textit{tgl}→\textit{eng} translation & 600 \\
        \midrule
        Total & & & 3,800 \\
        \bottomrule
    \end{tabular}
    \caption{Class distribution per task in \textsc{Batayan}.}
    \label{tab:dataset_distribution}
\end{table}
\renewcommand{\arraystretch}{1.0}

For all datasets except AS, only entries that were unanimously rated as demonstrating all 3/3 binary criteria were sampled for the final test splits. For AS, only entries that received a unanimous 3/3 binary criteria assessment and at least and average score of 2 for the scaled criteria were considered for inclusion, ensuring strict maintenance of factual integrity. In cases where initial sampling yielded insufficient qualifying entries, deficient samples were carefully corrected by authors through targeted improvements to grammar, translation of English passages, or other identified issues while maintaining authentic Filipino language patterns.

Joint agreement on our evaluation criteria is presented in Appendix \ref{sec:agreement}. We present joint agreement over traditional metrics such as Cohen’s kappa and Krippendorff’s alpha because metrics that use chance correction are less suitable for binary categorical tasks \cite{powers-2012-problem} and classification tasks in general \cite{delgado2019cohen, charles2016not}.

Overall, \textsc{Batayan} provides 8 distinct tasks with 3,800 test instances. Table \ref{tab:dataset_distribution} provides the distribution of classes for each dataset, and more detailed statistics can be found in Appendix \ref{sec:dataset_overview}. We also provided example splits composed of 5 entries that serve as exemplars for few-shot prompting.

\section{Challenges with Developing a Filipino Benchmark}
This section discusses issues in creating \textsc{Batayan}, shedding light on issues that future researchers may encounter in creating new Filipino datasets or adapting previous ones.

\subsection{Issues with English-adapted and English-sourced Data} \label{subsec:issues-english}

\textbf{Creating more relevant summaries for XL-Sum.} Previous work has noted that a significant portion of reference summaries in the XL-Sum dataset is highly abstractive, demonstrates factual errors, or contains information not mentioned in the provided articles \cite{guo-etal-2022-questioning}, putting to question its factuality and validity. To address this, we developed new Filipino summaries for each article. We ensured relevance and fluency by including only the most important information that can be directly lifted from the article using natural-sounding and grammatically-correct constructions.

\textbf{Limitations in adapting vocabulary.} Adapting the English tasks to Filipino revealed issues in maintaining the intelligibility of the text. Technical terms that exist in English were difficult to translate into Filipino, with several of them having no direct translations. For example, an instance from Belebele included the term ``rule of thirds'', which describes a specific photography technique. Originally, the translation of this phrase in the Tagalog subset of Belebele was ``\textit{tuntunin ng mga sangkatlo}'' (\textit{lit.} ``rule of thirds''), which does not make sense in Filipino and does not convey the intended meaning. For jargon such as this, we chose to keep the original English phrasing. 

English idiomatic expressions were also prevalent and required a more flexible approach in adaptation to ensure the integrity of meaning. For example, one premise from Balanced COPA was, ``The criminal turned himself in'', using the idiomatic expression ``turned himself in'' to mean ``surrendered''. We chose to translate this into Filipino as, ``\textit{Isinuko ng kriminal ang kanyang sarili}'' (\textit{lit.}: ``The criminal surrendered himself''). 

Moreover, the presence of English words that have homonyms added to the complexity of translating English sentences to Filipino. For example, some sentences from PAWS contained the adjective ``right'', which can refer to either the direction (in Filipino: ``\textit{kanan}'') or an assertion of the correctness of the object it is modifying (in Filipino: ``\textit{tama}''). In such cases, we inferred the most probable meaning of the words from the context of the entire passage and then identified the corresponding Filipino word to be incorporated in our Filipino translation.

\textbf{(Dis)fluency in expert and machine translations.} 
Our rigorous review of existing English datasets with machine and expert-guided translations in Filipino revealed weaknesses in their adaptation. The authors found that the automatic translations were unnatural, disfluent, and showed characteristics of translationese despite being grammatically correct. Notably, we observed that these initial translations demonstrated an unusual preference with using \textit{ay}-inversion, and used Filipino terms that were synonymous to their English counterparts but pragmatically-awkward. Hence, we applied re-translation to ensure that the samples sounded more native and natural.

For example, one passage from the English subset of Belebele was, ``The CCTV would certainly send a strong signal...'' The original Tagalog (machine) translation was, ``\textit{Ang CCTV ay tiyak na \underline{magpapadala} ng malakas na  \underline{hudyat}...}'' The term ``\textit{magpapadala}'' here means ``to send'' in the sense of transporting a thing from one place to another, while the term ``\textit{hudyat}'' implies a ``starting sign''. It also uses the unnatural DKA construction or \textit{ay}-inversion. Within the context of the original passage, a more fluent (human) translation would be, ``\textit{Tiyak na \underline{maghahatid} ang CCTV ng malakas na \underline{pahiwatig}...}'', where ``\textit{maghahatid}'' means ``to bring about'' and ``\textit{pahiwatig}'' conveys a sense of ``reminder'' or ``warning'', and the sentence follows the usual KA construction. 

\textbf{Inconsistencies in \textit{ay}-inversions in translation construction.} Styles in translation vary from person-to-person, depending on their valuations and reading of the original text \cite{castagnoli2020translation}. Likewise, \textit{ay}-inversion in the context of translation boils down to a stylistic choice of the translator. 

While KA is argued to be the natural pattern of speech for Filipino speakers \cite{yapan2017bagay}, this valuation of natural and unnatural is debated. \citet{magracia2001panumbas} notes that while DKA is correct in ``meaning, syntactic order, and grammaticality,'' it is not the ``natural'' way of speaking for Filipinos, as it reflects the speaker's language of thought—in this case, the influence of English. The same justification, in terms of source language for translated corpora, can be said for machine-translated text \cite{visweswariah2011word}, though this remains an open question for Filipino.

On the other hand, other research argues that DKA is a necessary aspect for ``discourse continuity'' \cite{bolata2022ang}.  Hence, the more complex the sentence, the higher the chance of utilizing the \textit{ay}-inversion \cite{fox1985word}. We observed this in some datasets such as NLI, which had longer, more context-rich sentences that required the use of DKA in contrast with the shorter, KA-ordered sentences found in the CR task.

These frameworks demonstrate the subjective nature of the selection of Filipino sentence structure. Hence, such inconsistencies may still be observed in \textsc{Batayan} despite being rigorously translated by native speakers.


\subsection{Issues with Natively-Sourced Data}
While the issues with most datasets in \textsc{Batayan} were related with fluency and naturalness, the Philippine election-related tweets \cite{cabasag2019hate} dataset used for the SA and TD tasks presented unique challenges. 

\textbf{Incomplete entries.}
Due to the contextual nature of Filipino and social media communication, many samples lacked sufficient contextual information. To maintain dataset authenticity, we prioritized entries with adequate information for sentiment or toxicity analysis rather than complete removal. This was accomplished by selecting high-agreement samples where native speakers demonstrated consistent task performance.

\textbf{Non-standard orthography.}
The dataset exhibited significant orthographic variation (spelling, capitalization, word boundaries, etc.) due to its natural language origins. We preserved this variation as it reflects authentic Filipino language use \cite{ilao2012comparative, javier2018pagsusuri, caroro2020rules} relevant to toxicity detection and sentiment analysis tasks, aligning with our methodological principles. We maintained quality control by verifying that orthographical variations remained comprehensible to native readers.

\textbf{Class imbalance.} The dataset showed a disproportionate distribution of negative sentiment for certain political entities. We rebalanced the class distribution and prioritized high-agreement samples to mitigate individual annotator bias.

\section{Evaluation}

\textsc{Batayan} serves as the Filipino component of the SEA-HELM leaderboard. The evaluation framework of SEA-HELM follows prior systems such as HELM \cite{liang2023holisticevaluationlanguagemodels}.

\subsection{Evaluation Design}
\label{sec:evaluation_design}

\textbf{Tasks.} The selection of tasks in \textsc{Batayan} mentioned in Section \ref{sec:task_and_dataset_curation} aims to comprehensively evaluate the Filipino language capabilities of LLMs. Each task is a collection of test and example instances composed of an input string, references, metadata, and a ground truth label.

\textbf{Prompts.} For \textsc{Batayan}, we developed prompt templates for each task written in Filipino. We ensured that the instructions for each task are consistent with the prompt template design already used in SEA-HELM. A comprehensive list of these prompt templates can be found in Appendix \ref{sec:prompt_templates}.

During model evaluation, input prompts are constructed using evaluation instances and the corresponding prompt template. The default evaluation setting of \textsc{Batayan} for instruction-tuned models is zero-shot prompting, where model prompts do not include in-context input-label examples. Base pre-trained models without instruction-tuning are evaluated under a five-shot setting. Given the input prompts and decoding parameters (see Appendix \ref{sec:experimental_setup}), a model then generates output completions.

\textbf{Metrics.} We adopt multiple metrics to quantify the models' performance on each task. For each metric, the model completion is treated as the prediction, while the instance label is used as the reference. For the NLU and NLR tasks, we report the macro F1 score. For machine translation (English→Filipino and Filipino→English), we use ChrF++ \cite{popovic-2017-chrf} and MetricX-24 using the \texttt{metricx-24-hybrid-xxl-v2p6-bfloat16} model \cite{juraska-etal-2024-metricx}. For abstractive summarization, we report three metrics: BERTScore \cite{bert-score}, ChrF++, and ROUGE-L F1 from the multilingual implementation of ROUGE \cite{lin-2004-rouge} used in XL-Sum \cite{hasan-etal-2021-xl}. Default package parameter settings are used, and performance scores are based on a single run of the benchmark.

\subsection{Results}

\begin{figure*}[t]
    \centering
    \includegraphics[width=\textwidth]{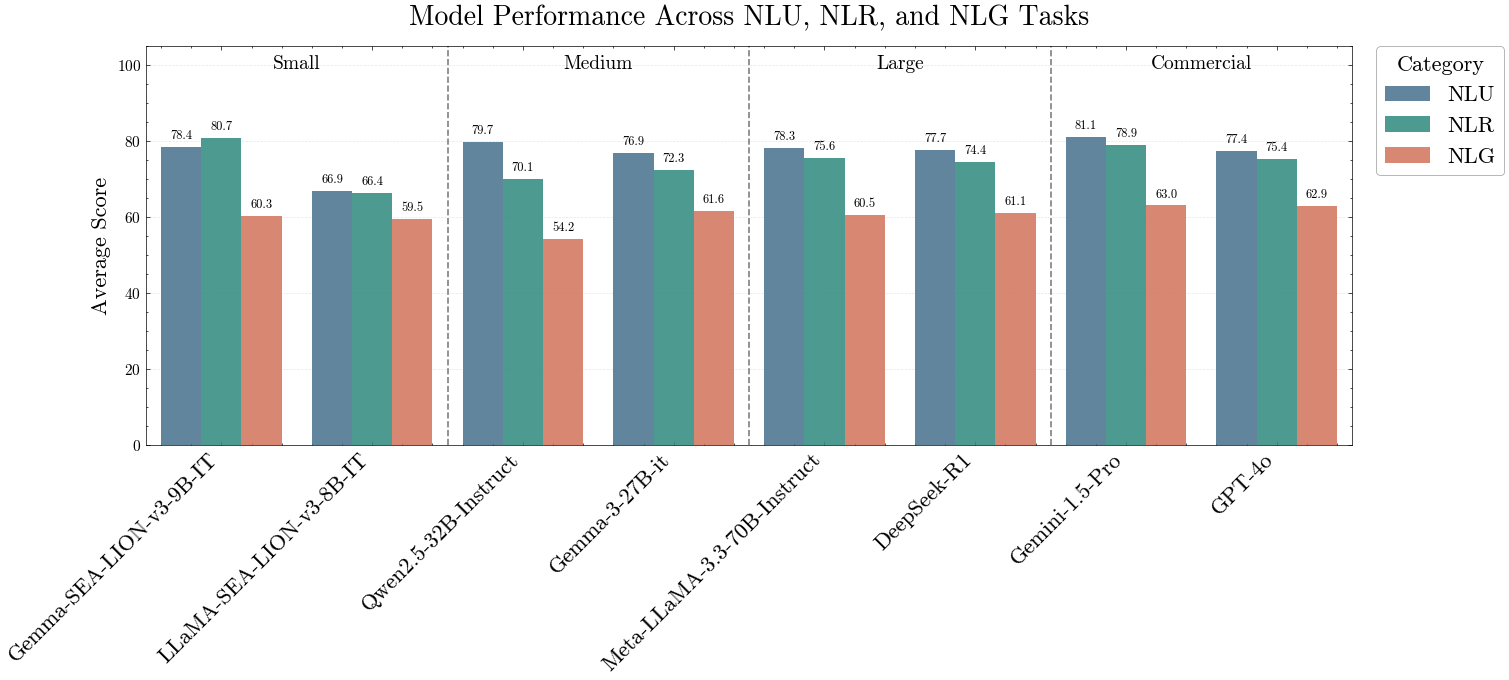}
    \caption{Model performance of selected models on NLU, NLR, and NLG tasks. NLU and NLR tasks are measured using macro F1 score, NLG tasks are measured using ChrF++, MetricX-24, BERTScore, and ROUGE-L F1. The unweighted average across tasks and scores is taken. Targeted Filipino instruction tuning allows \texttt{Gemma-SEA-LION-v3-9B-IT} to achieve competitive performance with significantly larger open-source and commercial models.}
    \label{fig:overall_results}
\end{figure*}

Our evaluation results include over 50 small ($\le$11B), medium ($\le$32B), and large ($\ge$32B) LLMs, including several commercial AI systems on the \textsc{Batayan} benchmark. The results underscore two principal findings: the substantial benefit of explicit Filipino language support and the scaling effects observed across model sizes. Figure \ref{fig:overall_results} highlights several model performances, and the complete experiment results can be found in Appendix \ref{sec:complete_experimental_results}.

\textbf{Filipino language support enhances task-specific performance.} Tables \ref{tab:small_nlu_nlr}–\ref{tab:commercial_nlu_nlr} in Appendix \ref{sec:complete_experimental_results} reveal that models with Filipino instruction tuning (such as the SEA-LION family of models) consistently outperform  counterparts on culturally nuanced tasks. In particular, the Filipino-adapted model \texttt{Gemma-SEA-LION-v3-9B-IT} achieves an average macro F1 of 79.23 among small models, with a striking CR score of 92.75. In contrast, non-Filipino-specific models, such as many variants in the Meta Llama 3 series, exhibit very low or zero performance in CR, which emphasizes the challenge of capturing Filipino’s complex causality and morphology without explicit tuning.

On the NLG competency, tables \ref{tab:small_nlg}–\ref{tab:commercial_nlg} in Appendix \ref{sec:complete_experimental_results} show that while ROUGE-L F1 scores show only modest differences between Filipino-tuned and general models, complementary evaluation using MetricX-24 provides a more sensitive measure of semantic fidelity. For instance, in \textit{eng}→\textit{tgl} translation, Filipino-supported systems like \texttt{Gemma-SEA-LION-v3-9B-IT} and \texttt{Sailor2-8B-Chat} achieve MetricX-24 scores exceeding 87, indicating superior preservation of Filipino-specific discourse and semantics. This observation shows that the apparent parity in ROUGE-L is only in part: when examined alongside semantic metrics, the benefit of dedicated language tuning is clear.

\textbf{Scaling effects have clear trends beyond parameter similarity.} Although many small models lie within a narrow 7B–9B parameter range, our broader evaluation—including medium and large models—illustrates a clear scaling effect. For example, among large models, \texttt{Gemma-SEA-LION-v3-9B-IT} attains an average macro F1 of 78.41 for NLU, macro F1 of 80.69 for NLR, and aggregated score of 60.35 in NLG, significantly outperforming similarly-tuned small models. This trend is evident even when models are compared within the same family: larger variants consistently demonstrate enhanced capability in handling both the nuanced understanding and generation challenges posed by Filipino text. The results thus suggest that increased model capacity—when combined with region-specific instruction tuning—enables a better representation of Filipino linguistic phenomena, including code-switching and agglutinative morphology.

\textbf{Insights from commercial systems.} In addition to open-source models, our evaluation also includes several commercial LLMs (Tables \ref{tab:commercial_nlu_nlr} and \ref{tab:commercial_nlg}). Commercial systems such as Gemini and GPT-4o variants exhibit competitive performance across both understanding and generation tasks. For instance, on NLU and NLR tasks, commercial models achieve average macro F1 scores ranging from 75.14 to 86.23. On the NLG competency, commercial systems also deliver robust results: Gemini and GPT-4o models maintain strong MetricX-24 scores above 88, indicative of high-quality semantic translation in both \textit{eng}→\textit{tgl} and \textit{tgl}→\textit{eng} directions.

While these commercial models benefit from large-scale pre-training and domain-specific optimization, our findings show that open-source, Filipino-tuned models achieve comparable—and in certain tasks, superior—performance. In particular, when cultural and linguistic nuances are critical, explicit regional instruction tuning offers a tangible advantage. The comparable performance of commercial systems reinforces that our dual strategy of targeted language support combined with model scaling is effective in academic settings as well as real-world applications.

Overall, our results provide strong evidence for a dual strategy in multilingual modeling: integrating region-specific instruction tuning with scalable architectures. Filipino-tuned models excel in tasks that require nuanced cultural and morphological sensitivity, and their performance improves markedly with increased capacity. Notably, open-source models fine-tuned with Filipino instructions are competitive with and even surpass state-of-the-art commercial systems on both understanding and reasoning tasks. These findings highlight that explicit Filipino language support not only enhances semantic fidelity and discourse preservation but also positions such models ideally for real-world, production-level applications in the Philippines.

\section{Conclusion}

In this paper, we introduced \textsc{Batayan} for holistically evaluating LLMs on a gamut of Filipino language tasks covering natural language understanding, reasoning, and generation. Our findings show that LLMs with explicit Filipino language support and fine-tuned on Filipino instructions demonstrate better performance on \textsc{Batayan} compared to other models.

As one of the major challenges in developing \textsc{Batayan} is maintaining the naturalness and fluency of the language, we plan to develop metrics and tools that can help discriminate from translationese and natural texts, inspired by prior research \cite{lovenia-etal-2024-seacrowd, riley_translationese_2020}.

\section*{Limitations}
While our work provides a unique and comprehensive Filipino language benchmark, we also reported in previous sections the challenges and limitations in developing high-quality NLP resources for the under-represented language. Additionally, prosodic characteristics such as stress and sarcasm are not immediately obvious in the datasets utilized. We account for this by selecting only high-agreement samples. We also note that the domain of both SA and TD tasks (which were tweets surrounding political events), limits the distributions of these tasks, and as such we recommend utilizing other datasets that cover a wider set of domains and use cases.

While we recognize that datasets with larger amounts of samples are preferred to ensure generalizability, we maintain our position to prioritize consistent, high-quality, and representative samples. The thoroughness and rigor of our review, annotation, and translation processes were essential and maintaining the quality and reliability of our benchmark. We believe that our benchmark serves as a valuable starting point for future research and can certainly be expanded in subsequent iterations.

Further, given the breadth of models and tasks, a comprehensive evaluation incorporating error cases was not possible. We also recognize the importance of evaluation LLMs that specialize on specific tasks such as translation. We hope that a public release of the \textsc{Batayan} codebase and benchmark on a leaderboard can help researchers better characterize the weaknesses of models, conduct their own analysis on the Filipino language capabilities of LLMs, and experiment with specialized models on \textsc{Batayan}.

\section*{Ethical Considerations}

The methodology used in this work was approved by an Institutional Review Board (NUS-IRB-2024-617). We have released the \textsc{Batayan} dataset as part of the SEA-HELM evaluation suite and leaderboard under the Creative Commons Attribution Share-Alike 4.0 (CC-BY-SA 4.0) license, respecting the licenses of the source datasets used in this study.

Due to the nature of the toxicity detection task, we note that the authors were exposed to offensive material. Nonetheless, they were encouraged to report inappropriate samples and were given the option to stop work if desired.

For the annotation of the sentiment analysis task, we involved a quality assurance team consisting of native Filipino speakers. The team, comprised of students from local universities in Singapore, were recruited through public advertisements. These stated the estimated work load and remuneration, which were consistent with university research guidelines and regulatory requirements.

We do not foresee negative social impacts from this paper. Our work introduces Filipino language resources that were reviewed by native Filipino speakers, paying due respect to local cultural sensitivities. We thus do not believe that our research will contribute to over-generalizations regarding Filipino culture.

\section*{Acknowledgments}

This research/project is supported by the National Research Foundation, Singapore under its National Large Language Models Funding Initiative. Any opinions, findings and conclusions or recommendations expressed in this material are those of the authors and do not reflect the views of National Research Foundation, Singapore.

The authors would like to thank National University of Singapore and AI Singapore, especially the AI Products and SEA-LION team, for their unwavering support with this endeavor.

The authors likewise thank the University of Cambridge, Department of Computer Science and Technology, and Magdalene College. David Africa's work is supported by the Cambridge Trust and the Jardine Foundation.

Lastly, the authors would like to thank all the Filipino natives involved in this study for their time and valuable contributions.

\bibliography{custom}

\begin{thebibliography}{91}
\providecommand{\natexlab}[1]{#1}

\bibitem[{Almario(2014)}]{almario2014masinop}
Virgilio~S. Almario. 2014.
\newblock \emph{{KWF Manwal sa Masinop na Pagsulat}}.
\newblock Komisyon sa Wikang Filipino.

\bibitem[{Almario(2016)}]{almario2016batayang}
Virgilio~S. Almario. 2016.
\newblock \emph{Batayang pagsasalin: Ilang patnubay at babasahin para sa baguhan}.
\newblock Komisyon sa Wikang Filipino.

\bibitem[{Archibald and O'Grady(2001)}]{archibald2001contemporary}
John Archibald and William~Delaney O'Grady. 2001.
\newblock \emph{Contemporary linguistics: An introduction}.
\newblock St. Martin's Press.

\bibitem[{Aryabumi et~al.(2024)Aryabumi, Dang, Talupuru, Dash, Cairuz, Lin, Venkitesh, Smith, Campos, Tan, Marchisio, Bartolo, Ruder, Locatelli, Kreutzer, Frosst, Gomez, Blunsom, Fadaee, Üstün, and Hooker}]{aryabumi2024aya23openweight}
Viraat Aryabumi, John Dang, Dwarak Talupuru, Saurabh Dash, David Cairuz, Hangyu Lin, Bharat Venkitesh, Madeline Smith, Jon~Ander Campos, Yi~Chern Tan, Kelly Marchisio, Max Bartolo, Sebastian Ruder, Acyr Locatelli, Julia Kreutzer, Nick Frosst, Aidan Gomez, Phil Blunsom, Marzieh Fadaee, Ahmet Üstün, and Sara Hooker. 2024.
\newblock \href {https://arxiv.org/abs/2405.15032} {Aya 23: Open weight releases to further multilingual progress}.
\newblock \emph{Preprint}, arXiv:2405.15032.

\bibitem[{Baklanova(2017)}]{baklanova2017types}
Ekaterina Baklanova. 2017.
\newblock {T}ypes of {B}orrowings in {T}agalog/{F}ilipino.
\newblock \emph{Kritika Kultura}, 28.

\bibitem[{Bandarkar et~al.(2024)Bandarkar, Liang, Muller, Artetxe, Shukla, Husa, Goyal, Krishnan, Zettlemoyer, and Khabsa}]{bandarkar-etal-2024-belebele}
Lucas Bandarkar, Davis Liang, Benjamin Muller, Mikel Artetxe, Satya~Narayan Shukla, Donald Husa, Naman Goyal, Abhinandan Krishnan, Luke Zettlemoyer, and Madian Khabsa. 2024.
\newblock \href {https://aclanthology.org/2024.acl-long.44} {The {B}elebele {B}enchmark: a {P}arallel {R}eading {C}omprehension {D}ataset in 122 {L}anguage {V}ariants}.
\newblock In \emph{Proceedings of the 62nd Annual Meeting of the Association for Computational Linguistics (Volume 1: Long Papers)}, pages 749--775, Bangkok, Thailand and virtual meeting. Association for Computational Linguistics.

\bibitem[{Bautista(1991)}]{bautista1991code}
Maria Lourdes~S Bautista. 1991.
\newblock {C}ode-switching studies in the {P}hilippines.
\newblock \emph{International Journal of the Sociology of Language}, 8(1).

\bibitem[{Bautista(2004)}]{bautista2004taglish}
Maria Lourdes~S. Bautista. 2004.
\newblock {T}agalog-{E}nglish {C}ode {S}witching as a {M}ode of {D}iscourse.
\newblock \emph{Asia Pacific Education Review)}, 5(2):226--233.

\bibitem[{Bolata(2022)}]{bolata2022ang}
Emmanuel Jayson~V Bolata. 2022.
\newblock Ang {N}oumenal at ang {N}ominal sa {P}anulaan ni {A}llan {P}opa.
\newblock \emph{Daluyan: Journal ng Wikang Filipino}, 28(2).

\bibitem[{Bowen(1971)}]{bowen1971hispanic}
J~Donald Bowen. 1971.
\newblock Hispanic languages and influence in {O}ceania.
\newblock \emph{Current trends in linguistics}, 8:938--52.

\bibitem[{Cabasag et~al.(2019)Cabasag, Chan, Lim, Gonzales, and Cheng}]{cabasag2019hate}
Neil~Vicente Cabasag, Vicente~Raphael Chan, Sean~Christian Lim, Mark~Edward Gonzales, and Charibeth Cheng. 2019.
\newblock Hate speech in {P}hilippine election-related tweets: {A}utomatic detection and classification using natural language processing.
\newblock \emph{Philippine Computing Journal, XIV}, 1.

\bibitem[{Cajote et~al.(2024)Cajote, Guevara, Bayona, and Lucas}]{cajote2024philippine}
Rhandley~D Cajote, Rowena Cristina~L Guevara, Michael Gringo Angelo~R Bayona, and Crisron Rudolf~G Lucas. 2024.
\newblock {P}hilippine {L}anguages {D}atabase: {A} {M}ultilingual {S}peech {C}orpora for {D}eveloping {S}ystems for {P}hilippine {S}poken {L}anguages.
\newblock \emph{LREC-COLING 2024}, page 264.

\bibitem[{Caroro et~al.(2020)Caroro, Paredes, and Lumasag}]{caroro2020rules}
Roseclaremath~A Caroro, Rolysent~K Paredes, and Jerry~M Lumasag. 2020.
\newblock Rules for orthographic word parsing of the {P}hilippines' {C}ebuano-{V}isayan language using context-free grammars.
\newblock \emph{International Journal of Software Science and Computational Intelligence (IJSSCI)}, 12(2):34--49.

\bibitem[{Castagnoli(2020)}]{castagnoli2020translation}
Sara Castagnoli. 2020.
\newblock Translation choices compared: Investigating variation in a learner translation corpus.
\newblock \emph{Translating and Comparing Languages: Corpus-based Insights. Corpora and Language in Use Proceedings}, 6:25--44.

\bibitem[{Chan-Yap(1980)}]{chan1980hokkien}
Gloria Chan-Yap. 1980.
\newblock \emph{Hokkien {C}hinese borrowings in {T}agalog}.
\newblock Dept. of Linguistics, Research School of Pacific Studies, Australian National University.

\bibitem[{Cohen(1960)}]{cohen1960coefficient}
Jacob Cohen. 1960.
\newblock A coefficient of agreement for nominal scales.
\newblock \emph{Educational and psychological measurement}, 20(1):37--46.

\bibitem[{Conneau et~al.(2018)Conneau, Rinott, Lample, Williams, Bowman, Schwenk, and Stoyanov}]{conneau2018xnli}
Alexis Conneau, Ruty Rinott, Guillaume Lample, Adina Williams, Samuel~R. Bowman, Holger Schwenk, and Veselin Stoyanov. 2018.
\newblock Xnli: {E}valuating {C}ross-lingual {S}entence {R}epresentations.
\newblock In \emph{Proceedings of the 2018 Conference on Empirical Methods in Natural Language Processing}. Association for Computational Linguistics.

\bibitem[{Cruz et~al.(2021)Cruz, Resabal, Lin, Velasco, and Cheng}]{cruz2021entailment}
Jan Christian~Blaise Cruz, Jose~Kristian Resabal, James Lin, Dan~John Velasco, and Charibeth Cheng. 2021.
\newblock Exploiting news article structure for automatic corpus generation of entailment datasets.
\newblock In \emph{PRICAI 2021: Trends in Artificial Intelligence}, pages 86--99, Cham. Springer International Publishing.

\bibitem[{Cruz et~al.(2020)Cruz, Tan, and Cheng}]{cruz-etal-2020-localization}
Jan Christian~Blaise Cruz, Julianne~Agatha Tan, and Charibeth Cheng. 2020.
\newblock \href {https://aclanthology.org/2020.lrec-1.316/} {Localization of fake news detection via multitask transfer learning}.
\newblock In \emph{Proceedings of the Twelfth Language Resources and Evaluation Conference}, pages 2596--2604, Marseille, France. European Language Resources Association.

\bibitem[{Dang et~al.(2024)Dang, Singh, D'souza, Ahmadian, Salamanca, Smith, Peppin, Hong, Govindassamy, Zhao, Kublik, Amer, Aryabumi, Campos, Tan, Kocmi, Strub, Grinsztajn, Flet-Berliac, Locatelli, Lin, Talupuru, Venkitesh, Cairuz, Yang, Chung, Ko, Shi, Shukayev, Bae, Piktus, Castagné, Cruz-Salinas, Kim, Crawhall-Stein, Morisot, Roy, Blunsom, Zhang, Gomez, Frosst, Fadaee, Ermis, Üstün, and Hooker}]{dang2024ayaexpansecombiningresearch}
John Dang, Shivalika Singh, Daniel D'souza, Arash Ahmadian, Alejandro Salamanca, Madeline Smith, Aidan Peppin, Sungjin Hong, Manoj Govindassamy, Terrence Zhao, Sandra Kublik, Meor Amer, Viraat Aryabumi, Jon~Ander Campos, Yi-Chern Tan, Tom Kocmi, Florian Strub, Nathan Grinsztajn, Yannis Flet-Berliac, Acyr Locatelli, Hangyu Lin, Dwarak Talupuru, Bharat Venkitesh, David Cairuz, Bowen Yang, Tim Chung, Wei-Yin Ko, Sylvie~Shang Shi, Amir Shukayev, Sammie Bae, Aleksandra Piktus, Roman Castagné, Felipe Cruz-Salinas, Eddie Kim, Lucas Crawhall-Stein, Adrien Morisot, Sudip Roy, Phil Blunsom, Ivan Zhang, Aidan Gomez, Nick Frosst, Marzieh Fadaee, Beyza Ermis, Ahmet Üstün, and Sara Hooker. 2024.
\newblock \href {https://arxiv.org/abs/2412.04261} {Aya expanse: Combining research breakthroughs for a new multilingual frontier}.
\newblock \emph{Preprint}, arXiv:2412.04261.

\bibitem[{DeepSeek-AI(2025)}]{deepseekai2025deepseekr1incentivizingreasoningcapability}
DeepSeek-AI. 2025.
\newblock \href {https://arxiv.org/abs/2501.12948} {Deepseek-r1: Incentivizing reasoning capability in llms via reinforcement learning}.
\newblock \emph{Preprint}, arXiv:2501.12948.

\bibitem[{Delgado and Tibau(2019)}]{delgado2019cohen}
Rosario Delgado and Xavier-Andoni Tibau. 2019.
\newblock Why {C}ohen’s {K}appa should be avoided as performance measure in classification.
\newblock \emph{PloS one}, 14(9):e0222916.

\bibitem[{Dita et~al.(2009)Dita, Roxas, and Inventado}]{dita2009building}
Shirley~N Dita, Rachel Edita~O Roxas, and Paul Inventado. 2009.
\newblock Building online corpora of {P}hilippine languages.
\newblock In \emph{Proceedings of the 23rd Pacific Asia Conference on Language, Information and Computation}, pages 646--653. Waseda University.

\bibitem[{Doddapaneni et~al.(2023)Doddapaneni, Aralikatte, Ramesh, Goyal, Khapra, Kunchukuttan, and Kumar}]{doddapaneni-etal-2023-towards}
Sumanth Doddapaneni, Rahul Aralikatte, Gowtham Ramesh, Shreya Goyal, Mitesh~M. Khapra, Anoop Kunchukuttan, and Pratyush Kumar. 2023.
\newblock \href {https://doi.org/10.18653/v1/2023.acl-long.693} {{T}owards {L}eaving {N}o {I}ndic {L}anguage {B}ehind: {B}uilding {M}onolingual {C}orpora, {B}enchmark and {M}odels for {I}ndic {L}anguages}.
\newblock In \emph{Proceedings of the 61st Annual Meeting of the Association for Computational Linguistics (Volume 1: Long Papers)}, pages 12402--12426, Toronto, Canada. Association for Computational Linguistics.

\bibitem[{Dou et~al.(2025)Dou, Liu, Zhou, Chen, Wang, Jin, Liu, Zhu, Du, Yang, Wang, Liu, Zhao, Feng, Mao, Yeung, Pipatanakul, Koto, Thu, Kydlíček, Liu, Lin, Sripaisarnmongkol, Sae-Khow, Thongchim, Konkaew, Borijindargoon, Dao, Maneegard, Artkaew, Yong, Nguyen, Phatthiyaphaibun, Tran, Zhang, Chen, Pang, Du, Wan, Lu, and Lin}]{dou2025sailor2sailingsoutheastasia}
Longxu Dou, Qian Liu, Fan Zhou, Changyu Chen, Zili Wang, Ziqi Jin, Zichen Liu, Tongyao Zhu, Cunxiao Du, Penghui Yang, Haonan Wang, Jiaheng Liu, Yongchi Zhao, Xiachong Feng, Xin Mao, Man~Tsung Yeung, Kunat Pipatanakul, Fajri Koto, Min~Si Thu, Hynek Kydlíček, Zeyi Liu, Qunshu Lin, Sittipong Sripaisarnmongkol, Kridtaphad Sae-Khow, Nirattisai Thongchim, Taechawat Konkaew, Narong Borijindargoon, Anh Dao, Matichon Maneegard, Phakphum Artkaew, Zheng-Xin Yong, Quan Nguyen, Wannaphong Phatthiyaphaibun, Hoang~H. Tran, Mike Zhang, Shiqi Chen, Tianyu Pang, Chao Du, Xinyi Wan, Wei Lu, and Min Lin. 2025.
\newblock \href {https://arxiv.org/abs/2502.12982} {Sailor2: Sailing in south-east asia with inclusive multilingual llms}.
\newblock \emph{Preprint}, arXiv:2502.12982.

\bibitem[{Feng and Zhao(2016)}]{charles2016not}
Guangchao~Charles Feng and Xinshu Zhao. 2016.
\newblock Do not force agreement: A response to {K}rippendorff (2016).
\newblock \emph{Methodology}.

\bibitem[{Fox(1985)}]{fox1985word}
Barbara~A Fox. 1985.
\newblock Word-order inversion and discourse continuity in {T}agalog.
\newblock \emph{Text-Interdisciplinary Journal for the Study of Discourse}, 5(1-2):39--54.

\bibitem[{Gellerstam(1986)}]{gellerstam1986translationese}
Martin Gellerstam. 1986.
\newblock Translationese in {Swedish} {Novels} {Translated} from {English}.
\newblock \emph{Translation studies in Scandinavia}, 1:88--95.

\bibitem[{Go and Nocon(2017)}]{go2017using}
Matthew Phillip~V Go and Nicco Nocon. 2017.
\newblock Using {S}tanford part-of-speech tagger for the morphologically-rich {F}ilipino language.
\newblock In \emph{Proceedings of the 31st Pacific Asia Conference on Language, Information and Computation}, pages 81--88. Waseda University.

\bibitem[{Go and Gustilo(2013)}]{go2013tagalog}
Mikhail~Alic Go and Leah Gustilo. 2013.
\newblock {T}agalog or {T}aglish: {T}he lingua franca of {F}ilipino urban factory workers.
\newblock \emph{Philippine ESL Journal}, 10:57--87.

\bibitem[{Gonzales(2022)}]{gonzales2022interactions}
Wilkinson Daniel~Wong Gonzales. 2022.
\newblock Interactions of {S}initic languages in the {P}hilippines: {S}inicization, {F}ilipinization, and {S}ino-{P}hilippine language creation.
\newblock In \emph{The Palgrave handbook of Chinese language studies}, pages 369--408. Springer.

\bibitem[{Goyal et~al.(2022)Goyal, Gao, Chaudhary, Chen, Wenzek, Ju, Krishnan, Ranzato, Guzm{\'a}n, and Fan}]{goyal-etal-2022-flores}
Naman Goyal, Cynthia Gao, Vishrav Chaudhary, Peng-Jen Chen, Guillaume Wenzek, Da~Ju, Sanjana Krishnan, Marc{'}Aurelio Ranzato, Francisco Guzm{\'a}n, and Angela Fan. 2022.
\newblock \href {https://doi.org/10.1162/tacl_a_00474} {The {F}lores-101 evaluation benchmark for low-resource and multilingual machine translation}.
\newblock \emph{Transactions of the Association for Computational Linguistics}, 10:522--538.

\bibitem[{Grattafiori et~al.(2024)Grattafiori, Dubey, Jauhri, Pandey, Kadian, Al-Dahle, Letman, Mathur, Schelten, Vaughan, Yang, Fan, Goyal, Hartshorn, Yang, Mitra, Sravankumar, Korenev, Hinsvark, Rao, Zhang, Rodriguez, Gregerson, Spataru, Roziere, Biron, Tang, Chern, Caucheteux, Nayak, Bi, Marra, McConnell, Keller, Touret, Wu, Wong, Ferrer, Nikolaidis, Allonsius, Song, Pintz, Livshits, Wyatt, Esiobu, Choudhary, Mahajan, Garcia-Olano, Perino, Hupkes, Lakomkin, AlBadawy, Lobanova, Dinan, Smith, Radenovic, Guzmán, Zhang, Synnaeve, Lee, Anderson, Thattai, Nail, Mialon, Pang, Cucurell, Nguyen, Korevaar, Xu, Touvron, Zarov, Ibarra, Kloumann, Misra, Evtimov, Zhang, Copet, Lee, Geffert, Vranes, Park, Mahadeokar, Shah, van~der Linde, Billock, Hong, Lee, Fu, Chi, Huang, Liu, Wang, Yu, Bitton, Spisak, Park, Rocca, Johnstun, Saxe, Jia, Alwala, Prasad, Upasani, Plawiak, Li, Heafield, Stone, El-Arini, Iyer, Malik, Chiu, Bhalla, Lakhotia, Rantala-Yeary, van~der Maaten, Chen, Tan, Jenkins, Martin, Madaan, Malo, Blecher,
  Landzaat, de~Oliveira, Muzzi, Pasupuleti, Singh, Paluri, Kardas, Tsimpoukelli, Oldham, Rita, Pavlova, Kambadur, Lewis, Si, Singh, Hassan, Goyal, Torabi, Bashlykov, Bogoychev, Chatterji, Zhang, Duchenne, Çelebi, Alrassy, Zhang, Li, Vasic, Weng, Bhargava, Dubal, Krishnan, Koura, Xu, He, Dong, Srinivasan, Ganapathy, Calderer, Cabral, Stojnic, Raileanu, Maheswari, Girdhar, Patel, Sauvestre, Polidoro, Sumbaly, Taylor, Silva, Hou, Wang, Hosseini, Chennabasappa, Singh, Bell, Kim, Edunov, Nie, Narang, Raparthy, Shen, Wan, Bhosale, Zhang, Vandenhende, Batra, Whitman, Sootla, Collot, Gururangan, Borodinsky, Herman, Fowler, Sheasha, Georgiou, Scialom, Speckbacher, Mihaylov, Xiao, Karn, Goswami, Gupta, Ramanathan, Kerkez, Gonguet, Do, Vogeti, Albiero, Petrovic, Chu, Xiong, Fu, Meers, Martinet, Wang, Wang, Tan, Xia, Xie, Jia, Wang, Goldschlag, Gaur, Babaei, Wen, Song, Zhang, Li, Mao, Coudert, Yan, Chen, Papakipos, Singh, Srivastava, Jain, Kelsey, Shajnfeld, Gangidi, Victoria, Goldstand, Menon, Sharma, Boesenberg,
  Baevski, Feinstein, Kallet, Sangani, Teo, Yunus, Lupu, Alvarado, Caples, Gu, Ho, Poulton, Ryan, Ramchandani, Dong, Franco, Goyal, Saraf, Chowdhury, Gabriel, Bharambe, Eisenman, Yazdan, James, Maurer, Leonhardi, Huang, Loyd, Paola, Paranjape, Liu, Wu, Ni, Hancock, Wasti, Spence, Stojkovic, Gamido, Montalvo, Parker, Burton, Mejia, Liu, Wang, Kim, Zhou, Hu, Chu, Cai, Tindal, Feichtenhofer, Gao, Civin, Beaty, Kreymer, Li, Adkins, Xu, Testuggine, David, Parikh, Liskovich, Foss, Wang, Le, Holland, Dowling, Jamil, Montgomery, Presani, Hahn, Wood, Le, Brinkman, Arcaute, Dunbar, Smothers, Sun, Kreuk, Tian, Kokkinos, Ozgenel, Caggioni, Kanayet, Seide, Florez, Schwarz, Badeer, Swee, Halpern, Herman, Sizov, Guangyi, Zhang, Lakshminarayanan, Inan, Shojanazeri, Zou, Wang, Zha, Habeeb, Rudolph, Suk, Aspegren, Goldman, Zhan, Damlaj, Molybog, Tufanov, Leontiadis, Veliche, Gat, Weissman, Geboski, Kohli, Lam, Asher, Gaya, Marcus, Tang, Chan, Zhen, Reizenstein, Teboul, Zhong, Jin, Yang, Cummings, Carvill, Shepard, McPhie,
  Torres, Ginsburg, Wang, Wu, U, Saxena, Khandelwal, Zand, Matosich, Veeraraghavan, Michelena, Li, Jagadeesh, Huang, Chawla, Huang, Chen, Garg, A, Silva, Bell, Zhang, Guo, Yu, Moshkovich, Wehrstedt, Khabsa, Avalani, Bhatt, Mankus, Hasson, Lennie, Reso, Groshev, Naumov, Lathi, Keneally, Liu, Seltzer, Valko, Restrepo, Patel, Vyatskov, Samvelyan, Clark, Macey, Wang, Hermoso, Metanat, Rastegari, Bansal, Santhanam, Parks, White, Bawa, Singhal, Egebo, Usunier, Mehta, Laptev, Dong, Cheng, Chernoguz, Hart, Salpekar, Kalinli, Kent, Parekh, Saab, Balaji, Rittner, Bontrager, Roux, Dollar, Zvyagina, Ratanchandani, Yuvraj, Liang, Alao, Rodriguez, Ayub, Murthy, Nayani, Mitra, Parthasarathy, Li, Hogan, Battey, Wang, Howes, Rinott, Mehta, Siby, Bondu, Datta, Chugh, Hunt, Dhillon, Sidorov, Pan, Mahajan, Verma, Yamamoto, Ramaswamy, Lindsay, Lindsay, Feng, Lin, Zha, Patil, Shankar, Zhang, Zhang, Wang, Agarwal, Sajuyigbe, Chintala, Max, Chen, Kehoe, Satterfield, Govindaprasad, Gupta, Deng, Cho, Virk, Subramanian, Choudhury,
  Goldman, Remez, Glaser, Best, Koehler, Robinson, Li, Zhang, Matthews, Chou, Shaked, Vontimitta, Ajayi, Montanez, Mohan, Kumar, Mangla, Ionescu, Poenaru, Mihailescu, Ivanov, Li, Wang, Jiang, Bouaziz, Constable, Tang, Wu, Wang, Wu, Gao, Kleinman, Chen, Hu, Jia, Qi, Li, Zhang, Zhang, Adi, Nam, Yu, Wang, Zhao, Hao, Qian, Li, He, Rait, DeVito, Rosnbrick, Wen, Yang, Zhao, and Ma}]{grattafiori2024llama3herdmodels}
Aaron Grattafiori, Abhimanyu Dubey, Abhinav Jauhri, Abhinav Pandey, Abhishek Kadian, Ahmad Al-Dahle, Aiesha Letman, Akhil Mathur, Alan Schelten, Alex Vaughan, Amy Yang, Angela Fan, Anirudh Goyal, Anthony Hartshorn, Aobo Yang, Archi Mitra, Archie Sravankumar, Artem Korenev, Arthur Hinsvark, Arun Rao, Aston Zhang, Aurelien Rodriguez, Austen Gregerson, Ava Spataru, Baptiste Roziere, Bethany Biron, Binh Tang, Bobbie Chern, Charlotte Caucheteux, Chaya Nayak, Chloe Bi, Chris Marra, Chris McConnell, Christian Keller, Christophe Touret, Chunyang Wu, Corinne Wong, Cristian~Canton Ferrer, Cyrus Nikolaidis, Damien Allonsius, Daniel Song, Danielle Pintz, Danny Livshits, Danny Wyatt, David Esiobu, Dhruv Choudhary, Dhruv Mahajan, Diego Garcia-Olano, Diego Perino, Dieuwke Hupkes, Egor Lakomkin, Ehab AlBadawy, Elina Lobanova, Emily Dinan, Eric~Michael Smith, Filip Radenovic, Francisco Guzmán, Frank Zhang, Gabriel Synnaeve, Gabrielle Lee, Georgia~Lewis Anderson, Govind Thattai, Graeme Nail, Gregoire Mialon, Guan Pang,
  Guillem Cucurell, Hailey Nguyen, Hannah Korevaar, Hu~Xu, Hugo Touvron, Iliyan Zarov, Imanol~Arrieta Ibarra, Isabel Kloumann, Ishan Misra, Ivan Evtimov, Jack Zhang, Jade Copet, Jaewon Lee, Jan Geffert, Jana Vranes, Jason Park, Jay Mahadeokar, Jeet Shah, Jelmer van~der Linde, Jennifer Billock, Jenny Hong, Jenya Lee, Jeremy Fu, Jianfeng Chi, Jianyu Huang, Jiawen Liu, Jie Wang, Jiecao Yu, Joanna Bitton, Joe Spisak, Jongsoo Park, Joseph Rocca, Joshua Johnstun, Joshua Saxe, Junteng Jia, Kalyan~Vasuden Alwala, Karthik Prasad, Kartikeya Upasani, Kate Plawiak, Ke~Li, Kenneth Heafield, Kevin Stone, Khalid El-Arini, Krithika Iyer, Kshitiz Malik, Kuenley Chiu, Kunal Bhalla, Kushal Lakhotia, Lauren Rantala-Yeary, Laurens van~der Maaten, Lawrence Chen, Liang Tan, Liz Jenkins, Louis Martin, Lovish Madaan, Lubo Malo, Lukas Blecher, Lukas Landzaat, Luke de~Oliveira, Madeline Muzzi, Mahesh Pasupuleti, Mannat Singh, Manohar Paluri, Marcin Kardas, Maria Tsimpoukelli, Mathew Oldham, Mathieu Rita, Maya Pavlova, Melanie Kambadur,
  Mike Lewis, Min Si, Mitesh~Kumar Singh, Mona Hassan, Naman Goyal, Narjes Torabi, Nikolay Bashlykov, Nikolay Bogoychev, Niladri Chatterji, Ning Zhang, Olivier Duchenne, Onur Çelebi, Patrick Alrassy, Pengchuan Zhang, Pengwei Li, Petar Vasic, Peter Weng, Prajjwal Bhargava, Pratik Dubal, Praveen Krishnan, Punit~Singh Koura, Puxin Xu, Qing He, Qingxiao Dong, Ragavan Srinivasan, Raj Ganapathy, Ramon Calderer, Ricardo~Silveira Cabral, Robert Stojnic, Roberta Raileanu, Rohan Maheswari, Rohit Girdhar, Rohit Patel, Romain Sauvestre, Ronnie Polidoro, Roshan Sumbaly, Ross Taylor, Ruan Silva, Rui Hou, Rui Wang, Saghar Hosseini, Sahana Chennabasappa, Sanjay Singh, Sean Bell, Seohyun~Sonia Kim, Sergey Edunov, Shaoliang Nie, Sharan Narang, Sharath Raparthy, Sheng Shen, Shengye Wan, Shruti Bhosale, Shun Zhang, Simon Vandenhende, Soumya Batra, Spencer Whitman, Sten Sootla, Stephane Collot, Suchin Gururangan, Sydney Borodinsky, Tamar Herman, Tara Fowler, Tarek Sheasha, Thomas Georgiou, Thomas Scialom, Tobias Speckbacher,
  Todor Mihaylov, Tong Xiao, Ujjwal Karn, Vedanuj Goswami, Vibhor Gupta, Vignesh Ramanathan, Viktor Kerkez, Vincent Gonguet, Virginie Do, Vish Vogeti, Vítor Albiero, Vladan Petrovic, Weiwei Chu, Wenhan Xiong, Wenyin Fu, Whitney Meers, Xavier Martinet, Xiaodong Wang, Xiaofang Wang, Xiaoqing~Ellen Tan, Xide Xia, Xinfeng Xie, Xuchao Jia, Xuewei Wang, Yaelle Goldschlag, Yashesh Gaur, Yasmine Babaei, Yi~Wen, Yiwen Song, Yuchen Zhang, Yue Li, Yuning Mao, Zacharie~Delpierre Coudert, Zheng Yan, Zhengxing Chen, Zoe Papakipos, Aaditya Singh, Aayushi Srivastava, Abha Jain, Adam Kelsey, Adam Shajnfeld, Adithya Gangidi, Adolfo Victoria, Ahuva Goldstand, Ajay Menon, Ajay Sharma, Alex Boesenberg, Alexei Baevski, Allie Feinstein, Amanda Kallet, Amit Sangani, Amos Teo, Anam Yunus, Andrei Lupu, Andres Alvarado, Andrew Caples, Andrew Gu, Andrew Ho, Andrew Poulton, Andrew Ryan, Ankit Ramchandani, Annie Dong, Annie Franco, Anuj Goyal, Aparajita Saraf, Arkabandhu Chowdhury, Ashley Gabriel, Ashwin Bharambe, Assaf Eisenman, Azadeh
  Yazdan, Beau James, Ben Maurer, Benjamin Leonhardi, Bernie Huang, Beth Loyd, Beto~De Paola, Bhargavi Paranjape, Bing Liu, Bo~Wu, Boyu Ni, Braden Hancock, Bram Wasti, Brandon Spence, Brani Stojkovic, Brian Gamido, Britt Montalvo, Carl Parker, Carly Burton, Catalina Mejia, Ce~Liu, Changhan Wang, Changkyu Kim, Chao Zhou, Chester Hu, Ching-Hsiang Chu, Chris Cai, Chris Tindal, Christoph Feichtenhofer, Cynthia Gao, Damon Civin, Dana Beaty, Daniel Kreymer, Daniel Li, David Adkins, David Xu, Davide Testuggine, Delia David, Devi Parikh, Diana Liskovich, Didem Foss, Dingkang Wang, Duc Le, Dustin Holland, Edward Dowling, Eissa Jamil, Elaine Montgomery, Eleonora Presani, Emily Hahn, Emily Wood, Eric-Tuan Le, Erik Brinkman, Esteban Arcaute, Evan Dunbar, Evan Smothers, Fei Sun, Felix Kreuk, Feng Tian, Filippos Kokkinos, Firat Ozgenel, Francesco Caggioni, Frank Kanayet, Frank Seide, Gabriela~Medina Florez, Gabriella Schwarz, Gada Badeer, Georgia Swee, Gil Halpern, Grant Herman, Grigory Sizov, Guangyi, Zhang, Guna
  Lakshminarayanan, Hakan Inan, Hamid Shojanazeri, Han Zou, Hannah Wang, Hanwen Zha, Haroun Habeeb, Harrison Rudolph, Helen Suk, Henry Aspegren, Hunter Goldman, Hongyuan Zhan, Ibrahim Damlaj, Igor Molybog, Igor Tufanov, Ilias Leontiadis, Irina-Elena Veliche, Itai Gat, Jake Weissman, James Geboski, James Kohli, Janice Lam, Japhet Asher, Jean-Baptiste Gaya, Jeff Marcus, Jeff Tang, Jennifer Chan, Jenny Zhen, Jeremy Reizenstein, Jeremy Teboul, Jessica Zhong, Jian Jin, Jingyi Yang, Joe Cummings, Jon Carvill, Jon Shepard, Jonathan McPhie, Jonathan Torres, Josh Ginsburg, Junjie Wang, Kai Wu, Kam~Hou U, Karan Saxena, Kartikay Khandelwal, Katayoun Zand, Kathy Matosich, Kaushik Veeraraghavan, Kelly Michelena, Keqian Li, Kiran Jagadeesh, Kun Huang, Kunal Chawla, Kyle Huang, Lailin Chen, Lakshya Garg, Lavender A, Leandro Silva, Lee Bell, Lei Zhang, Liangpeng Guo, Licheng Yu, Liron Moshkovich, Luca Wehrstedt, Madian Khabsa, Manav Avalani, Manish Bhatt, Martynas Mankus, Matan Hasson, Matthew Lennie, Matthias Reso, Maxim
  Groshev, Maxim Naumov, Maya Lathi, Meghan Keneally, Miao Liu, Michael~L. Seltzer, Michal Valko, Michelle Restrepo, Mihir Patel, Mik Vyatskov, Mikayel Samvelyan, Mike Clark, Mike Macey, Mike Wang, Miquel~Jubert Hermoso, Mo~Metanat, Mohammad Rastegari, Munish Bansal, Nandhini Santhanam, Natascha Parks, Natasha White, Navyata Bawa, Nayan Singhal, Nick Egebo, Nicolas Usunier, Nikhil Mehta, Nikolay~Pavlovich Laptev, Ning Dong, Norman Cheng, Oleg Chernoguz, Olivia Hart, Omkar Salpekar, Ozlem Kalinli, Parkin Kent, Parth Parekh, Paul Saab, Pavan Balaji, Pedro Rittner, Philip Bontrager, Pierre Roux, Piotr Dollar, Polina Zvyagina, Prashant Ratanchandani, Pritish Yuvraj, Qian Liang, Rachad Alao, Rachel Rodriguez, Rafi Ayub, Raghotham Murthy, Raghu Nayani, Rahul Mitra, Rangaprabhu Parthasarathy, Raymond Li, Rebekkah Hogan, Robin Battey, Rocky Wang, Russ Howes, Ruty Rinott, Sachin Mehta, Sachin Siby, Sai~Jayesh Bondu, Samyak Datta, Sara Chugh, Sara Hunt, Sargun Dhillon, Sasha Sidorov, Satadru Pan, Saurabh Mahajan,
  Saurabh Verma, Seiji Yamamoto, Sharadh Ramaswamy, Shaun Lindsay, Shaun Lindsay, Sheng Feng, Shenghao Lin, Shengxin~Cindy Zha, Shishir Patil, Shiva Shankar, Shuqiang Zhang, Shuqiang Zhang, Sinong Wang, Sneha Agarwal, Soji Sajuyigbe, Soumith Chintala, Stephanie Max, Stephen Chen, Steve Kehoe, Steve Satterfield, Sudarshan Govindaprasad, Sumit Gupta, Summer Deng, Sungmin Cho, Sunny Virk, Suraj Subramanian, Sy~Choudhury, Sydney Goldman, Tal Remez, Tamar Glaser, Tamara Best, Thilo Koehler, Thomas Robinson, Tianhe Li, Tianjun Zhang, Tim Matthews, Timothy Chou, Tzook Shaked, Varun Vontimitta, Victoria Ajayi, Victoria Montanez, Vijai Mohan, Vinay~Satish Kumar, Vishal Mangla, Vlad Ionescu, Vlad Poenaru, Vlad~Tiberiu Mihailescu, Vladimir Ivanov, Wei Li, Wenchen Wang, Wenwen Jiang, Wes Bouaziz, Will Constable, Xiaocheng Tang, Xiaojian Wu, Xiaolan Wang, Xilun Wu, Xinbo Gao, Yaniv Kleinman, Yanjun Chen, Ye~Hu, Ye~Jia, Ye~Qi, Yenda Li, Yilin Zhang, Ying Zhang, Yossi Adi, Youngjin Nam, Yu, Wang, Yu~Zhao, Yuchen Hao, Yundi
  Qian, Yunlu Li, Yuzi He, Zach Rait, Zachary DeVito, Zef Rosnbrick, Zhaoduo Wen, Zhenyu Yang, Zhiwei Zhao, and Zhiyu Ma. 2024.
\newblock \href {https://arxiv.org/abs/2407.21783} {The llama 3 herd of models}.
\newblock \emph{Preprint}, arXiv:2407.21783.

\bibitem[{Guo et~al.(2024)Guo, Dou, Zeng, Kok, Lu, and Liu}]{guo2024sailcompass}
Jia Guo, Longxu Dou, Guangtao Zeng, Stanley Kok, Wei Lu, and Qian Liu. 2024.
\newblock Sailcompass: Towards reproducible and robust evaluation for southeast asian languages.
\newblock \emph{arXiv preprint arXiv:2412.01186}.

\bibitem[{Guo et~al.(2022)Guo, Clavel, Kamal~Eddine, and Vazirgiannis}]{guo-etal-2022-questioning}
Yanzhu Guo, Chlo{\'e} Clavel, Moussa Kamal~Eddine, and Michalis Vazirgiannis. 2022.
\newblock \href {https://doi.org/10.18653/v1/2022.emnlp-main.386} {{Q}uestioning the {V}alidity of {S}ummarization {D}atasets and {I}mproving {T}heir {F}actual {C}onsistency}.
\newblock In \emph{Proceedings of the 2022 Conference on Empirical Methods in Natural Language Processing}, pages 5716--5727, Abu Dhabi, United Arab Emirates. Association for Computational Linguistics.

\bibitem[{Guo et~al.(2023)Guo, Jin, Liu, Huang, Shi, Yu, Liu, Li, Xiong, Xiong et~al.}]{guo2023evaluating}
Zishan Guo, Renren Jin, Chuang Liu, Yufei Huang, Dan Shi, Linhao Yu, Yan Liu, Jiaxuan Li, Bojian Xiong, Deyi Xiong, et~al. 2023.
\newblock Evaluating large language models: A comprehensive survey.
\newblock \emph{arXiv preprint arXiv:2310.19736}.

\bibitem[{Hadi et~al.(2023)Hadi, Qureshi, Shah, Irfan, Zafar, Shaikh, Akhtar, Wu, Mirjalili et~al.}]{hadi2023survey}
Muhammad~Usman Hadi, Rizwan Qureshi, Abbas Shah, Muhammad Irfan, Anas Zafar, Muhammad~Bilal Shaikh, Naveed Akhtar, Jia Wu, Seyedali Mirjalili, et~al. 2023.
\newblock A survey on large language models: {A}pplications, challenges, limitations, and practical usage.
\newblock \emph{Authorea Preprints}.

\bibitem[{Hasan et~al.(2021)Hasan, Bhattacharjee, Islam, Mubasshir, Li, Kang, Rahman, and Shahriyar}]{hasan-etal-2021-xl}
Tahmid Hasan, Abhik Bhattacharjee, Md.~Saiful Islam, Kazi Mubasshir, Yuan-Fang Li, Yong-Bin Kang, M.~Sohel Rahman, and Rifat Shahriyar. 2021.
\newblock \href {https://doi.org/10.18653/v1/2021.findings-acl.413} {{XL}-{S}um: {L}arge-scale {M}ultilingual {A}bstractive {S}ummarization for 44 {L}anguages}.
\newblock In \emph{Findings of the Association for Computational Linguistics: ACL-IJCNLP 2021}, pages 4693--4703, Online. Association for Computational Linguistics.

\bibitem[{Hu et~al.(2020)Hu, Ruder, Siddhant, Neubig, Firat, and Johnson}]{10.5555/3524938.3525348}
Junjie Hu, Sebastian Ruder, Aditya Siddhant, Graham Neubig, Orhan Firat, and Melvin Johnson. 2020.
\newblock Xtreme: a massively multilingual multi-task benchmark for evaluating cross-lingual generalization.
\newblock In \emph{Proceedings of the 37th International Conference on Machine Learning}, ICML'20. JMLR.org.

\bibitem[{Huang et~al.(2025)Huang, Vangani, Pham, Zou, Wang, Liu, and Aw}]{huang2025meraliontextllmcrosslingualunderstandinglarge}
Xin Huang, Tarun~Kumar Vangani, Minh~Duc Pham, Xunlong Zou, Bin Wang, Zhengyuan Liu, and Ai~Ti Aw. 2025.
\newblock \href {https://arxiv.org/abs/2501.08335} {Meralion-textllm: Cross-lingual understanding of large language models in chinese, indonesian, malay, and singlish}.
\newblock \emph{Preprint}, arXiv:2501.08335.

\bibitem[{Ilao et~al.(2012)Ilao, Santos, and Guevara}]{ilao2012comparative}
Joel~P Ilao, Timothy Israel~D Santos, and Rowena Cristina~L Guevara. 2012.
\newblock Comparative analysis of actual language usage and selected grammar and orthographical rules for {F}ilipino, {C}ebuano-{V}isayan and {I}lokano: a {C}orpus-based {A}pproach.
\newblock \emph{Electrical and Electronics Engineering Institute. University of the Philippines Diliman}.

\bibitem[{Javier(2018)}]{javier2018pagsusuri}
Jem~R Javier. 2018.
\newblock Pagsusuri sa ortograpiya ng kambal-katinig sa {F}ilipino batay sa korpus ({A} corpus-based analysis of consonant clusters in {F}ilipino orthography).
\newblock \emph{Social Science Diliman}, 14(1).

\bibitem[{Jubilado(2004)}]{jubilado2004philippine}
Rodney~C Jubilado. 2004.
\newblock Philippine linguistics, {F}ilipino language, and the {F}ilipino nation.
\newblock \emph{JATI-Journal of Southeast Asian Studies}, 9(1):43--53.

\bibitem[{Juraska et~al.(2024)Juraska, Deutsch, Finkelstein, and Freitag}]{juraska-etal-2024-metricx}
Juraj Juraska, Daniel Deutsch, Mara Finkelstein, and Markus Freitag. 2024.
\newblock \href {https://doi.org/10.18653/v1/2024.wmt-1.35} {{M}etric{X}-24: The {G}oogle submission to the {WMT} 2024 metrics shared task}.
\newblock In \emph{Proceedings of the Ninth Conference on Machine Translation}, pages 492--504, Miami, Florida, USA. Association for Computational Linguistics.

\bibitem[{Kavumba et~al.(2019)Kavumba, Inoue, Heinzerling, Singh, Reisert, and Inui}]{kavumba-etal-2019-choosing}
Pride Kavumba, Naoya Inoue, Benjamin Heinzerling, Keshav Singh, Paul Reisert, and Kentaro Inui. 2019.
\newblock \href {https://doi.org/10.18653/v1/D19-6004} {When {C}hoosing {P}lausible {A}lternatives, {C}lever {H}ans can be {C}lever}.
\newblock In \emph{Proceedings of the First Workshop on Commonsense Inference in Natural Language Processing}, pages 33--42, Hong Kong, China. Association for Computational Linguistics.

\bibitem[{Krippendorff(2018)}]{krippendorff2018content}
Klaus Krippendorff. 2018.
\newblock \emph{Content analysis: An introduction to its methodology}.
\newblock Sage publications.

\bibitem[{Kroeger(1993)}]{kroeger1993phrase}
Paul Kroeger. 1993.
\newblock \emph{Phrase structure and grammatical relations in {T}agalog}.
\newblock Center for the Study of Language (CSLI).

\bibitem[{Leong et~al.(2023)Leong, Ngui, Susanto, Rengarajan, Sarveswaran, and Tjhi}]{leong2023bhasaholisticsoutheastasian}
Wei~Qi Leong, Jian~Gang Ngui, Yosephine Susanto, Hamsawardhini Rengarajan, Kengatharaiyer Sarveswaran, and William~Chandra Tjhi. 2023.
\newblock \emph{Preprint}, arXiv:2309.06085.
\newblock \href {https://arxiv.org/abs/2309.06085} {[link]}.

\bibitem[{Liang et~al.(2023)Liang, Bommasani, Lee, Tsipras, Soylu, Yasunaga, Zhang, Narayanan, Wu, Kumar, Newman, Yuan, Yan, Zhang, Cosgrove, Manning, Ré, Acosta-Navas, Hudson, Zelikman, Durmus, Ladhak, Rong, Ren, Yao, Wang, Santhanam, Orr, Zheng, Yuksekgonul, Suzgun, Kim, Guha, Chatterji, Khattab, Henderson, Huang, Chi, Xie, Santurkar, Ganguli, Hashimoto, Icard, Zhang, Chaudhary, Wang, Li, Mai, Zhang, and Koreeda}]{liang2023holisticevaluationlanguagemodels}
Percy Liang, Rishi Bommasani, Tony Lee, Dimitris Tsipras, Dilara Soylu, Michihiro Yasunaga, Yian Zhang, Deepak Narayanan, Yuhuai Wu, Ananya Kumar, Benjamin Newman, Binhang Yuan, Bobby Yan, Ce~Zhang, Christian Cosgrove, Christopher~D. Manning, Christopher Ré, Diana Acosta-Navas, Drew~A. Hudson, Eric Zelikman, Esin Durmus, Faisal Ladhak, Frieda Rong, Hongyu Ren, Huaxiu Yao, Jue Wang, Keshav Santhanam, Laurel Orr, Lucia Zheng, Mert Yuksekgonul, Mirac Suzgun, Nathan Kim, Neel Guha, Niladri Chatterji, Omar Khattab, Peter Henderson, Qian Huang, Ryan Chi, Sang~Michael Xie, Shibani Santurkar, Surya Ganguli, Tatsunori Hashimoto, Thomas Icard, Tianyi Zhang, Vishrav Chaudhary, William Wang, Xuechen Li, Yifan Mai, Yuhui Zhang, and Yuta Koreeda. 2023.
\newblock \href {https://arxiv.org/abs/2211.09110} {Holistic {E}valuation of {L}anguage {M}odels}.
\newblock \emph{Preprint}, arXiv:2211.09110.

\bibitem[{Lin(2004)}]{lin-2004-rouge}
Chin-Yew Lin. 2004.
\newblock \href {https://aclanthology.org/W04-1013} {{ROUGE}: {A} {P}ackage for {A}utomatic {E}valuation of {S}ummaries}.
\newblock In \emph{Text Summarization Branches Out}, pages 74--81, Barcelona, Spain. Association for Computational Linguistics.

\bibitem[{Liu et~al.(2025)Liu, Zhang, Ying, Aljunied, Luu, and Bing}]{liu-etal-2025-seaexam}
Chaoqun Liu, Wenxuan Zhang, Jiahao Ying, Mahani Aljunied, Anh~Tuan Luu, and Lidong Bing. 2025.
\newblock \href {https://aclanthology.org/2025.findings-naacl.341/} {{S}ea{E}xam and {S}ea{B}ench: Benchmarking {LLM}s with local multilingual questions in {S}outheast {A}sia}.
\newblock In \emph{Findings of the Association for Computational Linguistics: NAACL 2025}, pages 6119--6136, Albuquerque, New Mexico. Association for Computational Linguistics.

\bibitem[{Liu et~al.(2021)Liu, Bugliarello, Ponti, Reddy, Collier, and Elliott}]{liu-etal-2021-visually}
Fangyu Liu, Emanuele Bugliarello, Edoardo~Maria Ponti, Siva Reddy, Nigel Collier, and Desmond Elliott. 2021.
\newblock \href {https://doi.org/10.18653/v1/2021.emnlp-main.818} {Visually {G}rounded {R}easoning across {L}anguages and {C}ultures}.
\newblock In \emph{Proceedings of the 2021 Conference on Empirical Methods in Natural Language Processing}, pages 10467--10485, Online and Punta Cana, Dominican Republic. Association for Computational Linguistics.

\bibitem[{Liu et~al.(2023)Liu, Yao, Ton, Zhang, Cheng, Klochkov, Taufiq, and Li}]{liu2023trustworthy}
Yang Liu, Yuanshun Yao, Jean-Francois Ton, Xiaoying Zhang, Ruocheng Guo~Hao Cheng, Yegor Klochkov, Muhammad~Faaiz Taufiq, and Hang Li. 2023.
\newblock Trustworthy {LLMs}: A survey and guideline for evaluating large language models' alignment.
\newblock \emph{arXiv preprint arXiv:2308.05374}.

\bibitem[{Lovenia et~al.(2024)Lovenia, Mahendra, Akbar, Miranda, Santoso, Aco, Fadhilah, Mansurov, Imperial, Kampman, Moniz, Habibi, Hudi, Montalan, Hadiwijaya, Lopo, Nixon, Karlsson, Jaya, Diandaru, Gao, Irawan, Wang, Cruz, Whitehouse, Parmonangan, Khelli, Zhang, Susanto, Ryanda, Hermawan, Velasco, Kautsar, Hendria, Moslem, Flynn, Adilazuarda, Li, Lee, Damanhuri, Sun, Qorib, Djanibekov, Leong, Do, Muennighoff, Pansuwan, Putra, Xu, Chia, Purwarianti, Ruder, Tjhi, Limkonchotiwat, Aji, Keh, Winata, Zhang, Koto, Yong, and Cahyawijaya}]{lovenia-etal-2024-seacrowd}
Holy Lovenia, Rahmad Mahendra, Salsabil~Maulana Akbar, Lester~Jamesalidad Miranda, Jennifer Santoso, Elyanah Aco, Akhdan Fadhilah, Jonibek Mansurov, Joseph~Marvin Imperial, Onno~P. Kampman, Joel Ruben~Antony Moniz, Muhammad Ravi~Shulthan Habibi, Frederikus Hudi, Jann~Railey Montalan, Ryan~Ignatius Hadiwijaya, Joanito~Agili Lopo, William Nixon, B{\"o}rje~F. Karlsson, James Jaya, Ryandito Diandaru, Yuze Gao, Patrick~Amadeus Irawan, Bin Wang, Jan Christian~Blaise Cruz, Chenxi Whitehouse, Ivan~Halim Parmonangan, Maria Khelli, Wenyu Zhang, Lucky Susanto, Reynard~Adha Ryanda, Sonny~Lazuardi Hermawan, Dan~John Velasco, Muhammad Dehan~Al Kautsar, Willy~Fitra Hendria, Yasmin Moslem, Noah Flynn, Muhammad~Farid Adilazuarda, Haochen Li, Johanes Lee, R.~Damanhuri, Shuo Sun, Muhammad~Reza Qorib, Amirbek Djanibekov, Wei~Qi Leong, Quyet~V. Do, Niklas Muennighoff, Tanrada Pansuwan, Ilham~Firdausi Putra, Yan Xu, Tai~Ngee Chia, Ayu Purwarianti, Sebastian Ruder, William~Chandra Tjhi, Peerat Limkonchotiwat, Alham~Fikri Aji,
  Sedrick Keh, Genta~Indra Winata, Ruochen Zhang, Fajri Koto, Zheng~Xin Yong, and Samuel Cahyawijaya. 2024.
\newblock \href {https://doi.org/10.18653/v1/2024.emnlp-main.296} {{SEAC}rowd: A multilingual multimodal data hub and benchmark suite for {S}outheast {A}sian languages}.
\newblock In \emph{Proceedings of the 2024 Conference on Empirical Methods in Natural Language Processing}, pages 5155--5203, Miami, Florida, USA. Association for Computational Linguistics.

\bibitem[{Magracia(2001)}]{magracia2001panumbas}
Emma~B Magracia. 2001.
\newblock Panumbas sa {F}ilipino ng {P}anghihiram: {P}agtinging {S}emantikal at {S}intaktikal.
\newblock \emph{Asia Pacific Journal of Social Innovation (formerly The Mindanao Forum)}, 16(2):31--42.

\bibitem[{Malicsi(2013)}]{malicsi2013gramar}
Jonathan~C. Malicsi. 2013.
\newblock \href {https://books.google.com.sg/books?id=iYO8rQEACAAJ} {\emph{Gramar ng {F}ilipino}}.
\newblock Aklat Sanyata. Sentro ng Wikang Filipino, Unibersidad ng Pilipinas.

\bibitem[{Miranda(2023{\natexlab{a}})}]{miranda2023developing}
Lester~James Miranda. 2023{\natexlab{a}}.
\newblock Developing a named entity recognition dataset for tagalog.
\newblock \emph{arXiv preprint arXiv:2311.07161}.

\bibitem[{Miranda(2023{\natexlab{b}})}]{miranda2023nlp}
Lester~James Miranda. 2023{\natexlab{b}}.
\newblock \href {https://ljvmiranda921.github.io/notebook/2023/02/04/tagalog-pipeline/} {Towards a {T}agalog {NLP} pipeline}.

\bibitem[{Ng et~al.(2025)Ng, Nguyen, Huang, Tai, Leong, Leong, Yong, Ngui, Susanto, Cheng, Rengarajan, Limkonchotiwat, Hulagadri, Teng, Tong, Siow, Teo, Lau, Tan, Ong, Ong, Montalan, Chan, Antonyrex, Lee, Choa, Tat-Wee, Liu, Tjhi, Cambria, and Teo}]{ng2025sealionsoutheastasianlanguages}
Raymond Ng, Thanh~Ngan Nguyen, Yuli Huang, Ngee~Chia Tai, Wai~Yi Leong, Wei~Qi Leong, Xianbin Yong, Jian~Gang Ngui, Yosephine Susanto, Nicholas Cheng, Hamsawardhini Rengarajan, Peerat Limkonchotiwat, Adithya~Venkatadri Hulagadri, Kok~Wai Teng, Yeo~Yeow Tong, Bryan Siow, Wei~Yi Teo, Wayne Lau, Choon~Meng Tan, Brandon Ong, Zhi~Hao Ong, Jann~Railey Montalan, Adwin Chan, Sajeban Antonyrex, Ren Lee, Esther Choa, David~Ong Tat-Wee, Bing Jie~Darius Liu, William~Chandra Tjhi, Erik Cambria, and Leslie Teo. 2025.
\newblock \href {https://arxiv.org/abs/2504.05747} {Sea-lion: Southeast asian languages in one network}.
\newblock \emph{Preprint}, arXiv:2504.05747.

\bibitem[{{NLLB Team} et~al.(2022){NLLB Team}, Costa-jussà, Cross, Çelebi, Elbayad, Heafield, Heffernan, Kalbassi, Lam, Licht, Jean~Maillard, Wang, Wenzek, Youngblood, Akula, Barrault, Gonzalez, Hansanti, Hoffman, Jarrett, Sadagopan, Rowe, Spruit, Tran, Andrews, Ayan, Bhosale, Edunov, Fan, Gao, Goswami, Guzmán, Koehn, Mourachko, Ropers, Saleem, Schwenk, and Wang}]{nllb2022}
{NLLB Team}, Marta~R. Costa-jussà, James Cross, Onur Çelebi, Maha Elbayad, Kenneth Heafield, Kevin Heffernan, Elahe Kalbassi, Janice Lam, Daniel Licht, Anna~Sun Jean~Maillard, Skyler Wang, Guillaume Wenzek, Al~Youngblood, Bapi Akula, Loic Barrault, Gabriel~Mejia Gonzalez, Prangthip Hansanti, John Hoffman, Semarley Jarrett, Kaushik~Ram Sadagopan, Dirk Rowe, Shannon Spruit, Chau Tran, Pierre Andrews, Necip~Fazil Ayan, Shruti Bhosale, Sergey Edunov, Angela Fan, Cynthia Gao, Vedanuj Goswami, Francisco Guzmán, Philipp Koehn, Alexandre Mourachko, Christophe Ropers, Safiyyah Saleem, Holger Schwenk, and Jeff Wang. 2022.
\newblock No {L}anguage {L}eft {B}ehind: {S}caling {H}uman-{C}entered {M}achine {T}ranslation.

\bibitem[{OpenAI et~al.(2024)OpenAI, :, Hurst, Lerer, Goucher, Perelman, Ramesh, Clark, Ostrow, Welihinda, Hayes, Radford, Mądry, Baker-Whitcomb, Beutel, Borzunov, Carney, Chow, Kirillov, Nichol, Paino, Renzin, Passos, Kirillov, Christakis, Conneau, Kamali, Jabri, Moyer, Tam, Crookes, Tootoochian, Tootoonchian, Kumar, Vallone, Karpathy, Braunstein, Cann, Codispoti, Galu, Kondrich, Tulloch, Mishchenko, Baek, Jiang, Pelisse, Woodford, Gosalia, Dhar, Pantuliano, Nayak, Oliver, Zoph, Ghorbani, Leimberger, Rossen, Sokolowsky, Wang, Zweig, Hoover, Samic, McGrew, Spero, Giertler, Cheng, Lightcap, Walkin, Quinn, Guarraci, Hsu, Kellogg, Eastman, Lugaresi, Wainwright, Bassin, Hudson, Chu, Nelson, Li, Shern, Conger, Barette, Voss, Ding, Lu, Zhang, Beaumont, Hallacy, Koch, Gibson, Kim, Choi, McLeavey, Hesse, Fischer, Winter, Czarnecki, Jarvis, Wei, Koumouzelis, Sherburn, Kappler, Levin, Levy, Carr, Farhi, Mely, Robinson, Sasaki, Jin, Valladares, Tsipras, Li, Nguyen, Findlay, Oiwoh, Wong, Asdar, Proehl, Yang, Antonow,
  Kramer, Peterson, Sigler, Wallace, Brevdo, Mays, Khorasani, Such, Raso, Zhang, von Lohmann, Sulit, Goh, Oden, Salmon, Starace, Brockman, Salman, Bao, Hu, Wong, Wang, Schmidt, Whitney, Jun, Kirchner, de~Oliveira~Pinto, Ren, Chang, Chung, Kivlichan, O'Connell, O'Connell, Osband, Silber, Sohl, Okuyucu, Lan, Kostrikov, Sutskever, Kanitscheider, Gulrajani, Coxon, Menick, Pachocki, Aung, Betker, Crooks, Lennon, Kiros, Leike, Park, Kwon, Phang, Teplitz, Wei, Wolfe, Chen, Harris, Varavva, Lee, Shieh, Lin, Yu, Weng, Tang, Yu, Jang, Candela, Beutler, Landers, Parish, Heidecke, Schulman, Lachman, McKay, Uesato, Ward, Kim, Huizinga, Sitkin, Kraaijeveld, Gross, Kaplan, Snyder, Achiam, Jiao, Lee, Zhuang, Harriman, Fricke, Hayashi, Singhal, Shi, Karthik, Wood, Rimbach, Hsu, Nguyen, Gu-Lemberg, Button, Liu, Howe, Muthukumar, Luther, Ahmad, Kai, Itow, Workman, Pathak, Chen, Jing, Guy, Fedus, Zhou, Mamitsuka, Weng, McCallum, Held, Ouyang, Feuvrier, Zhang, Kondraciuk, Kaiser, Hewitt, Metz, Doshi, Aflak, Simens, Boyd,
  Thompson, Dukhan, Chen, Gray, Hudnall, Zhang, Aljubeh, Litwin, Zeng, Johnson, Shetty, Gupta, Shah, Yatbaz, Yang, Zhong, Glaese, Chen, Janner, Lampe, Petrov, Wu, Wang, Fradin, Pokrass, Castro, de~Castro, Pavlov, Brundage, Wang, Khan, Murati, Bavarian, Lin, Yesildal, Soto, Gimelshein, Cone, Staudacher, Summers, LaFontaine, Chowdhury, Ryder, Stathas, Turley, Tezak, Felix, Kudige, Keskar, Deutsch, Bundick, Puckett, Nachum, Okelola, Boiko, Murk, Jaffe, Watkins, Godement, Campbell-Moore, Chao, McMillan, Belov, Su, Bak, Bakkum, Deng, Dolan, Hoeschele, Welinder, Tillet, Pronin, Tillet, Dhariwal, Yuan, Dias, Lim, Arora, Troll, Lin, Lopes, Puri, Miyara, Leike, Gaubert, Zamani, Wang, Donnelly, Honsby, Smith, Sahai, Ramchandani, Huet, Carmichael, Zellers, Chen, Chen, Nigmatullin, Cheu, Jain, Altman, Schoenholz, Toizer, Miserendino, Agarwal, Culver, Ethersmith, Gray, Grove, Metzger, Hermani, Jain, Zhao, Wu, Jomoto, Wu, Shuaiqi, Xia, Phene, Papay, Narayanan, Coffey, Lee, Hall, Balaji, Broda, Stramer, Xu, Gogineni,
  Christianson, Sanders, Patwardhan, Cunninghman, Degry, Dimson, Raoux, Shadwell, Zheng, Underwood, Markov, Sherbakov, Rubin, Stasi, Kaftan, Heywood, Peterson, Walters, Eloundou, Qi, Moeller, Monaco, Kuo, Fomenko, Chang, Zheng, Zhou, Manassra, Sheu, Zaremba, Patil, Qian, Kim, Cheng, Zhang, He, Zhang, Jin, Dai, and Malkov}]{openai2024gpt4ocard}
OpenAI, :, Aaron Hurst, Adam Lerer, Adam~P. Goucher, Adam Perelman, Aditya Ramesh, Aidan Clark, AJ~Ostrow, Akila Welihinda, Alan Hayes, Alec Radford, Aleksander Mądry, Alex Baker-Whitcomb, Alex Beutel, Alex Borzunov, Alex Carney, Alex Chow, Alex Kirillov, Alex Nichol, Alex Paino, Alex Renzin, Alex~Tachard Passos, Alexander Kirillov, Alexi Christakis, Alexis Conneau, Ali Kamali, Allan Jabri, Allison Moyer, Allison Tam, Amadou Crookes, Amin Tootoochian, Amin Tootoonchian, Ananya Kumar, Andrea Vallone, Andrej Karpathy, Andrew Braunstein, Andrew Cann, Andrew Codispoti, Andrew Galu, Andrew Kondrich, Andrew Tulloch, Andrey Mishchenko, Angela Baek, Angela Jiang, Antoine Pelisse, Antonia Woodford, Anuj Gosalia, Arka Dhar, Ashley Pantuliano, Avi Nayak, Avital Oliver, Barret Zoph, Behrooz Ghorbani, Ben Leimberger, Ben Rossen, Ben Sokolowsky, Ben Wang, Benjamin Zweig, Beth Hoover, Blake Samic, Bob McGrew, Bobby Spero, Bogo Giertler, Bowen Cheng, Brad Lightcap, Brandon Walkin, Brendan Quinn, Brian Guarraci, Brian Hsu,
  Bright Kellogg, Brydon Eastman, Camillo Lugaresi, Carroll Wainwright, Cary Bassin, Cary Hudson, Casey Chu, Chad Nelson, Chak Li, Chan~Jun Shern, Channing Conger, Charlotte Barette, Chelsea Voss, Chen Ding, Cheng Lu, Chong Zhang, Chris Beaumont, Chris Hallacy, Chris Koch, Christian Gibson, Christina Kim, Christine Choi, Christine McLeavey, Christopher Hesse, Claudia Fischer, Clemens Winter, Coley Czarnecki, Colin Jarvis, Colin Wei, Constantin Koumouzelis, Dane Sherburn, Daniel Kappler, Daniel Levin, Daniel Levy, David Carr, David Farhi, David Mely, David Robinson, David Sasaki, Denny Jin, Dev Valladares, Dimitris Tsipras, Doug Li, Duc~Phong Nguyen, Duncan Findlay, Edede Oiwoh, Edmund Wong, Ehsan Asdar, Elizabeth Proehl, Elizabeth Yang, Eric Antonow, Eric Kramer, Eric Peterson, Eric Sigler, Eric Wallace, Eugene Brevdo, Evan Mays, Farzad Khorasani, Felipe~Petroski Such, Filippo Raso, Francis Zhang, Fred von Lohmann, Freddie Sulit, Gabriel Goh, Gene Oden, Geoff Salmon, Giulio Starace, Greg Brockman, Hadi
  Salman, Haiming Bao, Haitang Hu, Hannah Wong, Haoyu Wang, Heather Schmidt, Heather Whitney, Heewoo Jun, Hendrik Kirchner, Henrique~Ponde de~Oliveira~Pinto, Hongyu Ren, Huiwen Chang, Hyung~Won Chung, Ian Kivlichan, Ian O'Connell, Ian O'Connell, Ian Osband, Ian Silber, Ian Sohl, Ibrahim Okuyucu, Ikai Lan, Ilya Kostrikov, Ilya Sutskever, Ingmar Kanitscheider, Ishaan Gulrajani, Jacob Coxon, Jacob Menick, Jakub Pachocki, James Aung, James Betker, James Crooks, James Lennon, Jamie Kiros, Jan Leike, Jane Park, Jason Kwon, Jason Phang, Jason Teplitz, Jason Wei, Jason Wolfe, Jay Chen, Jeff Harris, Jenia Varavva, Jessica~Gan Lee, Jessica Shieh, Ji~Lin, Jiahui Yu, Jiayi Weng, Jie Tang, Jieqi Yu, Joanne Jang, Joaquin~Quinonero Candela, Joe Beutler, Joe Landers, Joel Parish, Johannes Heidecke, John Schulman, Jonathan Lachman, Jonathan McKay, Jonathan Uesato, Jonathan Ward, Jong~Wook Kim, Joost Huizinga, Jordan Sitkin, Jos Kraaijeveld, Josh Gross, Josh Kaplan, Josh Snyder, Joshua Achiam, Joy Jiao, Joyce Lee, Juntang
  Zhuang, Justyn Harriman, Kai Fricke, Kai Hayashi, Karan Singhal, Katy Shi, Kavin Karthik, Kayla Wood, Kendra Rimbach, Kenny Hsu, Kenny Nguyen, Keren Gu-Lemberg, Kevin Button, Kevin Liu, Kiel Howe, Krithika Muthukumar, Kyle Luther, Lama Ahmad, Larry Kai, Lauren Itow, Lauren Workman, Leher Pathak, Leo Chen, Li~Jing, Lia Guy, Liam Fedus, Liang Zhou, Lien Mamitsuka, Lilian Weng, Lindsay McCallum, Lindsey Held, Long Ouyang, Louis Feuvrier, Lu~Zhang, Lukas Kondraciuk, Lukasz Kaiser, Luke Hewitt, Luke Metz, Lyric Doshi, Mada Aflak, Maddie Simens, Madelaine Boyd, Madeleine Thompson, Marat Dukhan, Mark Chen, Mark Gray, Mark Hudnall, Marvin Zhang, Marwan Aljubeh, Mateusz Litwin, Matthew Zeng, Max Johnson, Maya Shetty, Mayank Gupta, Meghan Shah, Mehmet Yatbaz, Meng~Jia Yang, Mengchao Zhong, Mia Glaese, Mianna Chen, Michael Janner, Michael Lampe, Michael Petrov, Michael Wu, Michele Wang, Michelle Fradin, Michelle Pokrass, Miguel Castro, Miguel Oom~Temudo de~Castro, Mikhail Pavlov, Miles Brundage, Miles Wang, Minal
  Khan, Mira Murati, Mo~Bavarian, Molly Lin, Murat Yesildal, Nacho Soto, Natalia Gimelshein, Natalie Cone, Natalie Staudacher, Natalie Summers, Natan LaFontaine, Neil Chowdhury, Nick Ryder, Nick Stathas, Nick Turley, Nik Tezak, Niko Felix, Nithanth Kudige, Nitish Keskar, Noah Deutsch, Noel Bundick, Nora Puckett, Ofir Nachum, Ola Okelola, Oleg Boiko, Oleg Murk, Oliver Jaffe, Olivia Watkins, Olivier Godement, Owen Campbell-Moore, Patrick Chao, Paul McMillan, Pavel Belov, Peng Su, Peter Bak, Peter Bakkum, Peter Deng, Peter Dolan, Peter Hoeschele, Peter Welinder, Phil Tillet, Philip Pronin, Philippe Tillet, Prafulla Dhariwal, Qiming Yuan, Rachel Dias, Rachel Lim, Rahul Arora, Rajan Troll, Randall Lin, Rapha~Gontijo Lopes, Raul Puri, Reah Miyara, Reimar Leike, Renaud Gaubert, Reza Zamani, Ricky Wang, Rob Donnelly, Rob Honsby, Rocky Smith, Rohan Sahai, Rohit Ramchandani, Romain Huet, Rory Carmichael, Rowan Zellers, Roy Chen, Ruby Chen, Ruslan Nigmatullin, Ryan Cheu, Saachi Jain, Sam Altman, Sam Schoenholz, Sam
  Toizer, Samuel Miserendino, Sandhini Agarwal, Sara Culver, Scott Ethersmith, Scott Gray, Sean Grove, Sean Metzger, Shamez Hermani, Shantanu Jain, Shengjia Zhao, Sherwin Wu, Shino Jomoto, Shirong Wu, Shuaiqi, Xia, Sonia Phene, Spencer Papay, Srinivas Narayanan, Steve Coffey, Steve Lee, Stewart Hall, Suchir Balaji, Tal Broda, Tal Stramer, Tao Xu, Tarun Gogineni, Taya Christianson, Ted Sanders, Tejal Patwardhan, Thomas Cunninghman, Thomas Degry, Thomas Dimson, Thomas Raoux, Thomas Shadwell, Tianhao Zheng, Todd Underwood, Todor Markov, Toki Sherbakov, Tom Rubin, Tom Stasi, Tomer Kaftan, Tristan Heywood, Troy Peterson, Tyce Walters, Tyna Eloundou, Valerie Qi, Veit Moeller, Vinnie Monaco, Vishal Kuo, Vlad Fomenko, Wayne Chang, Weiyi Zheng, Wenda Zhou, Wesam Manassra, Will Sheu, Wojciech Zaremba, Yash Patil, Yilei Qian, Yongjik Kim, Youlong Cheng, Yu~Zhang, Yuchen He, Yuchen Zhang, Yujia Jin, Yunxing Dai, and Yury Malkov. 2024.
\newblock \href {https://arxiv.org/abs/2410.21276} {Gpt-4o system card}.
\newblock \emph{Preprint}, arXiv:2410.21276.

\bibitem[{Pizarro-Guevara(2010)}]{pizarro2010revisiting}
Jed~Sam Pizarro-Guevara. 2010.
\newblock \href {http://www.linguistics.berkeley.edu/~mikkelsen/pizarro_guevara_thesis.pdf} {{R}evisiting {Person-Case} constraint on ay-inversion in {T}agalog}.

\bibitem[{Popovi{\'c}(2017)}]{popovic-2017-chrf}
Maja Popovi{\'c}. 2017.
\newblock \href {https://doi.org/10.18653/v1/W17-4770} {{chrF}++: words helping character n-grams}.
\newblock In \emph{Proceedings of the Second Conference on Machine Translation}, pages 612--618, Copenhagen, Denmark. Association for Computational Linguistics.

\bibitem[{Powers(2012)}]{powers-2012-problem}
David Martin~Ward Powers. 2012.
\newblock \href {https://aclanthology.org/E12-1035/} {The {P}roblem with {K}appa}.
\newblock In \emph{Proceedings of the 13th Conference of the {E}uropean Chapter of the Association for Computational Linguistics}, pages 345--355, Avignon, France. Association for Computational Linguistics.

\bibitem[{Quakenbush(2005)}]{quakenbush2005philippine}
J~Stephen Quakenbush. 2005.
\newblock Philippine linguistics from an {SIL} perspective: {T}rends and prospects.
\newblock \emph{Current issues in Philippine linguistics and anthropology: Parangal kay Lawrence A. Reid}, pages 3--27.

\bibitem[{Qwen et~al.(2025)Qwen, :, Yang, Yang, Zhang, Hui, Zheng, Yu, Li, Liu, Huang, Wei, Lin, Yang, Tu, Zhang, Yang, Yang, Zhou, Lin, Dang, Lu, Bao, Yang, Yu, Li, Xue, Zhang, Zhu, Men, Lin, Li, Tang, Xia, Ren, Ren, Fan, Su, Zhang, Wan, Liu, Cui, Zhang, and Qiu}]{qwen2025qwen25technicalreport}
Qwen, :, An~Yang, Baosong Yang, Beichen Zhang, Binyuan Hui, Bo~Zheng, Bowen Yu, Chengyuan Li, Dayiheng Liu, Fei Huang, Haoran Wei, Huan Lin, Jian Yang, Jianhong Tu, Jianwei Zhang, Jianxin Yang, Jiaxi Yang, Jingren Zhou, Junyang Lin, Kai Dang, Keming Lu, Keqin Bao, Kexin Yang, Le~Yu, Mei Li, Mingfeng Xue, Pei Zhang, Qin Zhu, Rui Men, Runji Lin, Tianhao Li, Tianyi Tang, Tingyu Xia, Xingzhang Ren, Xuancheng Ren, Yang Fan, Yang Su, Yichang Zhang, Yu~Wan, Yuqiong Liu, Zeyu Cui, Zhenru Zhang, and Zihan Qiu. 2025.
\newblock \href {https://arxiv.org/abs/2412.15115} {Qwen2.5 technical report}.
\newblock \emph{Preprint}, arXiv:2412.15115.

\bibitem[{Ramos(2021)}]{ramos2021tagalog}
Teresita~V Ramos. 2021.
\newblock \emph{Tagalog structures}.
\newblock University of Hawaii Press.

\bibitem[{Reid(2018)}]{reid2018modeling}
Lawrence~A Reid. 2018.
\newblock Modeling the linguistic situation in the {P}hilippines.
\newblock \emph{Senri ethnological studies}, 98:91--105.

\bibitem[{Riley et~al.(2020)Riley, Caswell, Freitag, and Grangier}]{riley_translationese_2020}
Parker Riley, Isaac Caswell, Markus Freitag, and David Grangier. 2020.
\newblock \href {https://doi.org/10.18653/v1/2020.acl-main.691} {Translationese as a {Language} in “{Multilingual}” {NMT}}.
\newblock In \emph{Proceedings of the 58th {Annual} {Meeting} of the {Association} for {Computational} {Linguistics}}, pages 7737--7746, Online. Association for Computational Linguistics.

\bibitem[{Rivera-Trigueros(2022)}]{rivera2022machine}
Irene Rivera-Trigueros. 2022.
\newblock Machine translation systems and quality assessment: a systematic review.
\newblock \emph{Language Resources and Evaluation}, 56(2):593--619.

\bibitem[{Son et~al.(2024)Son, Yoon, Suk, Aula-Blasco, Aslan, Kim, Islam, Prats-Cristi{\`a}, Tormo-Ba{\~n}uelos, and Kim}]{son2024mm}
Guijin Son, Dongkeun Yoon, Juyoung Suk, Javier Aula-Blasco, Mano Aslan, Vu~Trong Kim, Shayekh~Bin Islam, Jaume Prats-Cristi{\`a}, Luc{\'\i}a Tormo-Ba{\~n}uelos, and Seungone Kim. 2024.
\newblock Mm-eval: A {M}ultilingual {M}eta-{E}valuation {B}enchmark for {LLM}-as-a-judge and {R}eward {M}odels.
\newblock \emph{arXiv preprint arXiv:2410.17578}.

\bibitem[{Tanawan et~al.(2008)Tanawan, Nacin, and J.K.}]{tanawanetal2008istruktura}
D.S. Tanawan, A.A. Nacin, and Lartec J.K. 2008.
\newblock \emph{Istruktura ng {W}ikang {F}ilipino.}
\newblock Trinitas Publishing House.

\bibitem[{Team et~al.(2024{\natexlab{a}})Team, Georgiev, Lei, Burnell, Bai, Gulati, Tanzer, Vincent, Pan, Wang, Mariooryad, Ding, Geng, Alcober, Frostig, Omernick, Walker, Paduraru, Sorokin, Tacchetti, Gaffney, Daruki, Sercinoglu, Gleicher, Love, Voigtlaender, Jain, Surita, Mohamed, Blevins, Ahn, Zhu, Kawintiranon, Firat, Gu, Zhang, Rahtz, Faruqui, Clay, Gilmer, Co-Reyes, Penchev, Zhu, Morioka, Hui, Haridasan, Campos, Mahdieh, Guo, Hassan, Kilgour, Vezer, Cheng, de~Liedekerke, Goyal, Barham, Strouse, Noury, Adler, Sundararajan, Vikram, Lepikhin, Paganini, Garcia, Yang, Valter, Trebacz, Vodrahalli, Asawaroengchai, Ring, Kalb, Soares, Brahma, Steiner, Yu, Mentzer, He, Gonzalez, Xu, Kaufman, Shafey, Oh, Hennigan, van~den Driessche, Odoom, Lucic, Roelofs, Lall, Marathe, Chan, Ontanon, He, Teplyashin, Lai, Crone, Damoc, Ho, Riedel, Lenc, Yeh, Chowdhery, Xu, Kazemi, Amid, Petrushkina, Swersky, Khodaei, Chen, Larkin, Pinto, Yan, Badia, Patil, Hansen, Orr, Arnold, Grimstad, Dai, Douglas, Sinha, Yadav, Chen,
  Gribovskaya, Austin, Zhao, Patel, Komarek, Austin, Borgeaud, Friso, Goyal, Caine, Cao, Chung, Lamm, Barth-Maron, Kagohara, Olszewska, Chen, Shivakumar, Agarwal, Godhia, Rajwar, Snaider, Dotiwalla, Liu, Barua, Ungureanu, Zhang, Batsaikhan, Wirth, Qin, Danihelka, Doshi, Chadwick, Chen, Jain, Le, Kar, Gurumurthy, Li, Sang, Liu, Lamprou, Munoz, Lintz, Mehta, Howard, Reynolds, Aroyo, Wang, Blanco, Cassirer, Griffith, Das, Lee, Sygnowski, Fisher, Besley, Powell, Ahmed, Paulus, Reitter, Borsos, Joshi, Pope, Hand, Selo, Jain, Sethi, Goel, Makino, May, Yang, Schalkwyk, Butterfield, Hauth, Goldin, Hawkins, Senter, Brin, Woodman, Ritter, Noland, Giang, Bolina, Lee, Blyth, Mackinnon, Reid, Sarvana, Silver, Chen, Wang, Maggiore, Chang, Attaluri, Thornton, Chiu, Bunyan, Levine, Chung, Eltyshev, Si, Lillicrap, Brady, Aggarwal, Wu, Xu, McIlroy, Badola, Sandhu, Moreira, Stokowiec, Hemsley, Li, Tudor, Shyam, Rahimtoroghi, Haykal, Sprechmann, Zhou, Mincu, Li, Addanki, Krishna, Wu, Frechette, Eyal, Dafoe, Lacey, Whang,
  Avrahami, Zhang, Taropa, Lin, Toyama, Rutherford, Sano, Choe, Tomala, Safranek-Shrader, Kassner, Pajarskas, Harvey, Sechrist, Fortunato, Lyu, Elsayed, Kuang, Lottes, Chu, Jia, Chen, Humphreys, Baumli, Tao, Samuel, dos Santos, Andreassen, Rakićević, Grewe, Kumar, Winkler, Caton, Brock, Dalmia, Sheahan, Barr, Miao, Natsev, Devlin, Behbahani, Prost, Sun, Myaskovsky, Pillai, Hurt, Lazaridou, Xiong, Zheng, Pardo, Li, Horgan, Stanton, Ambar, Xia, Lince, Wang, Mustafa, Webson, Lee, Anil, Wicke, Dozat, Sinha, Piqueras, Dabir, Upadhyay, Boral, Hendricks, Fry, Djolonga, Su, Walker, Labanowski, Huang, Misra, Chen, Skerry-Ryan, Singh, Rijhwani, Yu, Castro-Ros, Changpinyo, Datta, Bagri, Hrafnkelsson, Maggioni, Zheng, Sulsky, Hou, Paine, Yang, Riesa, Rogozinska, Marcus, Badawy, Zhang, Wang, Miller, Greer, Sjos, Nova, Zen, Chaabouni, Rosca, Jiang, Chen, Liu, Sainath, Krikun, Polozov, Lespiau, Newlan, Cankara, Kwak, Xu, Chen, Coenen, Meyer, Tsihlas, Ma, Gottweis, Xing, Gu, Miao, Frank, Cankara, Ganapathy, Dasgupta,
  Hughes-Fitt, Chen, Reid, Rong, Fan, van Amersfoort, Zhuang, Cohen, Gu, Mohananey, Ilic, Tobin, Wieting, Bortsova, Thacker, Wang, Caveness, Chiu, Sezener, Kaskasoli, Baker, Millican, Elhawaty, Aisopos, Lebsack, Byrd, Dai, Jia, Wiethoff, Davoodi, Weston, Yagati, Ahuja, Gao, Pundak, Zhang, Azzam, Sim, Caelles, Keeling, Sharma, Swing, Li, Liu, Bostock, Bansal, Nado, Anand, Lipschultz, Karmarkar, Proleev, Ittycheriah, Yeganeh, Polovets, Faust, Sun, Rrustemi, Li, Shivanna, Liu, Welty, Lebron, Baddepudi, Krause, Parisotto, Soricut, Xu, Bloxwich, Johnson, Neyshabur, Mao-Jones, Wang, Ramasesh, Abbas, Guez, Segal, Nguyen, Svensson, Hou, York, Milan, Bridgers, Gworek, Tagliasacchi, Lee-Thorp, Chang, Guseynov, Hartman, Kwong, Zhao, Kashem, Cole, Miech, Tanburn, Phuong, Pavetic, Cevey, Comanescu, Ives, Yang, Du, Li, Zhang, Iinuma, Hu, Roy, Bijwadia, Zhu, Martins, Saputro, Gergely, Zheng, Jia, Antonoglou, Sadovsky, Gu, Bi, Andreev, Samangooei, Khan, Kocisky, Filos, Kumar, Bishop, Yu, Hodkinson, Mittal, Shah, Moufarek,
  Cheng, Bloniarz, Lee, Pejman, Michel, Spencer, Feinberg, Xiong, Savinov, Smith, Shakeri, Tran, Chesus, Bohnet, Tucker, von Glehn, Muir, Mao, Kazawa, Slone, Soparkar, Shrivastava, Cobon-Kerr, Sharman, Pavagadhi, Araya, Misiunas, Ghelani, Laskin, Barker, Li, Briukhov, Houlsby, Glaese, Lakshminarayanan, Schucher, Tang, Collins, Lim, Feng, Recasens, Lai, Magni, Cao, Siddhant, Ashwood, Orbay, Dehghani, Brennan, He, Xu, Gao, Saroufim, Molloy, Wu, Arnold, Chang, Schrittwieser, Buchatskaya, Radpour, Polacek, Giordano, Bapna, Tokumine, Hellendoorn, Sottiaux, Cogan, Severyn, Saleh, Thakoor, Shefey, Qiao, Gaba, yiin Chang, Swanson, Zhang, Lee, Rubenstein, Song, Kwiatkowski, Koop, Kannan, Kao, Schuh, Stjerngren, Ghiasi, Gibson, Vilnis, Yuan, Ferreira, Kamath, Klimenko, Franko, Xiao, Bhattacharya, Patel, Wang, Morris, Strudel, Sharma, Choy, Hashemi, Landon, Finkelstein, Jhakra, Frye, Barnes, Mauger, Daun, Baatarsukh, Tung, Farhan, Michalewski, Viola, de~Chaumont~Quitry, Lan, Hudson, Wang, Fischer, Zheng, White, Dragan,
  baptiste Alayrac, Ni, Pritzel, Iwanicki, Isard, Bulanova, Zilka, Dyer, Sachan, Srinivasan, Muckenhirn, Cai, Mandhane, Tariq, Rae, Wang, Ayoub, FitzGerald, Zhao, Han, Alberti, Garrette, Krishnakumar, Gimenez, Levskaya, Sohn, Matak, Iturrate, Chang, Xiang, Cao, Ranka, Brown, Hutter, Mirrokni, Chen, Yao, Egyed, Galilee, Liechty, Kallakuri, Palmer, Ghemawat, Liu, Tao, Thornton, Green, Jasarevic, Lin, Cotruta, Tan, Fiedel, Yu, Chi, Neitz, Heitkaemper, Sinha, Zhou, Sun, Kaed, Hulse, Mishra, Georgaki, Kudugunta, Farabet, Shafran, Vlasic, Tsitsulin, Ananthanarayanan, Carin, Su, Sun, V, Carvajal, Broder, Comsa, Repina, Wong, Chen, Hawkins, Filonov, Loher, Hirnschall, Wang, Ye, Burns, Cate, Wright, Piccinini, Zhang, Lin, Gog, Kulizhskaya, Sreevatsa, Song, Cobo, Iyer, Tekur, Garrido, Xiao, Kemp, Zheng, Li, Agarwal, Ngani, Goshvadi, Santamaria-Fernandez, Fica, Chen, Gorgolewski, Sun, Garg, Ye, Eslami, Hua, Simon, Joshi, Kim, Tenney, Potluri, Thiet, Yuan, Luisier, Chronopoulou, Scellato, Srinivasan, Chen, Koverkathu,
  Dalibard, Xu, Saeta, Anderson, Sellam, Fernando, Huot, Jung, Varadarajan, Quinn, Raul, Le, Habalov, Clark, Jalan, Bullard, Singhal, Luong, Wang, Rajayogam, Eisenschlos, Jia, Finchelstein, Yakubovich, Balle, Fink, Agarwal, Li, Dvijotham, Pal, Kang, Konzelmann, Beattie, Dousse, Wu, Crocker, Elkind, Jonnalagadda, Lee, Holtmann-Rice, Kallarackal, Liu, Vnukov, Vats, Invernizzi, Jafari, Zhou, Taylor, Prendki, Wu, Eccles, Liu, Kopparapu, Beaufays, Angermueller, Marzoca, Sarcar, Dib, Stanway, Perbet, Trdin, Sterneck, Khorlin, Li, Wu, Goenka, Madras, Goldshtein, Gierke, Zhou, Liu, Liang, White, Li, Singh, Bahargam, Epstein, Basu, Lao, Ozturel, Crous, Zhai, Lu, Tung, Gaur, Walton, Dixon, Zhang, Globerson, Uy, Bolt, Wiles, Nasr, Shumailov, Selvi, Piccinno, Aguilar, McCarthy, Khalman, Shukla, Galic, Carpenter, Villela, Zhang, Richardson, Martens, Bosnjak, Belle, Seibert, Alnahlawi, McWilliams, Singh, Louis, Ding, Popovici, Simicich, Knight, Mehta, Gupta, Shi, Fatehi, Mitrovic, Grills, Pagadora, Munkhdalai, Petrova,
  Eisenbud, Zhang, Yates, Mittal, Tripuraneni, Assael, Brovelli, Jain, Velimirovic, Akbulut, Mu, Macherey, Kumar, Xu, Qureshi, Comanici, Wiesner, Gong, Ruddock, Bauer, Felt, GP, Arnab, Zelle, Rothfuss, Rosgen, Shenoy, Seybold, Li, Mudigonda, Erdogan, Xia, Simsa, Michi, Yao, Yew, Kan, Caswell, Radebaugh, Elisseeff, Valenzuela, McKinney, Paterson, Cui, Latorre-Chimoto, Kim, Zeng, Durden, Ponnapalli, Sosea, Choquette-Choo, Manyika, Robenek, Vashisht, Pereira, Lam, Velic, Owusu-Afriyie, Lee, Bolukbasi, Parrish, Lu, Park, Venkatraman, Talbert, Rosique, Cheng, Sozanschi, Paszke, Kumar, Austin, Li, Salama, Perz, Kim, Dukkipati, Baryshnikov, Kaplanis, Sheng, Chervonyi, Unlu, de~Las~Casas, Askham, Tunyasuvunakool, Gimeno, Poder, Kwak, Miecnikowski, Mirrokni, Dimitriev, Parisi, Liu, Tsai, Shevlane, Kouridi, Garmon, Goedeckemeyer, Brown, Vijayakumar, Elqursh, Jazayeri, Huang, Carthy, Hoover, Kim, Kumar, Chen, Biles, Bingham, Rosen, Wang, Tan, Engel, Pongetti, de~Cesare, Hwang, Yu, Pullman, Narayanan, Levin, Gopal, Li,
  Aharoni, Trinh, Lo, Casagrande, Vij, Matthey, Ramadhana, Matthews, Carey, Johnson, Goranova, Shah, Ashraf, Dasgupta, Larsen, Wang, Vuyyuru, Jiang, Ijazi, Osawa, Smith, Boppana, Bilal, Koizumi, Xu, Altun, Shabat, Bariach, Korchemniy, Choo, Ronneberger, Iwuanyanwu, Zhao, Soergel, Hsieh, Cai, Iqbal, Sundermeyer, Chen, Bursztein, Malaviya, Biadsy, Shroff, Dhillon, Latkar, Dyer, Forbes, Nicosia, Nikolaev, Greene, Georgiev, Wang, Martin, Sedghi, Zhang, Banzal, Fritz, Rao, Wang, Zhang, Patraucean, Du, Mordatch, Jurin, Liu, Dubey, Mohan, Nowakowski, Ion, Wei, Tojo, Raad, Hudson, Keshava, Agrawal, Ramirez, Wu, Nguyen, Liu, Sewak, Petrini, Choi, Philips, Wang, Bica, Garg, Wilkiewicz, Agrawal, Li, Guo, Xue, Shaik, Leach, Khan, Wiesinger, Jerome, Chakladar, Wang, Ornduff, Abu, Ghaffarkhah, Wainwright, Cortes, Liu, Maynez, Terzis, Samangouei, Mansour, Kępa, Aubet, Algymr, Banica, Weisz, Orban, Senges, Andrejczuk, Geller, Santo, Anklin, Merey, Baeuml, Strohman, Bai, Petrov, Wu, Hassabis, Kavukcuoglu, Dean, and
  Vinyals}]{geminiteam2024gemini15unlockingmultimodal}
Gemini Team, Petko Georgiev, Ving~Ian Lei, Ryan Burnell, Libin Bai, Anmol Gulati, Garrett Tanzer, Damien Vincent, Zhufeng Pan, Shibo Wang, Soroosh Mariooryad, Yifan Ding, Xinyang Geng, Fred Alcober, Roy Frostig, Mark Omernick, Lexi Walker, Cosmin Paduraru, Christina Sorokin, Andrea Tacchetti, Colin Gaffney, Samira Daruki, Olcan Sercinoglu, Zach Gleicher, Juliette Love, Paul Voigtlaender, Rohan Jain, Gabriela Surita, Kareem Mohamed, Rory Blevins, Junwhan Ahn, Tao Zhu, Kornraphop Kawintiranon, Orhan Firat, Yiming Gu, Yujing Zhang, Matthew Rahtz, Manaal Faruqui, Natalie Clay, Justin Gilmer, JD~Co-Reyes, Ivo Penchev, Rui Zhu, Nobuyuki Morioka, Kevin Hui, Krishna Haridasan, Victor Campos, Mahdis Mahdieh, Mandy Guo, Samer Hassan, Kevin Kilgour, Arpi Vezer, Heng-Tze Cheng, Raoul de~Liedekerke, Siddharth Goyal, Paul Barham, DJ~Strouse, Seb Noury, Jonas Adler, Mukund Sundararajan, Sharad Vikram, Dmitry Lepikhin, Michela Paganini, Xavier Garcia, Fan Yang, Dasha Valter, Maja Trebacz, Kiran Vodrahalli, Chulayuth
  Asawaroengchai, Roman Ring, Norbert Kalb, Livio~Baldini Soares, Siddhartha Brahma, David Steiner, Tianhe Yu, Fabian Mentzer, Antoine He, Lucas Gonzalez, Bibo Xu, Raphael~Lopez Kaufman, Laurent~El Shafey, Junhyuk Oh, Tom Hennigan, George van~den Driessche, Seth Odoom, Mario Lucic, Becca Roelofs, Sid Lall, Amit Marathe, Betty Chan, Santiago Ontanon, Luheng He, Denis Teplyashin, Jonathan Lai, Phil Crone, Bogdan Damoc, Lewis Ho, Sebastian Riedel, Karel Lenc, Chih-Kuan Yeh, Aakanksha Chowdhery, Yang Xu, Mehran Kazemi, Ehsan Amid, Anastasia Petrushkina, Kevin Swersky, Ali Khodaei, Gowoon Chen, Chris Larkin, Mario Pinto, Geng Yan, Adria~Puigdomenech Badia, Piyush Patil, Steven Hansen, Dave Orr, Sebastien M.~R. Arnold, Jordan Grimstad, Andrew Dai, Sholto Douglas, Rishika Sinha, Vikas Yadav, Xi~Chen, Elena Gribovskaya, Jacob Austin, Jeffrey Zhao, Kaushal Patel, Paul Komarek, Sophia Austin, Sebastian Borgeaud, Linda Friso, Abhimanyu Goyal, Ben Caine, Kris Cao, Da-Woon Chung, Matthew Lamm, Gabe Barth-Maron, Thais
  Kagohara, Kate Olszewska, Mia Chen, Kaushik Shivakumar, Rishabh Agarwal, Harshal Godhia, Ravi Rajwar, Javier Snaider, Xerxes Dotiwalla, Yuan Liu, Aditya Barua, Victor Ungureanu, Yuan Zhang, Bat-Orgil Batsaikhan, Mateo Wirth, James Qin, Ivo Danihelka, Tulsee Doshi, Martin Chadwick, Jilin Chen, Sanil Jain, Quoc Le, Arjun Kar, Madhu Gurumurthy, Cheng Li, Ruoxin Sang, Fangyu Liu, Lampros Lamprou, Rich Munoz, Nathan Lintz, Harsh Mehta, Heidi Howard, Malcolm Reynolds, Lora Aroyo, Quan Wang, Lorenzo Blanco, Albin Cassirer, Jordan Griffith, Dipanjan Das, Stephan Lee, Jakub Sygnowski, Zach Fisher, James Besley, Richard Powell, Zafarali Ahmed, Dominik Paulus, David Reitter, Zalan Borsos, Rishabh Joshi, Aedan Pope, Steven Hand, Vittorio Selo, Vihan Jain, Nikhil Sethi, Megha Goel, Takaki Makino, Rhys May, Zhen Yang, Johan Schalkwyk, Christina Butterfield, Anja Hauth, Alex Goldin, Will Hawkins, Evan Senter, Sergey Brin, Oliver Woodman, Marvin Ritter, Eric Noland, Minh Giang, Vijay Bolina, Lisa Lee, Tim Blyth, Ian
  Mackinnon, Machel Reid, Obaid Sarvana, David Silver, Alexander Chen, Lily Wang, Loren Maggiore, Oscar Chang, Nithya Attaluri, Gregory Thornton, Chung-Cheng Chiu, Oskar Bunyan, Nir Levine, Timothy Chung, Evgenii Eltyshev, Xiance Si, Timothy Lillicrap, Demetra Brady, Vaibhav Aggarwal, Boxi Wu, Yuanzhong Xu, Ross McIlroy, Kartikeya Badola, Paramjit Sandhu, Erica Moreira, Wojciech Stokowiec, Ross Hemsley, Dong Li, Alex Tudor, Pranav Shyam, Elahe Rahimtoroghi, Salem Haykal, Pablo Sprechmann, Xiang Zhou, Diana Mincu, Yujia Li, Ravi Addanki, Kalpesh Krishna, Xiao Wu, Alexandre Frechette, Matan Eyal, Allan Dafoe, Dave Lacey, Jay Whang, Thi Avrahami, Ye~Zhang, Emanuel Taropa, Hanzhao Lin, Daniel Toyama, Eliza Rutherford, Motoki Sano, HyunJeong Choe, Alex Tomala, Chalence Safranek-Shrader, Nora Kassner, Mantas Pajarskas, Matt Harvey, Sean Sechrist, Meire Fortunato, Christina Lyu, Gamaleldin Elsayed, Chenkai Kuang, James Lottes, Eric Chu, Chao Jia, Chih-Wei Chen, Peter Humphreys, Kate Baumli, Connie Tao, Rajkumar
  Samuel, Cicero~Nogueira dos Santos, Anders Andreassen, Nemanja Rakićević, Dominik Grewe, Aviral Kumar, Stephanie Winkler, Jonathan Caton, Andrew Brock, Sid Dalmia, Hannah Sheahan, Iain Barr, Yingjie Miao, Paul Natsev, Jacob Devlin, Feryal Behbahani, Flavien Prost, Yanhua Sun, Artiom Myaskovsky, Thanumalayan~Sankaranarayana Pillai, Dan Hurt, Angeliki Lazaridou, Xi~Xiong, Ce~Zheng, Fabio Pardo, Xiaowei Li, Dan Horgan, Joe Stanton, Moran Ambar, Fei Xia, Alejandro Lince, Mingqiu Wang, Basil Mustafa, Albert Webson, Hyo Lee, Rohan Anil, Martin Wicke, Timothy Dozat, Abhishek Sinha, Enrique Piqueras, Elahe Dabir, Shyam Upadhyay, Anudhyan Boral, Lisa~Anne Hendricks, Corey Fry, Josip Djolonga, Yi~Su, Jake Walker, Jane Labanowski, Ronny Huang, Vedant Misra, Jeremy Chen, RJ~Skerry-Ryan, Avi Singh, Shruti Rijhwani, Dian Yu, Alex Castro-Ros, Beer Changpinyo, Romina Datta, Sumit Bagri, Arnar~Mar Hrafnkelsson, Marcello Maggioni, Daniel Zheng, Yury Sulsky, Shaobo Hou, Tom~Le Paine, Antoine Yang, Jason Riesa, Dominika
  Rogozinska, Dror Marcus, Dalia~El Badawy, Qiao Zhang, Luyu Wang, Helen Miller, Jeremy Greer, Lars~Lowe Sjos, Azade Nova, Heiga Zen, Rahma Chaabouni, Mihaela Rosca, Jiepu Jiang, Charlie Chen, Ruibo Liu, Tara Sainath, Maxim Krikun, Alex Polozov, Jean-Baptiste Lespiau, Josh Newlan, Zeyncep Cankara, Soo Kwak, Yunhan Xu, Phil Chen, Andy Coenen, Clemens Meyer, Katerina Tsihlas, Ada Ma, Juraj Gottweis, Jinwei Xing, Chenjie Gu, Jin Miao, Christian Frank, Zeynep Cankara, Sanjay Ganapathy, Ishita Dasgupta, Steph Hughes-Fitt, Heng Chen, David Reid, Keran Rong, Hongmin Fan, Joost van Amersfoort, Vincent Zhuang, Aaron Cohen, Shixiang~Shane Gu, Anhad Mohananey, Anastasija Ilic, Taylor Tobin, John Wieting, Anna Bortsova, Phoebe Thacker, Emma Wang, Emily Caveness, Justin Chiu, Eren Sezener, Alex Kaskasoli, Steven Baker, Katie Millican, Mohamed Elhawaty, Kostas Aisopos, Carl Lebsack, Nathan Byrd, Hanjun Dai, Wenhao Jia, Matthew Wiethoff, Elnaz Davoodi, Albert Weston, Lakshman Yagati, Arun Ahuja, Isabel Gao, Golan Pundak,
  Susan Zhang, Michael Azzam, Khe~Chai Sim, Sergi Caelles, James Keeling, Abhanshu Sharma, Andy Swing, YaGuang Li, Chenxi Liu, Carrie~Grimes Bostock, Yamini Bansal, Zachary Nado, Ankesh Anand, Josh Lipschultz, Abhijit Karmarkar, Lev Proleev, Abe Ittycheriah, Soheil~Hassas Yeganeh, George Polovets, Aleksandra Faust, Jiao Sun, Alban Rrustemi, Pen Li, Rakesh Shivanna, Jeremiah Liu, Chris Welty, Federico Lebron, Anirudh Baddepudi, Sebastian Krause, Emilio Parisotto, Radu Soricut, Zheng Xu, Dawn Bloxwich, Melvin Johnson, Behnam Neyshabur, Justin Mao-Jones, Renshen Wang, Vinay Ramasesh, Zaheer Abbas, Arthur Guez, Constant Segal, Duc~Dung Nguyen, James Svensson, Le~Hou, Sarah York, Kieran Milan, Sophie Bridgers, Wiktor Gworek, Marco Tagliasacchi, James Lee-Thorp, Michael Chang, Alexey Guseynov, Ale~Jakse Hartman, Michael Kwong, Ruizhe Zhao, Sheleem Kashem, Elizabeth Cole, Antoine Miech, Richard Tanburn, Mary Phuong, Filip Pavetic, Sebastien Cevey, Ramona Comanescu, Richard Ives, Sherry Yang, Cosmo Du, Bo~Li, Zizhao
  Zhang, Mariko Iinuma, Clara~Huiyi Hu, Aurko Roy, Shaan Bijwadia, Zhenkai Zhu, Danilo Martins, Rachel Saputro, Anita Gergely, Steven Zheng, Dawei Jia, Ioannis Antonoglou, Adam Sadovsky, Shane Gu, Yingying Bi, Alek Andreev, Sina Samangooei, Mina Khan, Tomas Kocisky, Angelos Filos, Chintu Kumar, Colton Bishop, Adams Yu, Sarah Hodkinson, Sid Mittal, Premal Shah, Alexandre Moufarek, Yong Cheng, Adam Bloniarz, Jaehoon Lee, Pedram Pejman, Paul Michel, Stephen Spencer, Vladimir Feinberg, Xuehan Xiong, Nikolay Savinov, Charlotte Smith, Siamak Shakeri, Dustin Tran, Mary Chesus, Bernd Bohnet, George Tucker, Tamara von Glehn, Carrie Muir, Yiran Mao, Hideto Kazawa, Ambrose Slone, Kedar Soparkar, Disha Shrivastava, James Cobon-Kerr, Michael Sharman, Jay Pavagadhi, Carlos Araya, Karolis Misiunas, Nimesh Ghelani, Michael Laskin, David Barker, Qiujia Li, Anton Briukhov, Neil Houlsby, Mia Glaese, Balaji Lakshminarayanan, Nathan Schucher, Yunhao Tang, Eli Collins, Hyeontaek Lim, Fangxiaoyu Feng, Adria Recasens, Guangda Lai,
  Alberto Magni, Nicola~De Cao, Aditya Siddhant, Zoe Ashwood, Jordi Orbay, Mostafa Dehghani, Jenny Brennan, Yifan He, Kelvin Xu, Yang Gao, Carl Saroufim, James Molloy, Xinyi Wu, Seb Arnold, Solomon Chang, Julian Schrittwieser, Elena Buchatskaya, Soroush Radpour, Martin Polacek, Skye Giordano, Ankur Bapna, Simon Tokumine, Vincent Hellendoorn, Thibault Sottiaux, Sarah Cogan, Aliaksei Severyn, Mohammad Saleh, Shantanu Thakoor, Laurent Shefey, Siyuan Qiao, Meenu Gaba, Shuo yiin Chang, Craig Swanson, Biao Zhang, Benjamin Lee, Paul~Kishan Rubenstein, Gan Song, Tom Kwiatkowski, Anna Koop, Ajay Kannan, David Kao, Parker Schuh, Axel Stjerngren, Golnaz Ghiasi, Gena Gibson, Luke Vilnis, Ye~Yuan, Felipe~Tiengo Ferreira, Aishwarya Kamath, Ted Klimenko, Ken Franko, Kefan Xiao, Indro Bhattacharya, Miteyan Patel, Rui Wang, Alex Morris, Robin Strudel, Vivek Sharma, Peter Choy, Sayed~Hadi Hashemi, Jessica Landon, Mara Finkelstein, Priya Jhakra, Justin Frye, Megan Barnes, Matthew Mauger, Dennis Daun, Khuslen Baatarsukh, Matthew
  Tung, Wael Farhan, Henryk Michalewski, Fabio Viola, Felix de~Chaumont~Quitry, Charline~Le Lan, Tom Hudson, Qingze Wang, Felix Fischer, Ivy Zheng, Elspeth White, Anca Dragan, Jean baptiste Alayrac, Eric Ni, Alexander Pritzel, Adam Iwanicki, Michael Isard, Anna Bulanova, Lukas Zilka, Ethan Dyer, Devendra Sachan, Srivatsan Srinivasan, Hannah Muckenhirn, Honglong Cai, Amol Mandhane, Mukarram Tariq, Jack~W. Rae, Gary Wang, Kareem Ayoub, Nicholas FitzGerald, Yao Zhao, Woohyun Han, Chris Alberti, Dan Garrette, Kashyap Krishnakumar, Mai Gimenez, Anselm Levskaya, Daniel Sohn, Josip Matak, Inaki Iturrate, Michael~B. Chang, Jackie Xiang, Yuan Cao, Nishant Ranka, Geoff Brown, Adrian Hutter, Vahab Mirrokni, Nanxin Chen, Kaisheng Yao, Zoltan Egyed, Francois Galilee, Tyler Liechty, Praveen Kallakuri, Evan Palmer, Sanjay Ghemawat, Jasmine Liu, David Tao, Chloe Thornton, Tim Green, Mimi Jasarevic, Sharon Lin, Victor Cotruta, Yi-Xuan Tan, Noah Fiedel, Hongkun Yu, Ed~Chi, Alexander Neitz, Jens Heitkaemper, Anu Sinha, Denny
  Zhou, Yi~Sun, Charbel Kaed, Brice Hulse, Swaroop Mishra, Maria Georgaki, Sneha Kudugunta, Clement Farabet, Izhak Shafran, Daniel Vlasic, Anton Tsitsulin, Rajagopal Ananthanarayanan, Alen Carin, Guolong Su, Pei Sun, Shashank V, Gabriel Carvajal, Josef Broder, Iulia Comsa, Alena Repina, William Wong, Warren~Weilun Chen, Peter Hawkins, Egor Filonov, Lucia Loher, Christoph Hirnschall, Weiyi Wang, Jingchen Ye, Andrea Burns, Hardie Cate, Diana~Gage Wright, Federico Piccinini, Lei Zhang, Chu-Cheng Lin, Ionel Gog, Yana Kulizhskaya, Ashwin Sreevatsa, Shuang Song, Luis~C. Cobo, Anand Iyer, Chetan Tekur, Guillermo Garrido, Zhuyun Xiao, Rupert Kemp, Huaixiu~Steven Zheng, Hui Li, Ananth Agarwal, Christel Ngani, Kati Goshvadi, Rebeca Santamaria-Fernandez, Wojciech Fica, Xinyun Chen, Chris Gorgolewski, Sean Sun, Roopal Garg, Xinyu Ye, S.~M.~Ali Eslami, Nan Hua, Jon Simon, Pratik Joshi, Yelin Kim, Ian Tenney, Sahitya Potluri, Lam~Nguyen Thiet, Quan Yuan, Florian Luisier, Alexandra Chronopoulou, Salvatore Scellato, Praveen
  Srinivasan, Minmin Chen, Vinod Koverkathu, Valentin Dalibard, Yaming Xu, Brennan Saeta, Keith Anderson, Thibault Sellam, Nick Fernando, Fantine Huot, Junehyuk Jung, Mani Varadarajan, Michael Quinn, Amit Raul, Maigo Le, Ruslan Habalov, Jon Clark, Komal Jalan, Kalesha Bullard, Achintya Singhal, Thang Luong, Boyu Wang, Sujeevan Rajayogam, Julian Eisenschlos, Johnson Jia, Daniel Finchelstein, Alex Yakubovich, Daniel Balle, Michael Fink, Sameer Agarwal, Jing Li, Dj~Dvijotham, Shalini Pal, Kai Kang, Jaclyn Konzelmann, Jennifer Beattie, Olivier Dousse, Diane Wu, Remi Crocker, Chen Elkind, Siddhartha~Reddy Jonnalagadda, Jong Lee, Dan Holtmann-Rice, Krystal Kallarackal, Rosanne Liu, Denis Vnukov, Neera Vats, Luca Invernizzi, Mohsen Jafari, Huanjie Zhou, Lilly Taylor, Jennifer Prendki, Marcus Wu, Tom Eccles, Tianqi Liu, Kavya Kopparapu, Francoise Beaufays, Christof Angermueller, Andreea Marzoca, Shourya Sarcar, Hilal Dib, Jeff Stanway, Frank Perbet, Nejc Trdin, Rachel Sterneck, Andrey Khorlin, Dinghua Li, Xihui Wu,
  Sonam Goenka, David Madras, Sasha Goldshtein, Willi Gierke, Tong Zhou, Yaxin Liu, Yannie Liang, Anais White, Yunjie Li, Shreya Singh, Sanaz Bahargam, Mark Epstein, Sujoy Basu, Li~Lao, Adnan Ozturel, Carl Crous, Alex Zhai, Han Lu, Zora Tung, Neeraj Gaur, Alanna Walton, Lucas Dixon, Ming Zhang, Amir Globerson, Grant Uy, Andrew Bolt, Olivia Wiles, Milad Nasr, Ilia Shumailov, Marco Selvi, Francesco Piccinno, Ricardo Aguilar, Sara McCarthy, Misha Khalman, Mrinal Shukla, Vlado Galic, John Carpenter, Kevin Villela, Haibin Zhang, Harry Richardson, James Martens, Matko Bosnjak, Shreyas~Rammohan Belle, Jeff Seibert, Mahmoud Alnahlawi, Brian McWilliams, Sankalp Singh, Annie Louis, Wen Ding, Dan Popovici, Lenin Simicich, Laura Knight, Pulkit Mehta, Nishesh Gupta, Chongyang Shi, Saaber Fatehi, Jovana Mitrovic, Alex Grills, Joseph Pagadora, Tsendsuren Munkhdalai, Dessie Petrova, Danielle Eisenbud, Zhishuai Zhang, Damion Yates, Bhavishya Mittal, Nilesh Tripuraneni, Yannis Assael, Thomas Brovelli, Prateek Jain, Mihajlo
  Velimirovic, Canfer Akbulut, Jiaqi Mu, Wolfgang Macherey, Ravin Kumar, Jun Xu, Haroon Qureshi, Gheorghe Comanici, Jeremy Wiesner, Zhitao Gong, Anton Ruddock, Matthias Bauer, Nick Felt, Anirudh GP, Anurag Arnab, Dustin Zelle, Jonas Rothfuss, Bill Rosgen, Ashish Shenoy, Bryan Seybold, Xinjian Li, Jayaram Mudigonda, Goker Erdogan, Jiawei Xia, Jiri Simsa, Andrea Michi, Yi~Yao, Christopher Yew, Steven Kan, Isaac Caswell, Carey Radebaugh, Andre Elisseeff, Pedro Valenzuela, Kay McKinney, Kim Paterson, Albert Cui, Eri Latorre-Chimoto, Solomon Kim, William Zeng, Ken Durden, Priya Ponnapalli, Tiberiu Sosea, Christopher~A. Choquette-Choo, James Manyika, Brona Robenek, Harsha Vashisht, Sebastien Pereira, Hoi Lam, Marko Velic, Denese Owusu-Afriyie, Katherine Lee, Tolga Bolukbasi, Alicia Parrish, Shawn Lu, Jane Park, Balaji Venkatraman, Alice Talbert, Lambert Rosique, Yuchung Cheng, Andrei Sozanschi, Adam Paszke, Praveen Kumar, Jessica Austin, Lu~Li, Khalid Salama, Bartek Perz, Wooyeol Kim, Nandita Dukkipati, Anthony
  Baryshnikov, Christos Kaplanis, XiangHai Sheng, Yuri Chervonyi, Caglar Unlu, Diego de~Las~Casas, Harry Askham, Kathryn Tunyasuvunakool, Felix Gimeno, Siim Poder, Chester Kwak, Matt Miecnikowski, Vahab Mirrokni, Alek Dimitriev, Aaron Parisi, Dangyi Liu, Tomy Tsai, Toby Shevlane, Christina Kouridi, Drew Garmon, Adrian Goedeckemeyer, Adam~R. Brown, Anitha Vijayakumar, Ali Elqursh, Sadegh Jazayeri, Jin Huang, Sara~Mc Carthy, Jay Hoover, Lucy Kim, Sandeep Kumar, Wei Chen, Courtney Biles, Garrett Bingham, Evan Rosen, Lisa Wang, Qijun Tan, David Engel, Francesco Pongetti, Dario de~Cesare, Dongseong Hwang, Lily Yu, Jennifer Pullman, Srini Narayanan, Kyle Levin, Siddharth Gopal, Megan Li, Asaf Aharoni, Trieu Trinh, Jessica Lo, Norman Casagrande, Roopali Vij, Loic Matthey, Bramandia Ramadhana, Austin Matthews, CJ~Carey, Matthew Johnson, Kremena Goranova, Rohin Shah, Shereen Ashraf, Kingshuk Dasgupta, Rasmus Larsen, Yicheng Wang, Manish~Reddy Vuyyuru, Chong Jiang, Joana Ijazi, Kazuki Osawa, Celine Smith, Ramya~Sree
  Boppana, Taylan Bilal, Yuma Koizumi, Ying Xu, Yasemin Altun, Nir Shabat, Ben Bariach, Alex Korchemniy, Kiam Choo, Olaf Ronneberger, Chimezie Iwuanyanwu, Shubin Zhao, David Soergel, Cho-Jui Hsieh, Irene Cai, Shariq Iqbal, Martin Sundermeyer, Zhe Chen, Elie Bursztein, Chaitanya Malaviya, Fadi Biadsy, Prakash Shroff, Inderjit Dhillon, Tejasi Latkar, Chris Dyer, Hannah Forbes, Massimo Nicosia, Vitaly Nikolaev, Somer Greene, Marin Georgiev, Pidong Wang, Nina Martin, Hanie Sedghi, John Zhang, Praseem Banzal, Doug Fritz, Vikram Rao, Xuezhi Wang, Jiageng Zhang, Viorica Patraucean, Dayou Du, Igor Mordatch, Ivan Jurin, Lewis Liu, Ayush Dubey, Abhi Mohan, Janek Nowakowski, Vlad-Doru Ion, Nan Wei, Reiko Tojo, Maria~Abi Raad, Drew~A. Hudson, Vaishakh Keshava, Shubham Agrawal, Kevin Ramirez, Zhichun Wu, Hoang Nguyen, Ji~Liu, Madhavi Sewak, Bryce Petrini, DongHyun Choi, Ivan Philips, Ziyue Wang, Ioana Bica, Ankush Garg, Jarek Wilkiewicz, Priyanka Agrawal, Xiaowei Li, Danhao Guo, Emily Xue, Naseer Shaik, Andrew Leach,
  Sadh~MNM Khan, Julia Wiesinger, Sammy Jerome, Abhishek Chakladar, Alek~Wenjiao Wang, Tina Ornduff, Folake Abu, Alireza Ghaffarkhah, Marcus Wainwright, Mario Cortes, Frederick Liu, Joshua Maynez, Andreas Terzis, Pouya Samangouei, Riham Mansour, Tomasz Kępa, François-Xavier Aubet, Anton Algymr, Dan Banica, Agoston Weisz, Andras Orban, Alexandre Senges, Ewa Andrejczuk, Mark Geller, Niccolo~Dal Santo, Valentin Anklin, Majd~Al Merey, Martin Baeuml, Trevor Strohman, Junwen Bai, Slav Petrov, Yonghui Wu, Demis Hassabis, Koray Kavukcuoglu, Jeff Dean, and Oriol Vinyals. 2024{\natexlab{a}}.
\newblock \href {https://arxiv.org/abs/2403.05530} {Gemini 1.5: Unlocking multimodal understanding across millions of tokens of context}.
\newblock \emph{Preprint}, arXiv:2403.05530.

\bibitem[{Team et~al.(2025)Team, Kamath, Ferret, Pathak, Vieillard, Merhej, Perrin, Matejovicova, Ramé, Rivière, Rouillard, Mesnard, Cideron, bastien Grill, Ramos, Yvinec, Casbon, Pot, Penchev, Liu, Visin, Kenealy, Beyer, Zhai, Tsitsulin, Busa-Fekete, Feng, Sachdeva, Coleman, Gao, Mustafa, Barr, Parisotto, Tian, Eyal, Cherry, Peter, Sinopalnikov, Bhupatiraju, Agarwal, Kazemi, Malkin, Kumar, Vilar, Brusilovsky, Luo, Steiner, Friesen, Sharma, Sharma, Gilady, Goedeckemeyer, Saade, Feng, Kolesnikov, Bendebury, Abdagic, Vadi, György, Pinto, Das, Bapna, Miech, Yang, Paterson, Shenoy, Chakrabarti, Piot, Wu, Shahriari, Petrini, Chen, Lan, Choquette-Choo, Carey, Brick, Deutsch, Eisenbud, Cattle, Cheng, Paparas, Sreepathihalli, Reid, Tran, Zelle, Noland, Huizenga, Kharitonov, Liu, Amirkhanyan, Cameron, Hashemi, Klimczak-Plucińska, Singh, Mehta, Lehri, Hazimeh, Ballantyne, Szpektor, Nardini, Pouget-Abadie, Chan, Stanton, Wieting, Lai, Orbay, Fernandez, Newlan, yeong Ji, Singh, Black, Yu, Hui, Vodrahalli, Greff, Qiu,
  Valentine, Coelho, Ritter, Hoffman, Watson, Chaturvedi, Moynihan, Ma, Babar, Noy, Byrd, Roy, Momchev, Chauhan, Sachdeva, Bunyan, Botarda, Caron, Rubenstein, Culliton, Schmid, Sessa, Xu, Stanczyk, Tafti, Shivanna, Wu, Pan, Rokni, Willoughby, Vallu, Mullins, Jerome, Smoot, Girgin, Iqbal, Reddy, Sheth, Põder, Bhatnagar, Panyam, Eiger, Zhang, Liu, Yacovone, Liechty, Kalra, Evci, Misra, Roseberry, Feinberg, Kolesnikov, Han, Kwon, Chen, Chow, Zhu, Wei, Egyed, Cotruta, Giang, Kirk, Rao, Black, Babar, Lo, Moreira, Martins, Sanseviero, Gonzalez, Gleicher, Warkentin, Mirrokni, Senter, Collins, Barral, Ghahramani, Hadsell, Matias, Sculley, Petrov, Fiedel, Shazeer, Vinyals, Dean, Hassabis, Kavukcuoglu, Farabet, Buchatskaya, Alayrac, Anil, Dmitry, Lepikhin, Borgeaud, Bachem, Joulin, Andreev, Hardin, Dadashi, and Hussenot}]{gemmateam2025gemma3technicalreport}
Gemma Team, Aishwarya Kamath, Johan Ferret, Shreya Pathak, Nino Vieillard, Ramona Merhej, Sarah Perrin, Tatiana Matejovicova, Alexandre Ramé, Morgane Rivière, Louis Rouillard, Thomas Mesnard, Geoffrey Cideron, Jean bastien Grill, Sabela Ramos, Edouard Yvinec, Michelle Casbon, Etienne Pot, Ivo Penchev, Gaël Liu, Francesco Visin, Kathleen Kenealy, Lucas Beyer, Xiaohai Zhai, Anton Tsitsulin, Robert Busa-Fekete, Alex Feng, Noveen Sachdeva, Benjamin Coleman, Yi~Gao, Basil Mustafa, Iain Barr, Emilio Parisotto, David Tian, Matan Eyal, Colin Cherry, Jan-Thorsten Peter, Danila Sinopalnikov, Surya Bhupatiraju, Rishabh Agarwal, Mehran Kazemi, Dan Malkin, Ravin Kumar, David Vilar, Idan Brusilovsky, Jiaming Luo, Andreas Steiner, Abe Friesen, Abhanshu Sharma, Abheesht Sharma, Adi~Mayrav Gilady, Adrian Goedeckemeyer, Alaa Saade, Alex Feng, Alexander Kolesnikov, Alexei Bendebury, Alvin Abdagic, Amit Vadi, András György, André~Susano Pinto, Anil Das, Ankur Bapna, Antoine Miech, Antoine Yang, Antonia Paterson, Ashish
  Shenoy, Ayan Chakrabarti, Bilal Piot, Bo~Wu, Bobak Shahriari, Bryce Petrini, Charlie Chen, Charline~Le Lan, Christopher~A. Choquette-Choo, CJ~Carey, Cormac Brick, Daniel Deutsch, Danielle Eisenbud, Dee Cattle, Derek Cheng, Dimitris Paparas, Divyashree~Shivakumar Sreepathihalli, Doug Reid, Dustin Tran, Dustin Zelle, Eric Noland, Erwin Huizenga, Eugene Kharitonov, Frederick Liu, Gagik Amirkhanyan, Glenn Cameron, Hadi Hashemi, Hanna Klimczak-Plucińska, Harman Singh, Harsh Mehta, Harshal~Tushar Lehri, Hussein Hazimeh, Ian Ballantyne, Idan Szpektor, Ivan Nardini, Jean Pouget-Abadie, Jetha Chan, Joe Stanton, John Wieting, Jonathan Lai, Jordi Orbay, Joseph Fernandez, Josh Newlan, Ju~yeong Ji, Jyotinder Singh, Kat Black, Kathy Yu, Kevin Hui, Kiran Vodrahalli, Klaus Greff, Linhai Qiu, Marcella Valentine, Marina Coelho, Marvin Ritter, Matt Hoffman, Matthew Watson, Mayank Chaturvedi, Michael Moynihan, Min Ma, Nabila Babar, Natasha Noy, Nathan Byrd, Nick Roy, Nikola Momchev, Nilay Chauhan, Noveen Sachdeva, Oskar
  Bunyan, Pankil Botarda, Paul Caron, Paul~Kishan Rubenstein, Phil Culliton, Philipp Schmid, Pier~Giuseppe Sessa, Pingmei Xu, Piotr Stanczyk, Pouya Tafti, Rakesh Shivanna, Renjie Wu, Renke Pan, Reza Rokni, Rob Willoughby, Rohith Vallu, Ryan Mullins, Sammy Jerome, Sara Smoot, Sertan Girgin, Shariq Iqbal, Shashir Reddy, Shruti Sheth, Siim Põder, Sijal Bhatnagar, Sindhu~Raghuram Panyam, Sivan Eiger, Susan Zhang, Tianqi Liu, Trevor Yacovone, Tyler Liechty, Uday Kalra, Utku Evci, Vedant Misra, Vincent Roseberry, Vlad Feinberg, Vlad Kolesnikov, Woohyun Han, Woosuk Kwon, Xi~Chen, Yinlam Chow, Yuvein Zhu, Zichuan Wei, Zoltan Egyed, Victor Cotruta, Minh Giang, Phoebe Kirk, Anand Rao, Kat Black, Nabila Babar, Jessica Lo, Erica Moreira, Luiz~Gustavo Martins, Omar Sanseviero, Lucas Gonzalez, Zach Gleicher, Tris Warkentin, Vahab Mirrokni, Evan Senter, Eli Collins, Joelle Barral, Zoubin Ghahramani, Raia Hadsell, Yossi Matias, D.~Sculley, Slav Petrov, Noah Fiedel, Noam Shazeer, Oriol Vinyals, Jeff Dean, Demis Hassabis,
  Koray Kavukcuoglu, Clement Farabet, Elena Buchatskaya, Jean-Baptiste Alayrac, Rohan Anil, Dmitry, Lepikhin, Sebastian Borgeaud, Olivier Bachem, Armand Joulin, Alek Andreev, Cassidy Hardin, Robert Dadashi, and Léonard Hussenot. 2025.
\newblock \href {https://arxiv.org/abs/2503.19786} {Gemma 3 technical report}.
\newblock \emph{Preprint}, arXiv:2503.19786.

\bibitem[{Team et~al.(2024{\natexlab{b}})Team, Riviere, Pathak, Sessa, Hardin, Bhupatiraju, Hussenot, Mesnard, Shahriari, Ramé, Ferret, Liu, Tafti, Friesen, Casbon, Ramos, Kumar, Lan, Jerome, Tsitsulin, Vieillard, Stanczyk, Girgin, Momchev, Hoffman, Thakoor, Grill, Neyshabur, Bachem, Walton, Severyn, Parrish, Ahmad, Hutchison, Abdagic, Carl, Shen, Brock, Coenen, Laforge, Paterson, Bastian, Piot, Wu, Royal, Chen, Kumar, Perry, Welty, Choquette-Choo, Sinopalnikov, Weinberger, Vijaykumar, Rogozińska, Herbison, Bandy, Wang, Noland, Moreira, Senter, Eltyshev, Visin, Rasskin, Wei, Cameron, Martins, Hashemi, Klimczak-Plucińska, Batra, Dhand, Nardini, Mein, Zhou, Svensson, Stanway, Chan, Zhou, Carrasqueira, Iljazi, Becker, Fernandez, van Amersfoort, Gordon, Lipschultz, Newlan, yeong Ji, Mohamed, Badola, Black, Millican, McDonell, Nguyen, Sodhia, Greene, Sjoesund, Usui, Sifre, Heuermann, Lago, McNealus, Soares, Kilpatrick, Dixon, Martins, Reid, Singh, Iverson, Görner, Velloso, Wirth, Davidow, Miller, Rahtz,
  Watson, Risdal, Kazemi, Moynihan, Zhang, Kahng, Park, Rahman, Khatwani, Dao, Bardoliwalla, Devanathan, Dumai, Chauhan, Wahltinez, Botarda, Barnes, Barham, Michel, Jin, Georgiev, Culliton, Kuppala, Comanescu, Merhej, Jana, Rokni, Agarwal, Mullins, Saadat, Carthy, Cogan, Perrin, Arnold, Krause, Dai, Garg, Sheth, Ronstrom, Chan, Jordan, Yu, Eccles, Hennigan, Kocisky, Doshi, Jain, Yadav, Meshram, Dharmadhikari, Barkley, Wei, Ye, Han, Kwon, Xu, Shen, Gong, Wei, Cotruta, Kirk, Rao, Giang, Peran, Warkentin, Collins, Barral, Ghahramani, Hadsell, Sculley, Banks, Dragan, Petrov, Vinyals, Dean, Hassabis, Kavukcuoglu, Farabet, Buchatskaya, Borgeaud, Fiedel, Joulin, Kenealy, Dadashi, and Andreev}]{gemmateam2024gemma2improvingopen}
Gemma Team, Morgane Riviere, Shreya Pathak, Pier~Giuseppe Sessa, Cassidy Hardin, Surya Bhupatiraju, Léonard Hussenot, Thomas Mesnard, Bobak Shahriari, Alexandre Ramé, Johan Ferret, Peter Liu, Pouya Tafti, Abe Friesen, Michelle Casbon, Sabela Ramos, Ravin Kumar, Charline~Le Lan, Sammy Jerome, Anton Tsitsulin, Nino Vieillard, Piotr Stanczyk, Sertan Girgin, Nikola Momchev, Matt Hoffman, Shantanu Thakoor, Jean-Bastien Grill, Behnam Neyshabur, Olivier Bachem, Alanna Walton, Aliaksei Severyn, Alicia Parrish, Aliya Ahmad, Allen Hutchison, Alvin Abdagic, Amanda Carl, Amy Shen, Andy Brock, Andy Coenen, Anthony Laforge, Antonia Paterson, Ben Bastian, Bilal Piot, Bo~Wu, Brandon Royal, Charlie Chen, Chintu Kumar, Chris Perry, Chris Welty, Christopher~A. Choquette-Choo, Danila Sinopalnikov, David Weinberger, Dimple Vijaykumar, Dominika Rogozińska, Dustin Herbison, Elisa Bandy, Emma Wang, Eric Noland, Erica Moreira, Evan Senter, Evgenii Eltyshev, Francesco Visin, Gabriel Rasskin, Gary Wei, Glenn Cameron, Gus Martins,
  Hadi Hashemi, Hanna Klimczak-Plucińska, Harleen Batra, Harsh Dhand, Ivan Nardini, Jacinda Mein, Jack Zhou, James Svensson, Jeff Stanway, Jetha Chan, Jin~Peng Zhou, Joana Carrasqueira, Joana Iljazi, Jocelyn Becker, Joe Fernandez, Joost van Amersfoort, Josh Gordon, Josh Lipschultz, Josh Newlan, Ju~yeong Ji, Kareem Mohamed, Kartikeya Badola, Kat Black, Katie Millican, Keelin McDonell, Kelvin Nguyen, Kiranbir Sodhia, Kish Greene, Lars~Lowe Sjoesund, Lauren Usui, Laurent Sifre, Lena Heuermann, Leticia Lago, Lilly McNealus, Livio~Baldini Soares, Logan Kilpatrick, Lucas Dixon, Luciano Martins, Machel Reid, Manvinder Singh, Mark Iverson, Martin Görner, Mat Velloso, Mateo Wirth, Matt Davidow, Matt Miller, Matthew Rahtz, Matthew Watson, Meg Risdal, Mehran Kazemi, Michael Moynihan, Ming Zhang, Minsuk Kahng, Minwoo Park, Mofi Rahman, Mohit Khatwani, Natalie Dao, Nenshad Bardoliwalla, Nesh Devanathan, Neta Dumai, Nilay Chauhan, Oscar Wahltinez, Pankil Botarda, Parker Barnes, Paul Barham, Paul Michel, Pengchong Jin,
  Petko Georgiev, Phil Culliton, Pradeep Kuppala, Ramona Comanescu, Ramona Merhej, Reena Jana, Reza~Ardeshir Rokni, Rishabh Agarwal, Ryan Mullins, Samaneh Saadat, Sara~Mc Carthy, Sarah Cogan, Sarah Perrin, Sébastien M.~R. Arnold, Sebastian Krause, Shengyang Dai, Shruti Garg, Shruti Sheth, Sue Ronstrom, Susan Chan, Timothy Jordan, Ting Yu, Tom Eccles, Tom Hennigan, Tomas Kocisky, Tulsee Doshi, Vihan Jain, Vikas Yadav, Vilobh Meshram, Vishal Dharmadhikari, Warren Barkley, Wei Wei, Wenming Ye, Woohyun Han, Woosuk Kwon, Xiang Xu, Zhe Shen, Zhitao Gong, Zichuan Wei, Victor Cotruta, Phoebe Kirk, Anand Rao, Minh Giang, Ludovic Peran, Tris Warkentin, Eli Collins, Joelle Barral, Zoubin Ghahramani, Raia Hadsell, D.~Sculley, Jeanine Banks, Anca Dragan, Slav Petrov, Oriol Vinyals, Jeff Dean, Demis Hassabis, Koray Kavukcuoglu, Clement Farabet, Elena Buchatskaya, Sebastian Borgeaud, Noah Fiedel, Armand Joulin, Kathleen Kenealy, Robert Dadashi, and Alek Andreev. 2024{\natexlab{b}}.
\newblock \href {https://arxiv.org/abs/2408.00118} {Gemma 2: Improving open language models at a practical size}.
\newblock \emph{Preprint}, arXiv:2408.00118.

\bibitem[{Tupas and Martin(2017)}]{tupas2017bilingual}
Ruanni Tupas and Isabel~Pefianco Martin. 2017.
\newblock Bilingual and mother tongue-based multilingual education in the {P}hilippines.
\newblock \emph{Bilingual and multilingual education}, 10:247--258.

\bibitem[{Villavicencio et~al.(2021)Villavicencio, Macrohon, Inbaraj, Jeng, and Hsieh}]{villavicencio2021twitter}
Charlyn Villavicencio, Julio~Jerison Macrohon, X~Alphonse Inbaraj, Jyh-Horng Jeng, and Jer-Guang Hsieh. 2021.
\newblock Twitter sentiment analysis towards {COVID}-19 vaccines in the {P}hilippines using {N}aive {B}ayes.
\newblock \emph{Information}, 12(5):204.

\bibitem[{Visweswariah et~al.(2011)Visweswariah, Rajkumar, Gandhe, Ramanathan, and Navr{\'a}til}]{visweswariah2011word}
Karthik Visweswariah, Rajakrishnan Rajkumar, Ankur Gandhe, Ananthakrishnan Ramanathan, and Ji{\v{r}}{\'\i} Navr{\'a}til. 2011.
\newblock A word reordering model for improved machine translation.
\newblock In \emph{Proceedings of the 2011 Conference on Empirical Methods in Natural Language Processing}, pages 486--496.

\bibitem[{Wang et~al.(2024)Wang, Liu, Huang, Jiao, Ding, Aw, and Chen}]{wang-etal-2024-seaeval}
Bin Wang, Zhengyuan Liu, Xin Huang, Fangkai Jiao, Yang Ding, AiTi Aw, and Nancy Chen. 2024.
\newblock \href {https://doi.org/10.18653/v1/2024.naacl-long.22} {{S}ea{E}val for multilingual foundation models: From cross-lingual alignment to cultural reasoning}.
\newblock In \emph{Proceedings of the 2024 Conference of the North American Chapter of the Association for Computational Linguistics: Human Language Technologies (Volume 1: Long Papers)}, pages 370--390, Mexico City, Mexico. Association for Computational Linguistics.

\bibitem[{Wardana et~al.(2022)Wardana, Widayati, Taib, and Iqbal}]{wardana2022lexicostatistics}
Muhammad~Kiki Wardana, Dwi Widayati, Rostina Taib, and Muhammad Iqbal. 2022.
\newblock Lexicostatistics of {M}alay, {T}agalog and {I}locano languages: A {C}omparisonal {H}istorical {L}inguistic {S}tudy.
\newblock \emph{Jurnal Education and Development}, 10(3):475--479.

\bibitem[{Wey(2024)}]{brickscurrent}
Tristan Koh~Ly Wey. 2024.
\newblock \href {https://aisingapore.org/ai-governance/current-llm-evaluations-do-not-sufficiently-measure-all-we-need/} {Current {LLM} evaluations do not sufficiently measure all we need}.

\bibitem[{Wilie et~al.(2020)Wilie, Vincentio, Winata, Cahyawijaya, Li, Lim, Soleman, Mahendra, Fung, Bahar, and Purwarianti}]{wilie-etal-2020-indonlu}
Bryan Wilie, Karissa Vincentio, Genta~Indra Winata, Samuel Cahyawijaya, Xiaohong Li, Zhi~Yuan Lim, Sidik Soleman, Rahmad Mahendra, Pascale Fung, Syafri Bahar, and Ayu Purwarianti. 2020.
\newblock \href {https://doi.org/10.18653/v1/2020.aacl-main.85} {{I}ndo{NLU}: {B}enchmark and {R}esources for {E}valuating {I}ndonesian {N}atural {L}anguage {U}nderstanding}.
\newblock In \emph{Proceedings of the 1st Conference of the Asia-Pacific Chapter of the Association for Computational Linguistics and the 10th International Joint Conference on Natural Language Processing}, pages 843--857, Suzhou, China. Association for Computational Linguistics.

\bibitem[{Wolff(2001)}]{wolff2001influence}
John~U Wolff. 2001.
\newblock The influence of {S}panish on {T}agalog.
\newblock \emph{Propio y lo ajeno en las lenguas austron{\'e}sicas y amerindias: procesos interculturales en el contacto de lenguas ind{\'\i}genas con el espa{\~n}ol en el Pac{\'\i}fico e Hispanoam{\'e}rica}, pages 233--252.

\bibitem[{Yang et~al.(2024)Yang, Yang, Hui, Zheng, Yu, Zhou, Li, Li, Liu, Huang, Dong, Wei, Lin, Tang, Wang, Yang, Tu, Zhang, Ma, Yang, Xu, Zhou, Bai, He, Lin, Dang, Lu, Chen, Yang, Li, Xue, Ni, Zhang, Wang, Peng, Men, Gao, Lin, Wang, Bai, Tan, Zhu, Li, Liu, Ge, Deng, Zhou, Ren, Zhang, Wei, Ren, Liu, Fan, Yao, Zhang, Wan, Chu, Liu, Cui, Zhang, Guo, and Fan}]{yang2024qwen2technicalreport}
An~Yang, Baosong Yang, Binyuan Hui, Bo~Zheng, Bowen Yu, Chang Zhou, Chengpeng Li, Chengyuan Li, Dayiheng Liu, Fei Huang, Guanting Dong, Haoran Wei, Huan Lin, Jialong Tang, Jialin Wang, Jian Yang, Jianhong Tu, Jianwei Zhang, Jianxin Ma, Jianxin Yang, Jin Xu, Jingren Zhou, Jinze Bai, Jinzheng He, Junyang Lin, Kai Dang, Keming Lu, Keqin Chen, Kexin Yang, Mei Li, Mingfeng Xue, Na~Ni, Pei Zhang, Peng Wang, Ru~Peng, Rui Men, Ruize Gao, Runji Lin, Shijie Wang, Shuai Bai, Sinan Tan, Tianhang Zhu, Tianhao Li, Tianyu Liu, Wenbin Ge, Xiaodong Deng, Xiaohuan Zhou, Xingzhang Ren, Xinyu Zhang, Xipin Wei, Xuancheng Ren, Xuejing Liu, Yang Fan, Yang Yao, Yichang Zhang, Yu~Wan, Yunfei Chu, Yuqiong Liu, Zeyu Cui, Zhenru Zhang, Zhifang Guo, and Zhihao Fan. 2024.
\newblock \href {https://arxiv.org/abs/2407.10671} {Qwen2 technical report}.
\newblock \emph{Preprint}, arXiv:2407.10671.

\bibitem[{Yang et~al.(2023)Yang, Chiang, Zheng, Gonzalez, and Stoica}]{yang2023rethinking}
Shuo Yang, Wei-Lin Chiang, Lianmin Zheng, Joseph~E Gonzalez, and Ion Stoica. 2023.
\newblock Rethinking benchmark and contamination for language models with rephrased samples.
\newblock \emph{arXiv preprint arXiv:2311.04850}.

\bibitem[{Yapan(2017)}]{yapan2017bagay}
Alvin Yapan. 2017.
\newblock \emph{Bagay: {G}abay sa pagsulat sa wikang {F}ilipino}.
\newblock BlueBooks.

\bibitem[{Zamar(2022)}]{zamar2022filipino}
Maria~Sheila Zamar. 2022.
\newblock \emph{Filipino: An {E}ssential {G}rammar}.
\newblock Routledge.

\bibitem[{Zhang et~al.(2020)Zhang, Kishore, Wu, Weinberger, and Artzi}]{bert-score}
Tianyi Zhang, Varsha Kishore, Felix Wu, Kilian~Q. Weinberger, and Yoav Artzi. 2020.
\newblock \href {https://openreview.net/forum?id=SkeHuCVFDr} {{BERTScore}: {E}valuating {T}ext {G}eneration with {BERT}}.
\newblock In \emph{Proceedings of the 8th International Conference on Learning Representations (ICLR 2020)}, Addis Ababa, Ethiopia.

\bibitem[{Zhang et~al.(2024)Zhang, Chan, Zhao, Aljunied, Wang, Liu, Deng, Hu, Xu, Chia, Li, and Bing}]{zhang2024seallms3openfoundation}
Wenxuan Zhang, Hou~Pong Chan, Yiran Zhao, Mahani Aljunied, Jianyu Wang, Chaoqun Liu, Yue Deng, Zhiqiang Hu, Weiwen Xu, Yew~Ken Chia, Xin Li, and Lidong Bing. 2024.
\newblock \href {https://arxiv.org/abs/2407.19672} {Seallms 3: Open foundation and chat multilingual large language models for southeast asian languages}.
\newblock \emph{Preprint}, arXiv:2407.19672.

\bibitem[{Zhang et~al.(2019)Zhang, Baldridge, and He}]{paws2019naacl}
Yuan Zhang, Jason Baldridge, and Luheng He. 2019.
\newblock \href {https://doi.org/10.18653/v1/N19-1131} {{PAWS}: Paraphrase adversaries from word scrambling}.
\newblock In \emph{Proceedings of the 2019 Conference of the North {A}merican Chapter of the Association for Computational Linguistics: Human Language Technologies, Volume 1 (Long and Short Papers)}, pages 1298--1308, Minneapolis, Minnesota. Association for Computational Linguistics.

\bibitem[{Zhao et~al.(2025)Zhao, Liu, Deng, Ying, Aljunied, Li, Bing, Chan, Rong, Zhao, and Zhang}]{zhao2025babelopenmultilinguallarge}
Yiran Zhao, Chaoqun Liu, Yue Deng, Jiahao Ying, Mahani Aljunied, Zhaodonghui Li, Lidong Bing, Hou~Pong Chan, Yu~Rong, Deli Zhao, and Wenxuan Zhang. 2025.
\newblock \href {https://arxiv.org/abs/2503.00865} {Babel: Open multilingual large language models serving over 90\% of global speakers}.
\newblock \emph{Preprint}, arXiv:2503.00865.

\end{thebibliography}

\appendix


\newpage
\onecolumn
\section{Overview of Tasks and Datasets}
\label{sec:dataset_overview}
As discussed in Section 3 of this paper, the selection of the eight tasks are presented below. We also provide Table \ref{tab:dataset_sources_licenses} describe the source datasets used for these tasks. The modifications and adaptations that were applied and the usage of the datasets for evaluation comply with their original intended uses.

\textbf{Abstractive Summarization (AS).} An LLM performing this task is given a paragraph and  is expected to summarize the content in a sentence. The model is tested not only for identifying the salient points of the text, but also paraphrasing the content into a concise and coherent text. For this task, we use XL-Sum \cite{hasan-etal-2021-xl}, a collection of annotated article-summary pairs.

\textbf{Causal Reasoning (CR).} This task requires the LLM to understand the relationship between events. In particular, the model is given a premise, a set of statements, and an instruction to determine which of the provided statements is the cause or effect of the premise. We employed Balanced COPA \cite{kavumba-etal-2019-choosing}, a dataset designed to evaluate commonsense causal reasoning with paired alternatives.

\textbf{Machine Translation (MT).} For this task, an LLM is given a text in one language, and is expected to provide the equivalent text translated into another language. In this study, we test for both English→Filipino and Filipino→English translation. This task leveraged the Filipino subset of FLORES 200 \cite{nllb2022}, which includes translations across numerous languages and domains.

\textbf{Natural Language Inference (NLI).} This is classification task where an LLM is provided two sentences (X and Y), and is expected to determine whether the sentences are related in one in the following ways: (a) X implies Y; (b) X contradicts Y; and (c) X neither implies nor contradicts Y.  This task utilized XNLI \cite{conneau2018xnli}, a dataset containing human-annotated examples for evaluating cross-lingual inference.

\textbf{Paraphrase Identification (PI).} This task requires an LLM to determine if two provided texts are paraphrased versions of each other; that is, whether both pieces of text convey the same idea. For this task, we used PAWS \cite{paws2019naacl}, which contains paraphrase and non-paraphrase pairs with high lexical overlap. 

\textbf{Question Answering (QA).} Given a passage and a question, an LLM performing this task must provide a span from the passage that answers the question. In this study, QA utilized Belebele \cite{bandarkar-etal-2024-belebele}, a multiple-choice reading comprehension dataset designed to assess understanding of passages.

\textbf{Toxicity Detection (TD) and Sentiment Analysis (SA).} These two NLU tasks both require an LLM to analyze a natural language text. The TD task requires the model to identify whether hate speech and abusive language is used in the text, while the SA task requires the model to classify the text as either positive, negative, or neutral in sentiment polarity. Both tasks are derived from the Philippine election-related tweets dataset \cite{cabasag2019hate}, which provided a resource for toxicity in political discourse. These tweets were collected during the 2016 presidential campaign. This dataset was also relabeled for sentiment analysis by three native Filipino speakers. There is substantial agreement between the annotations (Cohen’s kappa of 0.8202, Krippendorf’s alpha of 0.8268). 

\renewcommand{\arraystretch}{1.1}
\begin{table*}[htp]
\centering
\small
\begin{tabular}{llll}
\toprule
\textbf{Competency} & \textbf{Task} & \textbf{Dataset} & \textbf{License} \\
\midrule
\multirow{1}{*}{NLU} & PI & PAWS \cite{paws2019naacl} & CC BY 4.0 \\
& QA & Belebele \cite{bandarkar-etal-2024-belebele} & CC BY-NC 4.0 \\
& SA & PH Elections \cite{cabasag2019hate} & Unknown\\
& TD & PH Elections \cite{cabasag2019hate} & Unknown \\
\midrule
\multirow{1}{*}{NLR} & CR & Balanced COPA \cite{kavumba-etal-2019-choosing} & CC BY 4.0 \\
& NLI & XNLI \cite{conneau2018xnli} & CC BY-NC 4.0 \\
\midrule
\multirow{1}{*}{NLG} & AS & XL-Sum \cite{hasan-etal-2021-xl} & CC BY-NC-SA 4.0 \\
& MT & FLORES 200 \cite{nllb2022} & CC BY-SA 4.0 \\
\bottomrule
\end{tabular}
\caption{License details for datasets used in \textsc{Batayan}.}
\label{tab:dataset_sources_licenses}
\end{table*}
\renewcommand{\arraystretch}{1.0}

\renewcommand{\arraystretch}{1.2}
\begin{table*}[htp]
    \centering
    \small
    \begin{tabular}{llrr}
        \toprule
        \textbf{Competency} & \textbf{Task} & \textbf{\# test samples} & \textbf{Avg. No. Words/Sample} \\
        \midrule
        \multirow{1}{*}{NLU} & PI  & 400  & 22.50 (Sentence 1), 22.47 (Sentence 2) \\
        & QA  & 100  & 14.75 (Question), 4.26 (Choices) \\
        & SA  & 600  & 24.32 \\
        & TD  & 400  & 20.73 \\
        \midrule
        \multirow{1}{*}{NLR} & CR  & 400  & 7.24 (Premise), 6.05 (Choices) \\
        & NLI  & 600  & 23.30 (Sentence 1), 12.20 (Sentence 2) \\
        \midrule
        \multirow{1}{*}{NLG} & AS  & 100  & 120.8 \\
        & MT (English Text) & 600  & 21.42 \\
        & MT (Tagalog Text) & 600  & 25.05 \\
        \bottomrule
    \end{tabular}
    \caption{Summary of dataset statistics, including total rows and average words per text.}
    \label{tab:dataset_summary}
\end{table*}
\renewcommand{\arraystretch}{1.0}

Table \ref{tab:dataset_summary} presents quantitative statistics on dataset size and average word counts. The variation in text length reflects task-specific requirements: AS involves longer passages, while classification tasks (SA, TD, and PI) typically contain more concise text samples. The QA dataset, in particular, has relatively short questions and answer choices, mirroring real-world multiple-choice assessments. Further granularity is provided in Table \ref{tab:as_details}, which breaks down sentence, word, and character-level statistics. This statistics shows structural differences between tasks; for instance, MT has sentence-level parallelism, while CR and NLI involve distinct premise-hypothesis or question-choice relationships.

\renewcommand{\arraystretch}{1.2}
\begin{table*}[ht]
    \centering
    \small
    \begin{tabular}{lllrrr}
        \toprule
        \textbf{Competency} & \textbf{Task} & \textbf{Component} & \textbf{Avg. No. Sentences} & \textbf{Avg. No. Words} & \textbf{Avg. No. Characters} \\
        \midrule
        \multirow{1}{*}{NLU} & PI & Sentence 1 & 1.0 & 22.50 & 123.01 \\
        & PI & Sentence 2 & 1.0 & 22.47 & 122.99 \\
        & QA & Question & 1.0 & 14.75 & 80.62 \\
        & QA & Choice 1 & 1.0 & 4.66 & 29.10 \\
        & QA & Choice 2 & 1.0 & 4.61 & 28.60 \\
        & QA & Choice 3 & 1.0 & 4.76 & 29.74 \\
        & QA & Choice 4 & 1.0 & 3.0 & 28.58 \\
        & SA & Text & 2.21 & 24.32 & 120.44 \\
        & TD & Text & 2.45 & 20.73 & 99.37 \\
        \midrule
        \multirow{1}{*}{NLR} & CR & Premise & 1.0 & 7.24 & 37.55 \\
        & CR & Choice 1 & 1.0 & 6.05 & 30.91 \\
        & CR & Choice 2 & 1.0 & 6.05 & 30.62 \\
        & NLI & Sentence 1 & 1.0 & 23.30 & 127.88 \\
        & NLI & Sentence 2 & 1.0 & 12.20 & 67.04 \\
        \midrule
        \multirow{1}{*}{NLG} & AS & Text & 5.43 & 120.8 & 666.18 \\
        & MT & English Text & 1.12 & 21.42 & 113.07 \\
        & MT & Filipino Text & 1.12 & 25.05 & 140.58 \\
        \bottomrule
    \end{tabular}
    \caption{Detailed dataset statistics, including sentence, word, and character counts.}
    \label{tab:as_details}
\end{table*}
\renewcommand{\arraystretch}{1.0}

\newpage
\onecolumn
\section{Dataset Examples}
\label{sec:dataset_examples}
\begin{table}[h]
    \centering
    \small
    \renewcommand{\arraystretch}{1.4}
    \begin{tabular}{p{2cm} p{1cm} p{12cm}}
        \toprule
        \textbf{Competency} & \textbf{Task} & \textbf{Example} \\
        \midrule
        NLU & PI & \textbf{Sentence 1:} \textit{Wala pang isang taon matapos ang kaniyang nagdaang kasal, pinakasalan din ni Charlemagne si Desiderata at tinanggihan ang 13-anyos na Swabian na nagngangalang Hildegard.} \\ 
        & & \textbf{Sentence 2:} \textit{Pinakasalan ni Charlemagne si Desiderata wala pang isang taon pagkatapos ng kanyang kasal at isinawalang-bahala ang isang 13 taong gulang na Swabian na nagngangalang Hildegard.} \\
        & & \textbf{Label:} \textit{Paraprase} (Paraphrase) \\ \cline{3-3}
        & QA & \textbf{Text:} \textit{Lumipat ang mga hukbong Coalition at Afghan sa lugar na iyon upang tiyaking ligtas ang lokasyon, at nagpadala ng iba pang eroplano ng koalisyon upang tumulong. Naganap ang pagbagsak sa mataas na bahagi ng kabundukan, at pinaniniwalaang resulta ng pagbabaril ng kalaban. Isang malaking hamon ang masamang panahon at malubak na daan sa paghahanap.} \\ 
        & & \textbf{Question:} \textit{Ano ang pinaniniwalaang dahilan ng pagbagsak?} \\ 
        & & \textbf{Choices:} (1) \textit{Pangit na daan} (2) \textit{Masamang sunog} (3) \textit{Mabundok na lupain} (4) \textit{Masamang lagay ng panahon} \\
        & & \textbf{Label:} 1 \\ \cline{3-3}
        & SA & \textbf{Text:} \textit{Pucha.. PURO DILAWAN DITO SA REDDIT AH HAHAHA.. AKALA NYO NAMAN ANG LAKI NA NG ACCOMPLISHMENT NYO DAHIL SA VIDEO NA YAN MGA GUNGGONG!! HAHAHA \#DU30 PA RIN!! MGA TAE KAYO!} \\ 
        & & \textbf{Label:} \textit{Negatibo} (Negative) \\ \cline{3-3}
        & TD & \textbf{Text:} \textit{Ayun, nag-Filipino rin si Poe, ang presidente ko! Kitang-kita ang kanyang platapormang maka-mahirap at maka-tao. \#PiliPinasDebates2016} \\ 
        & & \textbf{Label:} \textit{Malinis} (Clean) \\
        \midrule
        NLR & CR & \textbf{Text:} \textit{Hindi tinanggap ang tsekeng ginawa ko.} \\ 
        & & \textbf{Question:} \textit{Sanhi} (Cause) \\ 
        & & \textbf{Choices:} (0) \textit{Walang laman ang bank account ko.} (1) \textit{Tumaas ang aking suweldo.} \\
        & & \textbf{Label:} 0 \\ \cline{3-3}
        & NLI & \textbf{Sentence 1:} \textit{Di mo ba natatandaan? Pupunta tayo ngayong araw sa birthday ni Tita Basia.} \\ 
        & & \textbf{Sentence 2:} \textit{Hindi kami pupunta sa birthday party ni Tita Basia ngayon.} \\ 
        & & \textbf{Label:} Contradiction \\
        \midrule
        NLG & AS & \textbf{Article:} \textit{Kinumpirma noong nakaraang buwan na humawa na ang virus sa isang tupa sa isla matapos isilang nang patay at wala sa hugis ang limang kordero sa isang bukid. Sinabi ng state veterinary officer na malamang na ang birus ay sanhi ng windborne midges.} \\ 
        & & \textbf{Summary:} \textit{Kinumpirma ng mga pagsusuri noong nakaraang buwan na nahawahan ng Schmallenberg virus ang mga tupa sa isla.} \\ \cline{3-3}
        & MT & \textbf{Text:} Former U.S. Speaker of the House Newt Gingrich came in second with 32 percent. \\ 
        & & \textbf{Translation:} \textit{Pumangalawa ang dating Speaker of the House ng U.S. na si Newt Gingrich nang may 32 porsyento.} \\
        \bottomrule
    \end{tabular}
    \caption{Example instances for each task in \textsc{Batayan}.}
    \label{tab:dataset_examples}
\end{table}

\onecolumn
\section{Inter-rater Agreement}
\label{sec:agreement}

\renewcommand{\arraystretch}{1.2}
\begin{table*}[ht]
\centering
\small
\begin{tabular}{llllrr}
\toprule
\textbf{Competency} & \textbf{Task} & \textbf{Dataset} & \textbf{Criteria} & \textbf{Joint agreement} & \textbf{Rating} \\
\midrule
\multirow{1}{*}{NLU} & \multicolumn{1}{l}{\multirow{1}{*}{PI}} & \multicolumn{1}{l}{\multirow{1}{*}{PAWS}} & Completeness & 0.9550 & - \\
& \multicolumn{1}{l}{} & \multicolumn{1}{l}{} & Fluency & 0.5875 & - \\
& \multicolumn{1}{l}{} & \multicolumn{1}{l}{} & Sensibility & 0.8025 & - \\
& \multicolumn{1}{l}{\multirow{1}{*}{QA}} & \multicolumn{1}{l}{\multirow{1}{*}{Belebele}} & Completeness & 0.9700 & - \\
& \multicolumn{1}{l}{} & \multicolumn{1}{l}{} & Fluency & 0.8300 & - \\
& \multicolumn{1}{l}{} & \multicolumn{1}{l}{} & Sensibility & 0.9500 & - \\
& \multicolumn{1}{l}{\multirow{1}{*}{SA}} & \multicolumn{1}{l}{\multirow{1}{*}{PH Election Tweets}} & Completeness & 0.6217  & - \\
& \multicolumn{1}{l}{} & \multicolumn{1}{l}{} & Fluency & 0.6767  & - \\
& \multicolumn{1}{l}{} & \multicolumn{1}{l}{} & Sensibility & 0.7050  & - \\
& \multicolumn{1}{l}{\multirow{1}{*}{TD}} & \multicolumn{1}{l}{\multirow{1}{*}{PH Election Tweets}} & Completeness & 0.8310  & - \\
& \multicolumn{1}{l}{} & \multicolumn{1}{l}{} & Fluency & 0.9060  & - \\
& \multicolumn{1}{l}{} & \multicolumn{1}{l}{} & Sensibility & 0.8680 & - \\
\midrule
\multirow{1}{*}{NLR} & \multicolumn{1}{l}{\multirow{1}{*}{CR}} & \multicolumn{1}{l}{\multirow{1}{*}{Balanced COPA}}  & Completeness & 0.8975 & - \\
& \multicolumn{1}{l}{} & \multicolumn{1}{l}{}  & Fluency & 0.7275 & - \\
& \multicolumn{1}{l}{} & \multicolumn{1}{l}{}  & Sensibility & 0.9575 & - \\
& \multicolumn{1}{l}{\multirow{1}{*}{NLI}} & \multicolumn{1}{l}{\multirow{1}{*}{XNLI}} & Completeness & 0.9716 & - \\
& \multicolumn{1}{l}{} & \multicolumn{1}{l}{} & Fluency & 0.8433 & - \\
& \multicolumn{1}{l}{} & \multicolumn{1}{l}{} & Sensibility & 0.9933 & - \\
\midrule
\multirow{1}{*}{NLG} & \multicolumn{1}{l}{\multirow{1}{*}{AS}} & \multicolumn{1}{l}{\multirow{1}{*}{XL-Sum}}  & Completeness & 0.8800 & - \\
& \multicolumn{1}{l}{} & \multicolumn{1}{l}{}  & Fluency & 0.9600 & - \\
& \multicolumn{1}{l}{} & \multicolumn{1}{l}{}  & Faithfulness & 1.0000 & - \\
& \multicolumn{1}{l}{} & \multicolumn{1}{l}{}  & Relevance of summary & - & 1.99 \\
& \multicolumn{1}{l}{} & \multicolumn{1}{l}{}  & Fluency of summary & - & 2.64\\
& \multicolumn{1}{l}{} & \multicolumn{1}{l}{}  & Coherence of summary & - & 2.56 \\
& \multicolumn{1}{l}{\multirow{1}{*}{MT}} & \multicolumn{1}{l}{\multirow{1}{*}{FLORES 200}} & Completeness & 0.7100 & - \\
& \multicolumn{1}{l}{} & \multicolumn{1}{l}{} & Fluency & 0.9900 & - \\
& \multicolumn{1}{l}{} & \multicolumn{1}{l}{} & Sensibility & 0.7750 & - \\
\bottomrule
\end{tabular}
\caption{Joint agreement is calculated as the percentage of times all the raters unanimously agree that the samples fulfill the criteria. Rating is calculated as the average score assigned by the raters to the samples under the given Likert-scaled (0-3) criteria.}
\label{tab:agreement}
\end{table*}
\renewcommand{\arraystretch}{1.0}

\newpage
\onecolumn
\section{Prompt Templates}
\label{sec:prompt_templates}

\renewcommand{\arraystretch}{1.2}
\small
\begin{longtable}{p{0.5cm} p{7.0cm} p{7.0cm}}
\toprule
\textbf{Task} & \textbf{Filipino Prompt Template} & \textbf{SEA-HELM English Prompt Template}
\\
\midrule
PI & 
\begin{lstlisting}
Bibigyan ka ng dalawang pangungusap,
  SENTENCE_1 at SENTENCE_2. Tukuyin kung
  alin sa sumusunod na pahayag ang pinaka-
  angkop para sa SENTENCE_1 at SENTENCE_2.
A: Paraprase ang SENTENCE_2 ng SENTENCE_1.
B: Hindi paraprase ang SENTENCE_2 ng
  SENTENCE_1.

Sumagot gamit ang sumusunod na format.
Sagot: $OPTION
Palitan ang $OPTION ng napiling sagot.
  Gumamit lang ng titik A o B sa sagot mo.
  {fewshot_examples}

SENTENCE_1:
```
{sentence1}
```
SENTENCE_2:
```
{sentence2}
```
\end{lstlisting}
&
\begin{lstlisting}
You will be given two sentences, SENTENCE_1
  and SENTENCE_2. Determine which of the
  following statements applies to
  SENTENCE_1 and SENTENCE_2 the best.
A: SENTENCE_2 is a paraphrase of SENTENCE_1.
B: SENTENCE_2 is not a paraphrase of
  SENTENCE_1.

Answer only using the following format:
Answer: $OPTION
Replace $OPTION with the selected option.
  Use the letters A or B only as the answer.
  {fewshot_examples}

SENTENCE_1:
```
{sentence1}
```
SENTENCE_2:
```
{sentence2}
```
\end{lstlisting}
\\
\midrule
QA & 
\begin{lstlisting}
Bibigyan ka ng isang talata, isang tanong,
  at apat na pagpipiliang sagot. Sumagot
  base sa talata sa pamamagitan ng pagpili
  ng isa sa mga opsiyong ibinigay.

Sumagot gamit ang sumusunod na format:
Sagot: $OPTION
Palitan ang $OPTION ng napiling sagot.
  Gumamit lang ng titik A, B, C, o D sa
  sagot mo.{fewshot_examples}

Talata:
```
{text}
```
Tanong: {question}
A: {choice1}
B: {choice2}
C: {choice3}
D: {choice4}
\end{lstlisting}
&
\begin{lstlisting}
You will be given one paragraph, one
  question and four answer options. Based
  on the paragraph, answer by choosing one
  of the options given.

Answer only using the following format:
Answer: $OPTION
Replace $OPTION with the selected option.
  Use the letter A, B, C, or D as the
  answer.{fewshot_examples}

Paragraph:
```
{text}
```
Question: {question}
A: {choice1}
B: {choice2}
C: {choice3}
D: {choice4}
\end{lstlisting}
\\
\midrule
SA & 
\begin{lstlisting}
Ano ang sentimyento sa sumusunod na
  pangungusap? Sumagot gamit ng isa sa mga
  sumusunod na pagpipilian: Positibo,
  Negatibo, o Neutral. 

Sumagot gamit ang sumusunod na format:
Sagot: $OPTION
Palitan ang $OPTION ng napiling sagot.
  {fewshot_examples}

Pangungusap:
```
{text}
```
\end{lstlisting}
&
\begin{lstlisting}
What is the sentiment of the following 
  sentence? Use one of the choices below:
  Positive, Negative, or Neutral.

Answer only using the following format:
Answer: $OPTION
Replace $OPTION with the selected option.
  {fewshot_examples}

Sentence:
```
{text}
```
\end{lstlisting}
\\
\midrule
TD & 
\begin{lstlisting}
Isa kang toxicity detector. Nangangahulugan 
  ang mga sumusunod na label:
Malinis: Walang toxicity.
Mapoot: May toxicity at direktang pag-atake
  sa isang tao.  

Batay sa kahulugan ng mga label, iuri ang
  mga sumusunod na pangungusap gamit lang
  ng isang salita: Malinis o Mapoot. 

Sumagot gamit ang sumusunod na format:
Sagot: $LABEL
Palitan ang $LABEL ng napiling label.
  {fewshot_examples}

Pangungusap:
```
{text}
```
\end{lstlisting}
&
\begin{lstlisting}
You are a toxicity detector. The following
  labels are defined as follows:
Clean: There is no toxicity.
Toxic: There is toxicity and direct attacks
  towards a person. 

Based on the label definitions, classify the
  following sentence only using one word:
  Clean or Toxic. 

Answer only using the following format:
Answer: $LABEL
Replace $LABEL with the selected label.
  {fewshot_examples}

Sentence:
```
{text}
```
\end{lstlisting}
\\
\midrule
CR & 
\begin{lstlisting}
Sumagot gamit ang sumusunod na format:
Sagot: $OPTION
Palitan ang $OPTION ng napiling sagot.
  Gumamit lang ng titik A or B sa sagot mo.
  {fewshot_examples}

Batay sa ibibigay na sitwasyon, alin sa 
  sumusunod na pagpipilian ang mas maaari
  na {sanhi/bunga}?

Sitwasyon:
```
{text}
```
Piliin ang pinaka-angkop na sagot mula sa
  sumusunod na pagpipilian:
A: {choice1}
B: {choice2}
\end{lstlisting}
&
\begin{lstlisting}
Answer only using the following format:
Answer: $OPTION
Replace $OPTION with the selected option.
  Use only the letters A or B as the answer.
  {fewshot_examples}

Based on the given situation, which of the
  following options is more likely to be the
  {cause/effect}?

Situation:
```
{text}
```
Choose the best answer from the following
  options:
A: {choice1}
B: {choice2}
\end{lstlisting}
\\
\midrule
NLI & 
\begin{lstlisting}
Bibigyan ka ng dalawang pangungusap,
  SENTENCE_1 at SENTENCE_2. Tukuyin kung
  alin sa sumusunod na pahayag ang pinaka-
  angkop para sa SENTENCE_1 at SENTENCE_2.
A: Kung totoo ang SENTENCE_1, dapat totoo
  din ang SENTENCE_2.
B: Sumasalungat ang SENTENCE_1 sa
  SENTENCE_2.
C: Kapag totoo ang SENTENCE_1, pwedeng totoo
  o hindi totoo ang SENTENCE_2. 

Sumagot gamit ang sumusunod na format.
Sagot: $OPTION
Palitan ang $OPTION ng napiling sagot.
  Gumamit lang ng titik A, B, o C sa sagot
  mo.{fewshot_examples}

SENTENCE_1:
```
{sentence1}
```
SENTENCE_2:
```
{sentence2}
```
\end{lstlisting}
&
\begin{lstlisting}
You will be given two sentences, SENTENCE_1
  and SENTENCE_2. Determine which of the
  following statements applies to
  SENTENCE_1 and SENTENCE_2 the best.
A: If SENTENCE_1 is true, SENTENCE_2 must be
  true.
B: SENTENCE_1 contradicts SENTENCE_2.
C: When SENTENCE_1 is true, SENTENCE_2 may
  or may not be true.

Answer only using the following format:
Answer: $OPTION
Replace $OPTION with the selected option.
  Use the letters A, B or C only as the
  answer.{fewshot_examples}

SENTENCE_1:
```
{sentence1}
```
SENTENCE_2:
```
{sentence2}
```
\end{lstlisting}
\\
\midrule
AS & 
\begin{lstlisting}
Ibuod ang sumusunod na artikulong Filipino
  sa isang talata na may isa o dalawang
  pangungusap.

Sumagot gamit ang sumusunod na format:
Buod: $SUMMARY
Palitan ang $SUMMARY ng buod.
  {fewshot_examples}

Artikulo:
```
{text}
```
\end{lstlisting}
&
\begin{lstlisting}
Summarize the following {language} article
  into a paragraph with 1 or 2 sentences.

Answer only using the following format:
Summary: $SUMMARY
Replace $SUMMARY with the summary.
  {fewshot_examples}

Article:
```
{text}
```
\end{lstlisting}
\\
\midrule
MT & 
\begin{lstlisting}
Isalin ang sumusunod na teksto sa
  {language}.

Sumagot gamit ang sumusunod na format:
Salin: $TRANSLATION
Palitan ang $TRANSLATION ng isinalin na
  teksto.{fewshot_examples}

Teksto: 
```
{text}
```
\end{lstlisting}
&
\begin{lstlisting}
Translate the following text into
  {language}.

Answer only using the following format:
Translation: $TRANSLATION
Replace $TRANSLATION with the translated
  text.{fewshot_examples}

Text:
```
{text}
```
\end{lstlisting}
\\
\bottomrule
\caption{Prompt templates used in evaluating LLMs on \textsc{Batayan}. Their corresponding English-language prompt templates are also provided.}
\label{tab:prompts}
\end{longtable}
\renewcommand{\arraystretch}{1.0}

\newpage
\section{Details of Models Evaluated}
\label{sec:models_evaluated}
\normalsize

\renewcommand{\arraystretch}{1.1}
\begin{table*}[htp]
\centering
\tiny
\begin{tabular}{llrl}
\toprule
Models & Source Link & No. params\\
\midrule
\textit{Base pre-trained models} \\
\midrule
aisingapore/Gemma-SEA-LION-v3-9B & \url{https://huggingface.co/aisingapore/Gemma-SEA-LION-v3-9B} & 9B \\
aisingapore/Llama-SEA-LION-v2-8B & \url{https://huggingface.co/aisingapore/Llama-SEA-LION-v2-8B} & 8B \\
aisingapore/Llama-SEA-LION-v3-8B & \url{https://huggingface.co/aisingapore/Llama-SEA-LION-v3-8B} & 8B \\
aisingapore/Llama-SEA-LION-v3-70B & \url{https://huggingface.co/aisingapore/Llama-SEA-LION-v3-70B} & 70B \\
google/gemma-2-9b & \url{https://huggingface.co/google/gemma-2-9b} & 9B \\
google/gemma-2-27b & \url{https://huggingface.co/google/gemma-2-27b} & 27B \\
google/gemma-3-12b-pt & \url{https://huggingface.co/google/gemma-3-12b-pt} & 12B \\
google/gemma-3-27b-pt & \url{https://huggingface.co/google/gemma-3-27b-pt} & 27B \\
meta-llama/Llama-3.1-8B & \url{https://huggingface.co/meta-llama/Llama-3.1-8B} & 8B\\
meta-llama/Llama-3.1-70B & \url{https://huggingface.co/meta-llama/Llama-3.1-70B} & 70B \\
Qwen/Qwen2-7B & \url{https://huggingface.co/Qwen/Qwen2-7B} & 7B \\
Qwen/Qwen2-72B & \url{https://huggingface.co/Qwen/Qwen2-72B} & 72B \\
Qwen/Qwen2.5-7B & \url{https://huggingface.co/Qwen/Qwen2.5-7B} & 7B \\
Qwen/Qwen2.5-14B & \url{https://huggingface.co/Qwen/Qwen2.5-14B} & 14B \\
Qwen/Qwen2.5-32B & \url{https://huggingface.co/Qwen/Qwen2.5-32B} & 32B \\
Qwen/Qwen2.5-72B & \url{https://huggingface.co/Qwen/Qwen2.5-72B} & 72B \\
sail/Sailor2-8B & \url{https://huggingface.co/sail/Sailor2-8B} & 8B \\
sail/Sailor2-20B & \url{https://huggingface.co/sail/Sailor2-20B} & 20B \\
SeaLLMs/SeaLLMs-v3-7B & \url{https://huggingface.co/SeaLLMs/SeaLLMs-v3-7B} & 7B \\
\midrule
\textit{Instruction-tuned models} \\
\midrule
aisingapore/Gemma-SEA-LION-v3-9B-IT & \url{https://huggingface.co/aisingapore/Gemma-SEA-LION-v3-9B-IT} & 9B\\
aisingapore/Llama-SEA-LION-v2-8B-IT & \url{https://huggingface.co/aisingapore/Llama-SEA-LION-v2-8B-IT} & 8B \\
aisingapore/Llama-SEA-LION-v3-8B-IT & \url{https://huggingface.co/aisingapore/Llama-SEA-LION-v3-8B-IT} & 8B \\
aisingapore/Llama-SEA-LION-v3-70B-IT & \url{https://huggingface.co/aisingapore/Llama-SEA-LION-v3-70B-IT} & 70B \\
CohereForAI/aya-23-8B & \url{https://huggingface.co/CohereForAI/aya-23-8B} & 8B \\
CohereForAI/aya-23-35B & \url{https://huggingface.co/CohereForAI/aya-23-35B} & 35B \\
CohereForAI/aya-expanse-8b & \url{https://huggingface.co/CohereForAI/aya-expanse-8b} & 8B \\
CohereForAI/aya-expanse-32b & \url{https://huggingface.co/CohereForAI/aya-expanse-32b} & 32B \\
deepseek-ai/DeepSeek-V3 & \url{https://huggingface.co/deepseek-ai/DeepSeek-V3} & 671B \\
google/gemma-2-9b-it & \url{https://huggingface.co/google/gemma-2-9b-it} & 9B \\
google/gemma-2-27b-it & \url{https://huggingface.co/google/gemma-2-27b-it} & 27B \\
google/gemma-3-12b-it & \url{https://huggingface.co/google/gemma-3-12b-it} & 12B \\
google/gemma-3-27b-it & \url{https://huggingface.co/google/gemma-3-27b-it} & 27B \\
MERaLiON/Llama-3-MERaLiON-8B-Instruct & \url{https://huggingface.co/MERaLiON/Llama-3-MERaLiON-8B-Instruct} & 8B \\
meta-llama/Llama-3.1-8B-Instruct & \url{https://huggingface.co/meta-llama/Llama-3.1-8B-Instruct} & 8B \\
meta-llama/Llama-3.1-70B-Instruct & \url{https://huggingface.co/meta-llama/Llama-3.1-70B-Instruct} & 70B \\
meta-llama/Llama-3.3-70B-Instruct & \url{https://huggingface.co/meta-llama/Llama-3.3-70B-Instruct} & 70B \\
Qwen/Qwen2-7B-Instruct & \url{https://huggingface.co/Qwen/Qwen2-7B-Instruct} & 7B \\
Qwen/Qwen2-72B-Instruct & \url{https://huggingface.co/Qwen/Qwen2-72B-Instruct} & 72B \\
Qwen/Qwen2.5-7B-Instruct & \url{https://huggingface.co/Qwen/Qwen2.5-7B-Instruct} & 7B \\
Qwen/Qwen2.5-14B-Instruct & \url{https://huggingface.co/Qwen/Qwen2.5-14B-Instruct} & 14B \\
Qwen/Qwen2.5-32B-Instruct & \url{https://huggingface.co/Qwen/Qwen2.5-32B-Instruct} & 32B \\
Qwen/Qwen2.5-72B-Instruct & \url{https://huggingface.co/Qwen/Qwen2.5-72B-Instruct} & 72B \\
sail/Sailor2-8B-Chat & \url{https://huggingface.co/sail/Sailor2-8B-Chat} & 8B \\
sail/Sailor2-20B-Chat & \url{https://huggingface.co/sail/Sailor2-20B-Chat} & 20B \\
SeaLLMs/SeaLLMs-v3-7B-Chat & \url{https://huggingface.co/SeaLLMs/SeaLLMs-v3-7B-Chat} & 7B \\
Tower-Babel/Babel-9B-Chat & \url{https://huggingface.co/Tower-Babel/Babel-9B-Chat} & 9B \\
Tower-Babel/Babel-83B-Chat & \url{https://huggingface.co/Tower-Babel/Babel-83B-Chat} & 83B \\
\midrule
\textit{Reasoning models} \\
\midrule
deepseek-ai/DeepSeek-R1 & \url{https://huggingface.co/deepseek-ai/DeepSeek-R1} & 671B \\
\midrule
\textit{Commercial models} \\
\midrule
google/gemini-1.5-flash-002 & \url{https://cloud.google.com/vertex-ai/generative-ai/docs/learn/model-versions} & - \\
google/gemini-1.5-pro-002 & \url{https://cloud.google.com/vertex-ai/generative-ai/docs/learn/model-versions} & - \\
google/gemini-2.0-flash-001 & \url{https://cloud.google.com/vertex-ai/generative-ai/docs/learn/model-versions} & - \\
google/gemini-2.0-flash-lite-001 & \url{https://cloud.google.com/vertex-ai/generative-ai/docs/learn/model-versions} & - \\
openai/gpt-4o-mini-2024-07-18 & \url{https://platform.openai.com/docs/models/gpt-4o-mini} & - \\
openai/gpt-4o-2024-08-06 & \url{https://platform.openai.com/docs/models/gpt-4o} & - \\
\bottomrule
\end{tabular}
\caption{Links and sizes of models evaluated in our experiments.}
\label{tab:model_links_sizes}
\end{table*}
\renewcommand{\arraystretch}{1.0}

\renewcommand{\arraystretch}{1.4}
\begin{table*}[htp]
\centering
\small
\begin{tabular}{lp{3cm}cp{6.5cm}}
\toprule
Model & Reference & Filipino IT & Rationale for selection
\\
\midrule
Aya 23 and Expanse & \cite{aryabumi2024aya23openweight, dang2024ayaexpansecombiningresearch} & & LLMs with multilingual support. It is unknown if these models were explicitly fine-tuned on Filipino instructions.
\\
Babel & \cite{zhao2025babelopenmultilinguallarge} & \checkmark & LLMs with multilingual support including Filipino. This model was pre-trained and fine-tuned on Filipino tokens.
\\
DeepSeek V3 and R1 & \cite{deepseekai2025deepseekr1incentivizingreasoningcapability} & & LLMs with multilingual support. It is unknown if these models were explicitly fine-tuned on Filipino instructions.
\\
Gemma 2 and 3 & \cite{gemmateam2024gemma2improvingopen, gemmateam2025gemma3technicalreport} & & LLMs with multilingual support. It is unknown if these models were explicitly fine-tuned on Filipino instructions.
\\
Gemini 1.5 and 2 & \cite{geminiteam2024gemini15unlockingmultimodal} & & Commercial AI systems with multilingual support. It is unknown if these models were explicitly fine-tuned on Filipino instructions.
\\
GPT-4o and 4o-mini & \cite{openai2024gpt4ocard} & & Commercial AI systems with multilingual support. It is unknown if these models were explicitly fine-tuned on Filipino instructions.
\\
Llama 3 & \cite{grattafiori2024llama3herdmodels} & &LLMs with multilingual support. It is unknown if these models were explicitly fine-tuned on Filipino instructions.
\\
MERaLiON & \cite{huang2025meraliontextllmcrosslingualunderstandinglarge} & & LLM with support for English, Chinese, and Indonesian. Llama 3 model was continue pre-trained on 120B English, Chinese, and Indonesian tokens. It is unknown if these models were explicitly fine-tuned on Filipino instructions.
\\
Qwen 2 and 2.5 & \cite{yang2024qwen2technicalreport, qwen2025qwen25technicalreport} & \checkmark & LLMs with multilingual support, including Tagalog.
\\
Sailor 2 & \cite{dou2025sailor2sailingsoutheastasia} & \checkmark & LLMs with support for Southeast Asian languages. Qwen 2.5 7B model was continue pre-trained on 400B SEA tokens, and was specifically fine-tuned on Filipino instructions.
\\
SEA-LION v2 and v3 & \cite{ng2025sealionsoutheastasianlanguages} & \checkmark & LLMs with support for Southeast Asian languages. Llama 3 and Gemma 2 series models were continue pre-trained on 200B tokens across 11 Southeast Asian languages, and was specifically fine-tuned on Filipino instructions.
\\
SeaLLMs v3 & \cite{zhang2024seallms3openfoundation} & \checkmark & LLMs with support for Southeast Asian languages. These models were pre-trained and fine-tuned on Filipino tokens.
\\
\bottomrule
\end{tabular}
\caption{Descriptions of model series evaluated in our experiments.}
\label{tab:model_descriptions}
\end{table*}
\renewcommand{\arraystretch}{1.0}

\newpage
\section{Experimental Setup}
\label{sec:experimental_setup}
\normalsize

The default evaluation setting for \textsc{Batayan} is zero-shot prompting, however base pre-trained models with no instruction-tuning are evaluated on a five-shot setting. Performance scores are calculated based on a single run. Table \ref{tab:inference_parameters} details the decoding parameters used in the experiments.

\renewcommand{\arraystretch}{1.2}
\begin{table*}[htp]
\centering
\small
\begin{tabular}{llrrrrr}
\toprule
\textbf{Competency} &\textbf{Task} & \texttt{max\_tokens} & \texttt{temperature} & \texttt{top\_p} & \texttt{top\_k} & \texttt{repetition\_penalty} \\
\midrule
\multirow{1}{*}{NLU} & PI & 32 & 0.0 & 1.0 & 1.0 & 1.0 \\
& QA & 32 & 0.0 & 1.0 & 1.0 & 1.0 \\
& SA & 32 & 0.0 & 1.0 & 1.0 & 1.0 \\
& TD & 32 & 0.0 & 1.0 & 1.0 & 1.0 \\
\midrule
\multirow{1}{*}{NLR} & CR & 32 & 0.0 & 1.0 & 1.0 & 1.0 \\
& NLI & 32 & 0.0 & 1.0 & 1.0 & 1.0 \\
\midrule
\multirow{1}{*}{NLG} &AS & 512 & 0.3 & 1.0 & 1.0 & 1.0 \\
& MT & 256 & 0.0 & 1.0 & 1.0 & 1.0 \\
\bottomrule
\end{tabular}
\caption{Inference Parameters per task.}
\label{tab:inference_parameters}
\end{table*}
\renewcommand{\arraystretch}{1.0}

\newpage
\onecolumn
\section{Complete Experimental Results}
\label{sec:complete_experimental_results}

\renewcommand{\arraystretch}{1.1} 
\begin{table*}[h]
\centering
\small
\begin{threeparttable}
\begin{tabular}{lrrrrrrrr}
\toprule
Model & PI & QA & SA & TD & NLU Avg. & CR & NLI & NLR Avg. \\
\midrule
\textit{Base pre-trained models (five-shot)} \\
\midrule
aisingapore/Gemma-SEA-LION-v3-9B                & 82.40 & \textbf{89.02} & 67.60 & 51.54 & 72.64 & 0.00\tnote{a}  & \textbf{63.93} & \textbf{31.97} \\
aisingapore/Llama-SEA-LION-v2-8B                & 71.21 & 69.54 & 48.86 & 52.89 & 60.62 & 0.00\tnote{a}  & 28.99 & 14.49 \\
aisingapore/Llama-SEA-LION-v3-8B                & 71.96 & 71.07 & 38.58 & 66.44 & 62.01 & 0.00\tnote{a}  & 48.59 & 24.30 \\
google/gemma-2-9b                               & 70.64 & 85.79 & 51.09 & \textbf{74.26} & 70.44 & 0.00\tnote{a}  & 46.04 & 23.02 \\
meta-llama/Llama-3.1-8B                         & 60.31 & 71.91 & 50.16 & 58.51 & 60.22 & 0.00\tnote{a}  & 26.82 & 13.41 \\
Qwen/Qwen2-7B                                   & 84.69 & 74.12 & 72.11 & 65.48 & \textbf{74.10} & 0.00\tnote{a}  & 45.85 & 22.92 \\
Qwen/Qwen2.5-7B                                 & 84.48 & 69.89 & \textbf{72.22} & 64.71 & 72.82 & 0.00\tnote{a}  & 41.11 & 20.55 \\
sail/Sailor2-8B                                 & 79.44 & 75.93 & 70.25 & 67.64 & 73.32 & 0.00\tnote{a}  & 43.84 & 21.92 \\
SeaLLMs/SeaLLMs-v3-7B                           & \textbf{84.95} & 74.05 & 71.12 & 63.54 & 73.41 & 0.00\tnote{a}  & 41.46 & 20.73 \\
\midrule
\textit{Instruction-tuned models (zero-shot)} \\
\midrule
aisingapore/Gemma-SEA-LION-v3-9B-IT             & 82.00 & 82.80 & \textbf{75.99} & \textbf{72.85} & \textbf{78.41} & \textbf{92.75} & \textbf{68.64} & \textbf{80.69} \\
aisingapore/Llama-SEA-LION-v2-8B-IT             & 68.50 & 45.41 & 49.81 & 58.04 & 55.44 & 50.23 & 37.63 & 43.93 \\
aisingapore/Llama-SEA-LION-v3-8B-IT             & 74.97 & 80.69 & 51.24 & 60.57 & 66.87 & 79.44 & 65.87 & 72.65 \\
CohereForAI/aya-23-8B                           & 24.68 & 33.50 & 33.49 & 32.70 & 31.09 & 28.12 & 16.67 & 22.39 \\
CohereForAI/aya-expanse-8b                      & 32.65 & 2.97\tnote{a} & 57.44 & 34.39 & 31.86 & 1.96\tnote{a}  & 25.27 & 13.61 \\
google/gemma-2-9b-it                            & 82.74 & \textbf{83.96} & 70.80 & 64.48 & 75.50 & 91.25 & 60.24 & 75.74 \\
MERaLiON/Llama-3-MERaLiON-8B-Instruct           & 23.47 & 64.90 & 35.16 & 27.64 & 37.79 & 28.97 & 23.35 & 26.16 \\
meta-llama/Llama-3.1-8B-Instruct                & 61.38 & 74.90 & 52.55 & 50.00 & 59.71 & 72.66 & 56.28 & 64.47 \\
Qwen/Qwen2-7B-Instruct                          & 73.08 & 56.01 & 52.63 & 26.10 & 51.96 & 43.99 & 32.35 & 38.17 \\
Qwen/Qwen2.5-7B-Instruct                        & \textbf{85.22} & 69.07 & 67.82 & 60.84 & 70.74 & 46.15 & 57.80 & 51.98 \\
sail/Sailor2-8B-Chat                            & 73.60 & 59.48 & 48.67 & 47.24 & 57.25 & 57.52 & 61.21 & 59.36 \\
SeaLLMs/SeaLLMs-v3-7B-Chat                      & 37.85 & 49.72 & 70.63 & 61.48 & 54.92 & 71.82 & 20.72 & 46.27 \\
Tower-Babel/Babel-9B-Chat                       & 78.76 & 61.68 & 72.52 & 56.74 & 67.42 & 83.00 & 26.89 & 54.95 \\
\bottomrule
\end{tabular}
\begin{tablenotes}
    \item[a] Could not follow answer instructions.
\end{tablenotes}
\caption{Performance of small LLMs (<11B parameters) on \textsc{Batayan} natural language understanding (NLU) and natural language reasoning (NLR) tasks. Macro F1 scores are reported.}
\label{tab:small_nlu_nlr}
\end{threeparttable}
\end{table*}
\renewcommand{\arraystretch}{1.0} 

\renewcommand{\arraystretch}{1.1}
\begin{table*}[t]
\centering
\small
\begin{threeparttable}
\begin{tabular}{lrrrrrrrr}
\toprule
Model & PI & QA & SA & TD & NLU Avg. & CR & NLI & NLR Avg. \\
\midrule
\textit{Base pre-trained models (five-shot)} \\
\midrule
google/gemma-2-27b           & 72.63 & \textbf{86.04} & 69.60 & \textbf{75.95} & 76.05 & 0.00\tnote{a} & 56.21 & 28.11 \\
google/gemma-3-12b-pt        & 71.42 & 86.02 & \textbf{79.30} & 51.60 & 72.08 & 0.00\tnote{a} & 59.47 & 29.74 \\ 
google/gemma-3-27b-pt        & 59.29 & 84.83 & 69.84 & 52.34 & 66.58 & 0.00\tnote{a} & 55.13 & 27.56 \\
Qwen/Qwen2.5-14B             & \textbf{89.25} & 82.03 & 68.36 & 62.89 & 75.63 & 0.00\tnote{a} & 53.10 & 26.55 \\
Qwen/Qwen2.5-32B             & 87.69 & 84.98 & 67.71 & 66.62 & \textbf{76.75} & 0.00\tnote{a} & 55.09 & 27.54 \\
sail/Sailor2-20B             & 85.46 & 84.94 & 68.47 & 48.83 & 71.92 & 0.00\tnote{a} & \textbf{63.09} & \textbf{31.55} \\
\midrule
\textit{Instruction-tuned models (zero-shot)} \\
\midrule
CohereForAI/aya-expanse-32b  & 76.39 & 54.82 & 63.25 & 74.70 & 67.29 & 55.46 & 66.57 & 61.01 \\
google/gemma-3-12b-it        & 82.27 & 83.94 & 62.10 & 80.50 & 77.20 & 93.75 & 70.64 & \textbf{82.19} \\
google/gemma-2-27b-it        & 70.68 & 69.18 & 70.39 & 80.36 & 72.65 & 67.76 & \textbf{80.25} & 73.04 \\
google/gemma-3-27b-it        & 70.20 & \textbf{87.91} & 67.77 & \textbf{81.74} & 76.91 & \textbf{94.75} & 67.73 & 81.24 \\
Qwen/Qwen2.5-14B-Instruct    & 63.75 & 64.64 & 72.34 & 72.73 & 68.37 & 53.47 & 43.83 & 48.65 \\
Qwen/Qwen2.5-32B-Instruct    & \textbf{83.70} & 84.97 & \textbf{74.60} & 75.50 & \textbf{79.69} & 85.98 & 60.58 & 73.28 \\
sail/Sailor2-20B-Chat        & 39.17 & 13.81 & 54.41 & 74.03 & 45.36 & 22.98 & 47.88 & 35.43 \\
\bottomrule
\end{tabular}
\begin{tablenotes}
    \item[a] Could not follow answer instructions.
\end{tablenotes}
\caption{Performance of medium LLMs (<32B parameters)  on \textsc{Batayan} NLU/NLR tasks. Macro F1 scores are reported.}
\label{tab:medium_nlu_nlr}
\end{threeparttable}
\end{table*}
\renewcommand{\arraystretch}{1.0}

\renewcommand{\arraystretch}{1.1}
\begin{table*}[t]
\centering
\small
\begin{threeparttable}
\begin{tabular}{lrrrrrrrr}
\toprule
Model & PI & QA & SA & TD & NLU Avg. & CR & NLI & NLR Avg. \\
\midrule
\textit{Base pre-trained models (five-shot)} \\
\midrule
aisingapore/Llama-SEA-LION-v3-70B                 & 83.48 & 87.02 & 45.89 & 77.49 & 73.47 & 0.00\tnote{a}  & \textbf{79.62} & \textbf{39.81} \\
meta-llama/Llama-3.1-70B                          & 83.49 & 86.05 & 59.72 & 51.70 & 70.24 & 0.00\tnote{a}  & 55.78 & 27.89 \\
Qwen/Qwen2-72B                                    & \textbf{90.25} & \textbf{88.86} & 67.88 & 77.51 & 81.13 & 0.00\tnote{a}  & 69.63 & 34.81 \\
Qwen/Qwen2.5-72B                                  & 86.89 & 86.79 & \textbf{74.73} & \textbf{79.61} & \textbf{82.01} & 0.00\tnote{a}  & 67.58 & 33.79 \\
\midrule
\textit{Instruction-tuned models (zero-shot)} \\
\midrule
aisingapore/Llama-SEA-LION-v3-70B-IT              & 84.07 & 87.01 & 68.79 & 76.99 & 79.21 & \textbf{94.75} & \textbf{75.34} & \textbf{85.04} \\
CohereForAI/aya-23-35B                            & 81.83 & 61.46 & 61.68 & 53.06 & 64.51 & 74.16 & 40.54 & 57.35 \\
deepseek-ai/DeepSeek-V3                           & 83.07 & \textbf{90.00} & \textbf{81.78} & 76.21 & \textbf{82.77} & 94.25 & 63.25 & 78.75 \\
meta-llama/Llama-3.1-70B-Instruct                 & 83.57 & 88.13 & 67.29 & 63.80 & 75.70 & 93.49 & 72.62 & 83.06 \\
meta-llama/Llama-3.3-70B-Instruct                 & 82.62 & 86.96 & 66.12 & 77.40 & 78.28 & 94.25 & 72.97 & 83.61 \\
Qwen/Qwen2-72B-Instruct                           & \textbf{85.37} & 86.10 & 76.16 & 76.38 & 81.00 & 58.54 & 68.39 & 63.47 \\
Qwen/Qwen2.5-72B-Instruct                         & 80.40 & 84.00 & 75.32 & 73.44 & 78.29 & 91.75 & 63.95 & 77.85 \\
Tower-Babel/Babel-83B-Chat                        & 85.06 & 33.12 & 73.20 & \textbf{81.24} & 68.15 & 49.62 & 68.14 & 58.88 \\
\midrule
\textit{Reasoning models (zero-shot)} \\
\midrule
deepseek-ai/DeepSeek-R1                           & 79.84 & 87.93 & 71.47 & 71.47 & 77.68 & 97.00 & 71.06 & 84.03 \\
\bottomrule
\end{tabular}
\begin{tablenotes}
    \item[a] Could not follow answer instructions.
\end{tablenotes}
\caption{Performance of large LLMs (>=32B parameters) on \textsc{Batayan} NLU/NLR tasks. Macro F1 scores are reported.}
\label{tab:large_nlu_nlr}
\end{threeparttable}
\end{table*}
\renewcommand{\arraystretch}{1.0}

\renewcommand{\arraystretch}{1.1}
\begin{table*}[t]
\centering
\small
\begin{tabular}{lrrrrrrrr}
\toprule
Model & PI & QA & SA & TD & NLU Avg. & CR & NLI & NLR Avg. \\
\midrule
google/gemini-1.5-flash-002           & 85.71 & 84.98 & 63.81 & 66.04 & 75.14 & 94.24 & 69.19 & 81.72 \\
google/gemini-1.5-pro-002             & 85.62 & \textbf{89.95} & 69.48 & 79.27 & 81.08 & 95.74 & \textbf{76.71} & \textbf{86.23} \\
google/gemini-2.0-flash-001           & 83.98 & 88.05 & 73.93 & \textbf{80.49} & \textbf{81.61} & \textbf{97.74} & 70.47 & 84.11 \\
google/gemini-2.0-flash-lite-001      & 85.02 & 85.97 & 64.33 & 78.94 & 78.57 & 96.49 & 70.20 & 83.35 \\
openai/gpt-4o-mini-2024-07-18         & 86.11 & 85.04 & 70.77 & 70.77 & 78.17 & 92.99 & 68.18 & 80.59 \\
openai/gpt-4o-2024-08-06              & \textbf{88.21} & 71.88 & \textbf{74.78} & 74.78 & 77.41 & 97.25 & 73.33 & 85.29 \\
\bottomrule
\end{tabular}
\caption{Performance of commercial LLMs on \textsc{Batayan} NLU/NLR tasks. Macro F1 scores are reported.}
\label{tab:commercial_nlu_nlr}
\end{table*}
\renewcommand{\arraystretch}{1.0}

\renewcommand{\arraystretch}{1.1}
\begin{table*}[t]
\centering
\small
\begin{tabular}{lrrrrrrrrrr}
\toprule
\multicolumn{1}{l}{\multirow{2}{*}{Models}} & \multicolumn{3}{c}{\multirow{1}{*}{AS}} & \multicolumn{2}{c}{\multirow{1}{*}{MT (\textit{eng}→\textit{tgl})}} & \multicolumn{2}{c}{\multirow{1}{*}{MT (\textit{tgl}→\textit{eng})}} & NLG
\\ \cline{2-8} 
& {\tiny BERTScore} & {\tiny ChrF++} & {\tiny ROUGE-L} & {\tiny ChrF++} & {\tiny MetricX-24} & {\tiny ChrF++} & {\tiny MetricX-24} & Avg. \\
\midrule
\textit{Base pre-trained models (five-shot)} \\
\midrule
aisingapore/Gemma-SEA-LION-v3-9B                         & 77.01 & \textbf{34.19} & 33.71 & 58.90 & 82.75 & 56.63 & 91.15 & 62.05 \\
aisingapore/Llama-SEA-LION-v2-8B                         & 75.86 & 27.17 & 30.88 & 48.45 & 62.74 & 58.17 & 87.10 & 55.77 \\
aisingapore/Llama-SEA-LION-v3-8B                         & 75.73 & 32.41 & 30.56 & 53.10 & 72.41 & 57.70 & 88.91 & 58.69 \\
google/gemma-2-9b                                        & 76.60 & 31.72 & 31.69 & 59.06 & 82.01 & 63.27 & \textbf{91.16} & \textbf{62.22} \\
meta-llama/Llama-3.1-8B                                  & 75.80 & 29.93 & 30.13 & 51.94 & 71.10 & 58.44 & 88.77 & 58.01 \\
Qwen/Qwen2-7B                                            & 73.84 & 27.64 & 25.71 & 42.61 & 47.37 & 53.60 & 82.13 & 50.41 \\
Qwen/Qwen2.5-7B                                          & 73.26 & 26.71 & 25.04 & 38.89 & 40.28 & 42.59 & 81.58 & 46.91 \\
sail/Sailor2-8B                                          & \textbf{77.05} & 30.69 & \textbf{34.50} & \textbf{62.01} & \textbf{86.30} & \textbf{63.41} & 91.05 & 63.57 \\
SeaLLMs/SeaLLMs-v3-7B                                    & 73.78 & 28.37 & 26.57 & 43.45 & 47.79 & 44.54 & 83.74 & 49.75 \\
\midrule
\textit{Instruction-tuned models (zero-shot)} \\
\midrule
aisingapore/Gemma-SEA-LION-v3-9B-IT                      & 73.00 & 34.16 & 20.99 & \textbf{57.79} & \textbf{87.24} & \textbf{58.63} & \textbf{90.56} & \textbf{60.35} \\
aisingapore/Llama-SEA-LION-v2-8B-IT                      & 73.39 & 32.91 & 24.14 & 50.90 & 64.62 & 48.49 & 55.04 & 49.93 \\
aisingapore/Llama-SEA-LION-v3-8B-IT                      & 73.34 & 33.80 & 23.43 & 57.14 & 82.67 & 58.33 & 87.69 & 59.49 \\
CohereForAI/aya-23-8B                                    & 67.37 & 24.01 & 11.76 & 15.61 &  9.68 & 46.20 & 58.59 & 33.32 \\
CohereForAI/aya-expanse-8b                               & 59.48 &  6.69 &  1.64 & 39.30 & 37.47 & 49.81 & 78.88 & 39.04 \\
google/gemma-2-9b-it                                     & 72.73 & 34.19 & 21.81 & 56.80 & 84.23 & 50.97 & 76.62 & 56.76 \\
MERaLiON/Llama-3-MERaLiON-8B-Instruct                    & 74.08 & 31.79 & 25.68 & 49.77 & 64.85 & 41.66 & 60.14 & 49.71 \\
meta-llama/Llama-3.1-8B-Instruct                         & \textbf{74.24} & \textbf{34.27} & \textbf{26.94} & 44.79 & 51.76 & 46.33 & 70.65 & 49.85 \\
Qwen/Qwen2-7B-Instruct                                   & 69.51 & 26.09 & 13.58 & 42.22 & 42.86 & 45.13 & 50.01 & 41.34 \\
Qwen/Qwen2.5-7B-Instruct                                 & 70.61 & 28.89 & 15.86 & 38.99 & 34.57 & 49.02 & 76.36 & 44.91 \\
sail/Sailor2-8B-Chat                                     & 71.82 & 31.83 & 19.75 & 54.04 & 80.93 & 35.41 & 19.66 & 44.78 \\
SeaLLMs/SeaLLMs-v3-7B-Chat                               & 70.34 & 27.84 & 16.49 & 46.36 & 61.07 & 43.27 & 60.37 & 46.53 \\
Tower-Babel/Babel-9B-Chat                                & 71.61 & 31.55 & 18.93 & 49.97 & 72.50 & 48.69 & 75.69 & 52.70 \\
\bottomrule
\end{tabular}
\caption{Performance of small LLMs (<11B parameters) on \textsc{Batayan} natural language generation (NLG) tasks.}
\label{tab:small_nlg}
\end{table*}
\renewcommand{\arraystretch}{1.0}

\renewcommand{\arraystretch}{1.1}
\begin{table*}[t]
\centering
\small
\begin{tabular}{lrrrrrrrrrr}
\toprule
\multicolumn{1}{l}{\multirow{2}{*}{Models}} & \multicolumn{3}{c}{\multirow{1}{*}{AS}} & \multicolumn{2}{c}{\multirow{1}{*}{MT (\textit{eng}→\textit{tgl})}} & \multicolumn{2}{c}{\multirow{1}{*}{MT (\textit{tgl}→\textit{eng})}} & NLG
\\ \cline{2-8} 
& {\tiny BERTScore} & {\tiny ChrF++} & {\tiny ROUGE-L} & {\tiny ChrF++} & {\tiny MetricX-24} & {\tiny ChrF++} & {\tiny MetricX-24} & Avg. \\
\midrule
\textit{Base pre-trained models (five-shot)} \\
\midrule
google/gemma-2-27b                          & 77.27 & 33.71 & 33.50 & 61.16 & 85.56 & 64.77 & 91.82 & 63.97 \\
google/gemma-3-12b-pt                       & 77.42 & 32.12 & 34.32 & 61.66 & 85.52 & 64.69 & 92.08 & 63.97 \\
google/gemma-3-27b-pt                       & \textbf{77.70} & \textbf{35.53} & \textbf{35.36} & \textbf{62.89} & \textbf{87.06} & \textbf{64.98} & \textbf{92.27} & \textbf{65.11} \\
Qwen/Qwen2.5-14B                            & 74.39 & 30.34 & 27.87 & 49.01 & 62.69 & 59.41 & 88.63 & 56.05 \\
Qwen/Qwen2.5-32B                            & 75.13 & 29.73 & 29.31 & 50.94 & 65.06 & 60.18 & 89.34 & 57.10 \\
sail/Sailor2-20B                            & 76.73 & 32.41 & 33.47 & 61.75 & 86.30 & 63.88 & 91.73 & 63.75 \\
\midrule
\textit{Instruction-tuned models (zero-shot)} \\
\midrule
CohereForAI/aya-expanse-32b                 & 71.18 & 30.12 & 16.85 & 49.14 & 62.76 & 57.35 & 85.05 & 53.21 \\
google/gemma-2-27b-it                       & \textbf{73.41} & \textbf{35.11} & \textbf{23.2}1 & 59.51 & 88.19 & 61.87 & 91.68 & \textbf{61.86} \\
google/gemma-3-12b-it                       & 72.10 & 32.80 & 19.43 & 60.09 & 89.39 & 61.85 & 91.61 & 61.04 \\
google/gemma-3-27b-it                       & 72.25 & 33.26 & 19.55 & \textbf{61.02} & \textbf{90.71} & \textbf{62.09} & \textbf{92.18} & 61.58 \\
Qwen/Qwen2.5-14B-Instruct                   & 71.92 & 31.85 & 19.24 & 44.78 & 54.51 & 58.01 & 86.08 & 52.34 \\
Qwen/Qwen2.5-32B-Instruct                   & 72.47 & 33.43 & 21.22 & 47.47 & 59.47 & 58.77 & 86.90 & 54.25 \\
sail/Sailor2-20B-Chat                       & 71.47 & 31.20 & 18.41 & 49.45 & 72.51 & 49.52 & 72.79 & 52.19 \\
\bottomrule
\end{tabular}
\caption{Performance of medium LLMs (<32B parameters) on \textsc{Batayan} NLG tasks.}
\label{tab:medium_nlg}
\end{table*}
\renewcommand{\arraystretch}{1.0}

\renewcommand{\arraystretch}{1.1}
\begin{table*}[t]
\centering
\small
\begin{tabular}{lrrrrrrrrrr}
\toprule
\multicolumn{1}{l}{\multirow{2}{*}{Models}} & \multicolumn{3}{c}{\multirow{1}{*}{AS}} & \multicolumn{2}{c}{\multirow{1}{*}{MT (\textit{eng}→\textit{tgl})}} & \multicolumn{2}{c}{\multirow{1}{*}{MT (\textit{tgl}→\textit{eng})}} & NLG
\\ \cline{2-8} 
& {\tiny BERTScore} & {\tiny ChrF++} & {\tiny ROUGE-L} & {\tiny ChrF++} & {\tiny MetricX-24} & {\tiny ChrF++} & {\tiny MetricX-24} & Avg. \\
\midrule
\textit{Base pre-trained models (five-shot)} \\
\midrule
aisingapore/Llama-SEA-LION-v3-70B                 & \textbf{76.87} & \textbf{35.49} & \textbf{33.66} & \textbf{60.70} & \textbf{85.28} & 63.39 & 91.52 & \textbf{63.85} \\
meta-llama/Llama-3.1-70B                          & 76.80 & 33.98 & 33.54 & 60.20 & 84.32 & \textbf{65.80} & \textbf{91.85} & 63.79 \\
Qwen/Qwen2-72B                                    & 76.81 & 34.15 & 33.24 & 54.75 & 74.84 & 65.13 & 90.87 & 61.40 \\
Qwen/Qwen2.5-72B                                  & 76.41 & 34.12 & 31.96 & 54.69 & 74.92 & 64.50 & 90.85 & 61.06 \\
\midrule
\textit{Instruction-tuned models (zero-shot)} \\
\midrule
aisingapore/Llama-SEA-LION-v3-70B-IT              & 73.25 & 34.82 & 23.81 & 61.02 & 86.24 & 63.11 & 91.28 & 61.93 \\
CohereForAI/aya-23-35B                            & 72.81 & 29.94 & 22.08 & 46.00 & 56.79 & 49.54 & 72.45 & 49.94 \\
deepseek-ai/DeepSeek-V3                           & 73.19 & 35.16 & 22.99 & \textbf{61.61} & \textbf{89.21} & 63.06 & \textbf{92.28} & \textbf{62.50} \\
meta-llama/Llama-3.1-70B-Instruct                 & \textbf{74.76} & \textbf{35.56} & \textbf{27.94} & 60.32 & 84.73 & 62.78 & 91.20 & 62.47 \\
meta-llama/Llama-3.3-70B-Instruct                 & 72.90 & 34.03 & 22.63 & 58.98 & 83.05 & 61.24 & 90.72 & 60.51 \\
Qwen/Qwen2-72B-Instruct                           & 67.05 & 25.08 & 11.38 & 52.84 & 70.63 & 60.05 & 88.73 & 53.68 \\
Qwen/Qwen2.5-72B-Instruct                         & 71.62 & 31.52 & 18.02 & 52.01 & 72.30 & \textbf{64.33} & 90.02 & 57.12 \\
Tower-Babel/Babel-83B-Chat                        & 67.72 & 20.58 & 13.35 & 53.29 & 76.61 & 51.56 & 85.42 & 52.65 \\
\midrule
\textit{Reasoning models (zero-shot)} \\
\midrule
deepseek-ai/DeepSeek-R1                           & 71.97 & 32.70 & 19.59 & 59.15 & 90.91 & 61.01 & 92.30 & 61.09 \\
\bottomrule
\end{tabular}
\caption{Performance of large LLMs (>=32B parameters) on \textsc{Batayan} NLG tasks.}
\label{tab:large_nlg}
\end{table*}
\renewcommand{\arraystretch}{1.0}

\renewcommand{\arraystretch}{1.1}
\begin{table*}[t]
\centering
\small
\begin{tabular}{lrrrrrrrrrr}
\toprule
\multicolumn{1}{l}{\multirow{2}{*}{Models}} & \multicolumn{3}{c}{\multirow{1}{*}{AS}} & \multicolumn{2}{c}{\multirow{1}{*}{MT (\textit{eng}→\textit{tgl})}} & \multicolumn{2}{c}{\multirow{1}{*}{MT (\textit{tgl}→\textit{eng})}} & NLG
\\ \cline{2-8} 
& {\tiny BERTScore} & {\tiny ChrF++} & {\tiny ROUGE-L} & {\tiny ChrF++} & {\tiny MetricX-24} & {\tiny ChrF++} & {\tiny MetricX-24} & Avg.\\
\midrule
google/gemini-1.5-flash-002           & 72.44 & 34.00 & 21.20 & 61.01 & 90.11 & 65.23 & 92.04 & 62.29 \\
google/gemini-1.5-pro-002             & \textbf{73.62} & \textbf{36.16} & \textbf{24.00} & 61.07 & \textbf{90.90} & 64.14 & 91.30 & 63.03 \\
google/gemini-2.0-flash-001           & 72.99 & 34.85 & 21.96 & \textbf{63.88} & 90.87 & \textbf{65.64} & \textbf{92.53} & \textbf{63.25} \\
google/gemini-2.0-flash-lite-001      & 72.83 & 34.80 & 21.87 & 63.45 & 90.06 & 65.32 & 91.51 & 62.83 \\
openai/gpt-4o-mini-2024-07-18         & 72.70 & 33.97 & 21.67 & 62.46 & 88.59 & 63.12 & 91.68 & 62.03 \\
openai/gpt-4o-2024-08-06              & 72.83 & 34.36 & 21.55 & 63.86 & 90.19 & 64.89 & 92.51 & 62.88 \\
\bottomrule
\end{tabular}
\caption{Performance of commercial LLMs on \textsc{Batayan} NLG tasks.}
\label{tab:commercial_nlg}
\end{table*}
\renewcommand{\arraystretch}{1.0}

\end{document}